\documentclass{elsarticle}
\pdfoutput=1
\usepackage[titletoc]{appendix}
\usepackage{pdflscape}
\usepackage{booktabs}
\usepackage{flushend}
\usepackage{rotating}
\usepackage[left=2.5cm, right=2.5cm, top=2cm]{geometry}
\usepackage[usenames, dvipsnames]{color}
\usepackage{todonotes}
\usepackage{arydshln}
\usepackage[caption=false,font=normalsize,labelfont=sf,textfont=sf]{subfig}
\usepackage[cmex10]{amsmath}
\usepackage{multirow}
\usepackage{framed,lipsum}

\usepackage{algcompatible}
\usepackage{algorithm}
\usepackage{varwidth}
\usepackage[noend]{algpseudocode}
\DeclareCaptionType{mycapequ}[Equation][]
\algnewcommand{\LineComment}[1]{\State \(\triangleright\) #1}

\makeatletter
\def\BState{\State\hskip-\ALG@thistlm}
\makeatother

\algdef{SE}[VARIABLES]{Variables}{EndVariables}
   {\algorithmicvariables}
   {\algorithmicend\ \algorithmicvariables}
\algnewcommand{\algorithmicvariables}{\textbf{global variables}}

\usepackage{hyperref}
\hypersetup{
    colorlinks=true,
    linkcolor=blue,
    filecolor=magenta,
    urlcolor=cyan,
}

\makeatletter
\def\ps@pprintTitle{%
 \let\@oddhead\@empty
 \let\@evenhead\@empty
 \def\@oddfoot{}%
 \let\@evenfoot\@oddfoot}
\makeatother

\newcommand{\eg}{{\textit{e.g.}}}
\newcommand{\ie}{{\textit{i.e.}}}
\newcommand{\etal}{{\textit{et al.}}}
\newcommand{\lstm}{LSTM RNNs }

\begin{document}

\begin{frontmatter}

\title{Optimizing Long Short-Term Memory Recurrent Neural Networks Using Ant Colony Optimization to Predict Turbine Engine Vibration}

\author[mymainaddress]{AbdElRahman ElSaid\corref{mycorrespondingauthor}}
\cortext[mycorrespondingauthor]{Corresponding author}
\ead{abdelrahman.elsaid@ndus.edu}

\author[mymainaddress]{Fatima El Jamiy}
\ead{fatima.eljamiy@und.edu}

\author[mysecondaryaddress]{James Higgins}
\ead{ jhiggins@aero.und.edu}

\author[mysecondaryaddress]{Brandon Wild}
\ead{bwild@aero.und.edu}

\author[mymainaddress]{Travis Desell}
\ead[url]{http://people.cs.und.edu/~tdesell}
\ead{tdesell@cs.und.edu}

\address[mymainaddress]{Department of Computer Science}
\address[mysecondaryaddress]{Department of Aviation}
\address{University of North Dakota, Grand Forks, North Dakota 58202}

\begin{abstract}
	
This article expands on research that has been done to develop a recurrent neural network (RNN) capable of predicting aircraft engine vibrations using long short-term memory (LSTM) neurons. LSTM RNNs can provide a more generalizable and robust method for prediction over analytical calculations of engine vibration, as analytical calculations must be solved iteratively based on specific empirical engine parameters, making this approach ungeneralizable across multiple engines. In initial work, multiple LSTM RNN architectures were proposed, evaluated and compared. This research improves the performance of the most effective LSTM network design proposed in the previous work by using a promising neuroevolution method based on ant colony optimization (ACO) to develop and enhance the LSTM cell structure of the network. A parallelized version of the ACO neuroevolution algorithm has been developed and the evolved LSTM RNNs were compared to the previously used fixed topology. The evolved networks were trained on a large database of flight data records obtained from an airline containing flights that suffered from excessive vibration. Results were obtained using MPI (Message Passing Interface) on a high performance computing (HPC) cluster, evolving 1000 different LSTM cell structures using 168 cores over 4 days.  The new evolved LSTM cells showed an improvement of 1.35\%, reducing prediction error from 5.51\% to 4.17\% when predicting excessive engine vibrations 10 seconds in the future, while at the same time dramatically reducing the number of weights from 21,170 to 11,810.
\end{abstract}

\begin{keyword}
Ant Colony Optimization \sep ACO \sep Long Short Term Memory Recurrent Neural Network \sep LSTM \sep Recurrent Neural Network \sep RNN \sep  Time Series Prediction \sep Aviation \sep Aerospace Engineering \sep Turbomachinery \sep Turbine engine vibration \sep Flight Parameters Prediction
\MSC[2010] 00-01\sep  99-00
\end{keyword}

\end{frontmatter}


\section{Introduction}
\label{sec:introduction}

Aircraft engine vibration is of critical interest to the aviation industry, and accurate predictions of excessive engine vibration have the potential to save time, effort, money as well as human lives in the aviation industry. An aircraft engine, as turbo-machinery, should normally vibrate as it has many dynamic parts. However, it is not supposed to exceed resonance limits as to not destroy the engines~\cite{flutter&resonance}. As an example, A. V. Srinivasan~\cite{flutter&resonance} describes vibrations generated from engine blades' fluttering. Engine blades are the rotating engine components that have the largest dimensions among the other engine components. When rotating at high speeds, they will withstand high centrifugal forces that would logically give the highest contribution to engine vibrations.
Engine vibrations are not that simple to calculate or predict analytically because of the fact that various parameters contribute to their occurrence. This fact is always a problem for aviation performance monitors, especially as engines vary in design, size, operation conditions, service life span, the aircraft they are mounted on, and many other parameters. Most of these parameters’ contributions can be translated in some key parameters measured and recorded on the flight data recorder. Nonetheless, vibrations are likely to be a result of a mixture of these contributions, making it very hard to predict the real cause behind the excess in vibrations.

As such, engine vibration is a complex problem depending on an unknown number of parameters interacting over an unknown extended period of time, which poses a challenge in analyzing its causes and triggers. Holistic computation methods represent a promising solution for this problem by letting the computers find relations and anomalies that might lead to the problem through a learning process using time series data from flight data recorders (FDR). Traditional neural networks, however, lack the required capabilities to capture those relations and anomalies as they work on current time series without taking the effect of the previous time instances' parameters on the current or future time instants. Due to this, recurrent neural networks have been developed which utilize memory neurons that retain information from previous passes for use with the current experienced data, giving a chance for the neural network to know which parameter really have higher contributions to the investigated problem. 

However, these complicated neural network designs in turn posses their own challenges. Regardless of the difficulty of implementing it to a specific problem, the learning process is the main concern when dealing with such neural networks with a large number of interactive connections. When supervised learning is considered and the back-propagation is implemented to update the weights of the connections of the neural network, vanishing and exploding gradients are very serious obstacles for the successfully training recurrent neural networks. As noted by Hochrieter and Schmidhuber~\cite{hochreiter1997long}, {\it"Learning to store information over extended period of time intervals via recurrent backpropagation takes a very long time, mostly due to insufficient, decaying error back flow."} 

While this drawback hindered the application of such sophisticated neural network designs, RNNs which utilize LSTM memory cells offer a solution for this problem as the memory cells provide forget and remember gates which prevent or lessen vanishing or exploding gradients. \lstm have been used successfully in many studies on involving time series data~\cite{diartificial,lstm,felder2010wind,choi2015doctor,maknickiene2012application} and were chosen by this study to examine them as a solution to predicting aircraft engine vibration.

For many years, neural networks have strongly proven their prediction potential and applied in different many areas~\cite{yao1999evolving, siebel2009efficient, schmidhuber2015deep}. Various strategies and techniques exist to automatically generate the structure of neural networks and the most common used ones are based on evolutionary algorithms. These neuroevolution strategies examine the use of learning and evolution as concepts to improve the performance of neural networks. In traditional neuroevolution~\cite{zhang1993evolving}, an evolutionary algorithm is used to train a neural networks' connection weights with a fixed structure, but significant benefit has been demonstrated in using these techniques to both optimize and evolve connections and topologies~\cite{kohl2009learning, whiteson2005improving, floreano2008neuroevolution}, as weights are not the only key parameter for best performance of neural networks~\cite{turner2013importance}. These strategies are of particular interest as determining the optimal structure for a neural network is still an open question.  This particular work focuses on evolving the structure of LSTM neurons with  an ant colony optimization~\cite{desell2015evolving} based algorithm.

\subsection{Previous work}
\begin{figure}
\begin{center}
\includegraphics[width=.95\textwidth, height=.95\textheight]{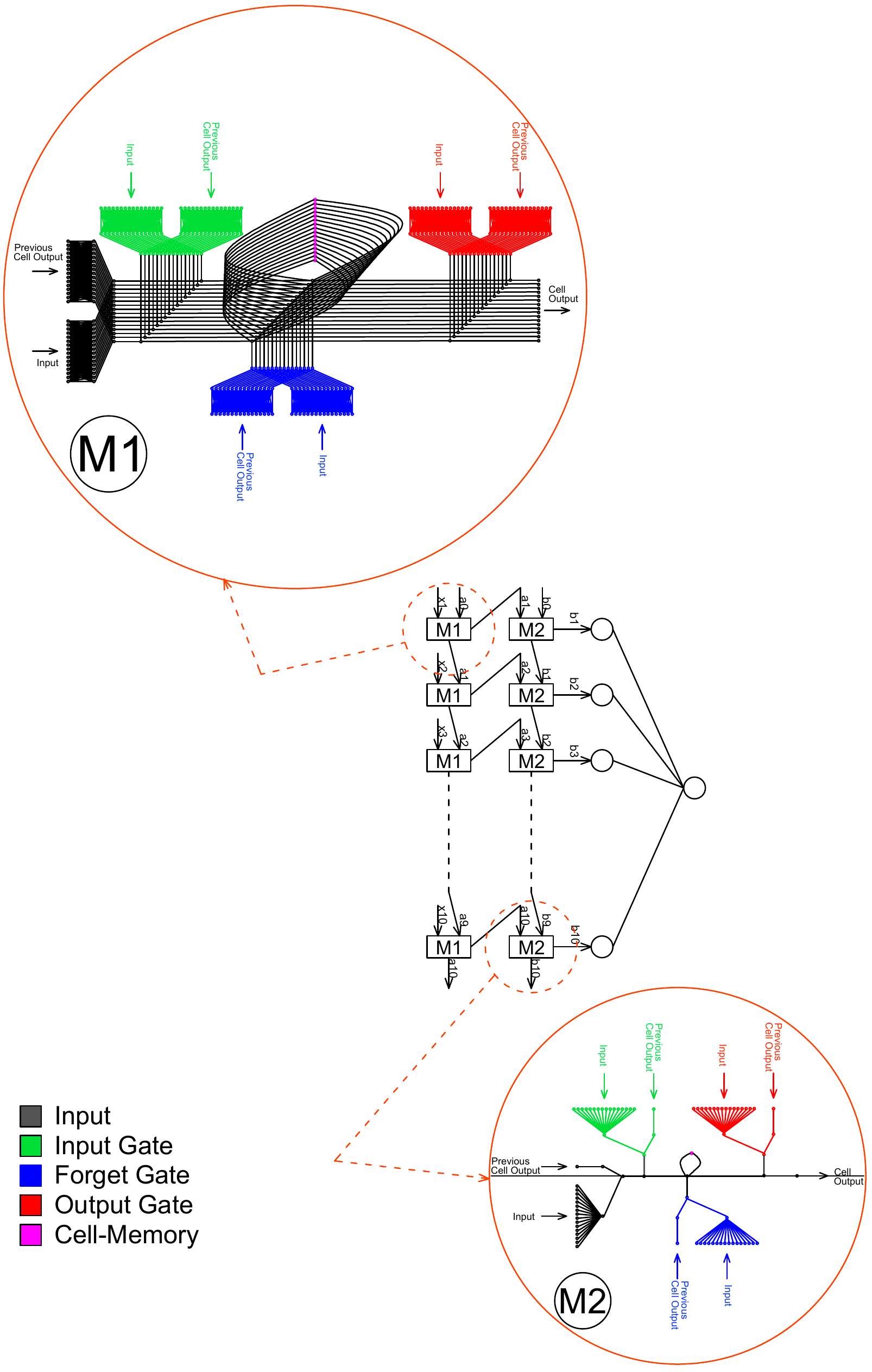}
\end{center}
\center{\caption{ \label{fig:lstm_strct} Neural network structure}}
\end{figure}

This study's ultimate goal is to explore the utilization of \lstm to predict future engine vibration in order to be used in a warning system to give indications for the problem before it occurs in order to avoid or mitigate it. An initial work examined building viable Recurrent Neural Networks (RNN) using Long Short Term Memory (LSTM) neurons to predict aircraft engine vibrations~\cite{elsaid2016using}. To achieve this, three different LSTM RNNs architectures were examined to find which would provide better results. The three architectures varied in both complexity and depth of layers. The different networks were trained on time series flight data records obtained from a regional airline containing flights that suffered from excessive vibration. The structure of the \lstm used in this study is shown in Figure~\ref{fig:lstm_strct}. After selecting an initial set of 15 relevant parameters, these LSTM RNNs were able to predict vibration values for 1, 5, 10, and 20 seconds in the future, with 2.84\% 3.3\%, 5.51\% and 10.19\% mean absolute error, respectively.

\subsection{Ant Colony Optimization}
Ant colony optimization (ACO) is a metaheuristic used to find approximate solutions to many combinatorial problems. It belongs to a family of bio inspired metaheuristics. ACO is a distributed approach using agents called artificial ants. These artificial ants resemble biological ants, in that each ant is independent and communicates with other members of the colony through a chemical called pheromone. Ants randomly explore areas, however they tend to follow paths with pheromones, and upon finding food they mark their return path with more pheromone. Pheromones decay over time, and paths with the most pheromone represent the most promising paths to food. The first algorithm of this type (the ``Ant System''~\cite{dorigo1996ant}) was designed for the traveling salesman problem, but failed to produce competitive results. However, subsequent research has shown this algorithm to be effective on this problem~\cite{dorigo1997ant,bianchi2002ant,manfrin2006parallel} and interest for the metaphor has launched many algorithms inspired by it in various fields, including continuous optimization problems~\cite{34, 36, 35, 5, 27, dreo2002new}, and even training neural networks~\cite{R.Ashena, blum2005training, J.B.Li, Unal}.

\subsection{Study's Contribution}
This work improves the performance of the LSTM network architectures proposed in the previous work by optimizing LSTM cell structure. Since the contributions of the input parameters are not of the same magnitude to the problem (vibration), the weights of the neural network are adjusted through backpropagation. However, a fully connected neural network can pose additional training complexity and unwanted noise through some of the connections coming out of some of the inputs. Therefore, there is a need for a way to examine these connections and try to eliminate those which contribute to the prediction error.  ACO has been chosen mainly because it has proven its effectiveness in evolving RNNs~\cite{desell2015evolving} for time series data prediction. The results have demonstrated that the evolved LSTM architecture increased the best previous results' performance by 1.35\% in predicting vibration 10 seconds in the future, while at the same time only requiring nearly half of the connections (the number of weights was reduced from 21,170 to 11,810). The prediction accuracy was improved from 94.49\% to 95.83\% based on mean absolute error calculations.

\section{Related Work}
\label{sec:related_work}

\subsection{Turbine Engine Vibration}
According to A. V. Srinivasan ~\cite{flutter&resonance}: {\it``The most common types of vibration problems that concern the designer of jet engines include {\it (a)} resonant vibration occurring at an integral order, \ie ~multiple of rotation speed, and {\it (b)} flutter, an aeroelastic instability occurring generally as a nonintegral order vibration, having the potential to escalate, \underline{unless checked by any means available to the operator}, into larger and larger stresses resulting in serious damage to the machine. The associated failures of engine blades are referred to as high cycle fatigue failures"}. The {\it means available to the operator} in practical aviation operations are mainly the implementation of manufacturers' maintenance program which relies on reliability observations.

Any aircraft maintenance program has four objectives: {\it i)} guaranteeing the inherit safety and reliability levels of the systems and subsystems, {\it ii)} restore those levels to their original levels if deviations occur, {\it iii)} gather information about design enhancement for systems and subsystems that showed deficiency in there expected reliability, and {\it iv)} accomplish these targets at the the most overall cost efficiency possible. To achieve these goals a maintenance program consists of checks performed over specific intervals to perform on-condition checks which might be visual inspections, test runs, or non-destructive tests performed on the systems' or subsystems components. In addition to that, and as mentioned earlier, preventive maintenance is also part of the maintenance program where parts are replaced or overhauled based on reliability observations~\cite{moubray1997reliability}. 

There are also other methods to mitigate the risk coming from excessive engine vibration using statistical and holistic computation methods. This is accomplished by monitoring the engine performance through its flight data history, which is logged as time series data, in order to forecast the excessive vibration occurrence. However, since these methods are not exact, reasonable safety factors should be considered. As discussed in the following sections, machine learning and artificial intelligence becomes the heart of such methods.

\subsection{Time Series Data Prediction}
From a statistical point of view, the main goal of prediction is to provide vital information for decision makers, economists, planners optimizers, industrialists and critical systems operators. There are two sides for prediction: the qualitative side and the quantitative side. The qualitative side utilizes methods known as the judgmental or subjective prediction methods which covers methods relaying on intuition, judgement or opinions of some kind of a referee as consumers, experts and/or supporting information. Qualitative methods are considered in cases when past data is not available. On the other hand, quantitative methods include univariate and multivariate methods. For many study cases related to different scientific and real life problems, the time series data are available on several dependent variables, and in such cases multivariate prediction methods are used~\cite{chatfield2016analysis}.

The models in the time series predictions realm mainly falls in two categories: {\it a)} statistical prediction models which include, \eg, the autoregressive (AR) model, the moving average (MA) model, and hybrid models that derive from them such as autoregressive moving average (ARMA), autoregressive integrated moving average (ARIMA), seasonal ARMIA (SARIMA)~\cite{boukary2016comparison}, vector autoregressive (VAR) models and {\it b)} artificial neural networks (ANN) prediction models. While successful studies have managed to achieve good results by using such methods and even reported results that out perform neural networks~\cite{goelmultivariate}, the main draw back of the statistical methods is that they generally can not be applied to non-linear systems~\cite{boukary2016comparison}. On the other hand, ANNs have shown good performance with such systems. There are also a third category which are the hybrid models. These models (hybrid) offer the benefits of both statistical and ANN models. Consequently, as the studied system involving prediction of aircraft engine vibration is complicated in its nature and is expected to be extremely non-linear, statistical and hybrid models are considered beyond the scope of the study.

\subsection{Time Series Prediction in Aviation}
Some effort has been done using neural networks to classify engine abnormalities without doing analytical computation, \eg, Alexandre Nairac \etal ~\cite{nairac1999system} have performed research to detect abnormalities in engine vibrations based on recorded data. To achieve that, the work used two modules. One of the modules uses the overall shape of the vibration curve to detect unusual vibration signatures. The second one reports sudden unexpected transitions in the signature curves. Their approach to detect defects is not to introduce examples of faulty engines to the neural network, rather, only examples of healthy engines are introduced to the neural networks in the training phase. This approach was taken to overcome the lack of existence of adequate faulty engine data, which was not enough for training. In this context, the paper introduces the term `normality' to describe the behavior of normal engines and `abnormality' to describe the behavior of faulty engines. Using statistical models, the faulty engines detection would be described as `novelty' detection based on the deviation from the data distribution. The best results this work achieved was the prediction of faulty engines with 84\% successful classifications.  

David A. Clifton \etal ~\cite{clifton2007framework} presented work for predicting abnormalities in engine vibration based on statistical analysis of vibration signatures. The paper presents two modes of prediction. One is ground-based (off-line), where prediction is done by run-by-run analysis to predict abnormalities based on previous engine runs. The success in this approach was predicting abnormalities two flights ahead. The other mode is a flight based-mode (online) in which detection is done either by sending reduced data to the ground-base or processing it onboard the aircraft. The paper mentions that they could successfully predict vibration events 2.5 hours in the future. However, this prediction is done after half an hour of flight data collection, which might be a critical time as well, as excess vibration may occur during this data collection time. The paper did not mention how much data was required to have a sound prediction.

\subsubsection{RNN for Predicting Flight Parameters}
Having an advantage over standard FFNNs\footnote{Feed Forward Neural Networks}, RNNs can deal with sequential input data, using their internal memory to process sequences of inputs and use previously stored information to aid in future predictions. This is done by feedback connections or by looping between neurons, which allows them to be of predicting more complex data~\cite{gers2002learning}.

This presented work is in part inspired by previous work on predicting flight parameters~\cite{desell-ppsn-2014,desell2015evolving}. Which first utilized evolutionary algorithms such as particle swarm optimization~\cite{kennedy2011particle,poli2007particle} and differential evolution~\cite{storn1997differential} to optimize the weights of the network~\cite{desell-ppsn-2014}, and then an ant colony optimization based algorithm to evolve different recurrent neural network structures~\cite{desell2015evolving}. The neural networks evolved with ant colony optimization predicted airspeed, altitude and, pitch with a 63\%, 97\% and 120\% improvement respectively over the previously best published results.
The research used recurrent neural networks and applied an ant-colony optimization (ACO) algorithm ~\cite{ant_colony_alg1, ant_colony_alg2, ant_colony_alg3}, an optimization technique used in the beginning on discrete problems, mainly on the Traveling Salesman Problem~\cite{18}. Later it was used in continuous optimization problems~\cite{34, 36, 35, 5, 27, dreo2002new}, including training neural networks~\cite{R.Ashena, blum2005training, J.B.Li, Unal}.

\subsubsection{LSTM RNN}
LSTM RNNs were first introduced by S. Hochrieter \& J. Schmidhuber ~\cite{lstm}. While the work by T. Desell \etal ~utilized non-gradient based evolutionary algorithms to optimize RNN weights, LSTM neurons provide a solution for the exploding/vanishing gradients problem by utilizing various gates, which allow backpropagation to be used in large RNNs (S. Hochrieter in 1991). This work has paved the way for many interesting projects.

Later, J. Schmidhuber \etal~\cite{Learning_to_Forget} emphasized the forget gate in the LSTM RNNs. The paper mentions that {\it``We identify a weakness of LSTM networks processing continual input streams that are not a priori segmented into subsequences with explicitly marked ends at which the network's internal state could be reset. Without resets, the state may grow indefinitely and eventually cause the network to break down. Our remedy is a novel, adaptive “forget gate” that enables an LSTM cell to learn to reset itself at appropriate times, thus releasing internal resources. We review illustrative benchmark problems on which standard LSTM outperforms other RNN algorithms. All algorithms (including LSTM) fail to solve continual versions of these problems. LSTM with forget gates, however, easily solves them, and in an elegant way.''} However, Felix A. Gers \etal~\cite{Gers2001} suggest that {\it``LSTM RNNs does not carry over to certain simpler time series prediction tasks solvable by time window approaches''}. The paper suggests to use LSTM when {\it``simpler traditional approaches fails''}. 

LSTM RNNs have been used with strong performance in image recognition ~\cite{Donahue_2015_CVPR}, audio visual emotion recognition ~\cite{arXiv:1603.08321v1}, music composition ~\cite{eck2002first} and other areas. Regarding time series prediction, for example, LSTM RNNs have been used for stock market forecasting~\cite{diartificial} and forex market forecasting~\cite{maknickiene2012application}. Also forecasting wind speeds~\cite{lstm, felder2010wind} for wind energy mills, and even predicting diagnoses for patients based on health records~\cite{choi2015doctor}.

\subsection{Evolutionary Optimization Methods}
Several methods for evolving topologies along with weights have been searched and deployed. In~\cite{stanley2002evolving, stanley2002evolving}, NeuroEvolution of Augmenting Topologies (NEAT) has been developed. It is a genetic algorithm that evolves increasingly complex neural network topologies, while at the same time evolving the connection weights. Genes are tracked using historical markings with innovation numbers to perform crossover among different structures and enable efficient recombination. Innovation is protected through speciation and the population initially starts small without hidden layers and gradually grows through generations~\cite{annunziato2002adaptive, larochelle2009exploring, kandel2000principles}. Experimentations have demonstrated that NEAT presents an efficient way for evolving neural networks for weights and topologies in parallel or separately. Its power resides in its ability to combine all the four main aspects discussed above and expand to complex solutions along the generation process. However NEAT still has some limitations when it comes evolving neural networks with weights or LSTM cells for time series prediction tasks as it has been claimed in~\cite{desell2015evolving}.

\section{Methodology}
\label{ch:methodology}

This work utilizes the ACO method developed by Desell \etal ~to evolve the structure of LSTM cells, in part due to strong previous results and because it allows any method to be used to determine the optimal weights of connections. This is particularly important as it allows backpropagation to be used on a large scale LSTM RNN, which is significantly more efficient than the non-gradient based evolutionary algorithms used in previous work.

\subsection{Experimental Data}
\label{sec:test_data}

The flight data used consists of 76 different parameters recorded on the aircraft Flight Data Recorder (FDR), inclusive of the engine vibration parameters. During the data processing phase of the project, two efforts were done to identify the parameters that most contributed to the engine vibration. 

\subsubsection{Data Correlation Parameter Selection}
\label{sec:cross_corr}
Primarily, cross-correlation analysis~\cite{lewis1995fast} was exercised to find the potential parameters that highly contribute to vibration. Every parameter from each flight was cross-correlated to vibration then plotted to pick the highest correlated parameters. Cross correlation was calculated using the following Equation:

\begin{equation}
    Cross\_Correlation = \sum_{a = -\infty}^{\infty} x[a] \cdot vib[a]
    \label{eq:cross_correlation}
\end{equation}

Where the highest correlation was determined by calculating the area under the plotted curve for the normalized data. The top correlated parameters to vibration were:

\begin{enumerate}
\item Right InBoard Spoiler
\item Right OutBoard Spoiler
\item Left InBoard Spoiler
\item Left OutBoard Spoiler
\item Static Air Temperature
\item Pitch 2
\item Pitch
\item Slat Configuration
\item Main Landing Gear Lock Down Sensors
\item Flap Configuration
\end{enumerate}

\begin{figure}
\begin{center}
\includegraphics[height=.20\textheight]{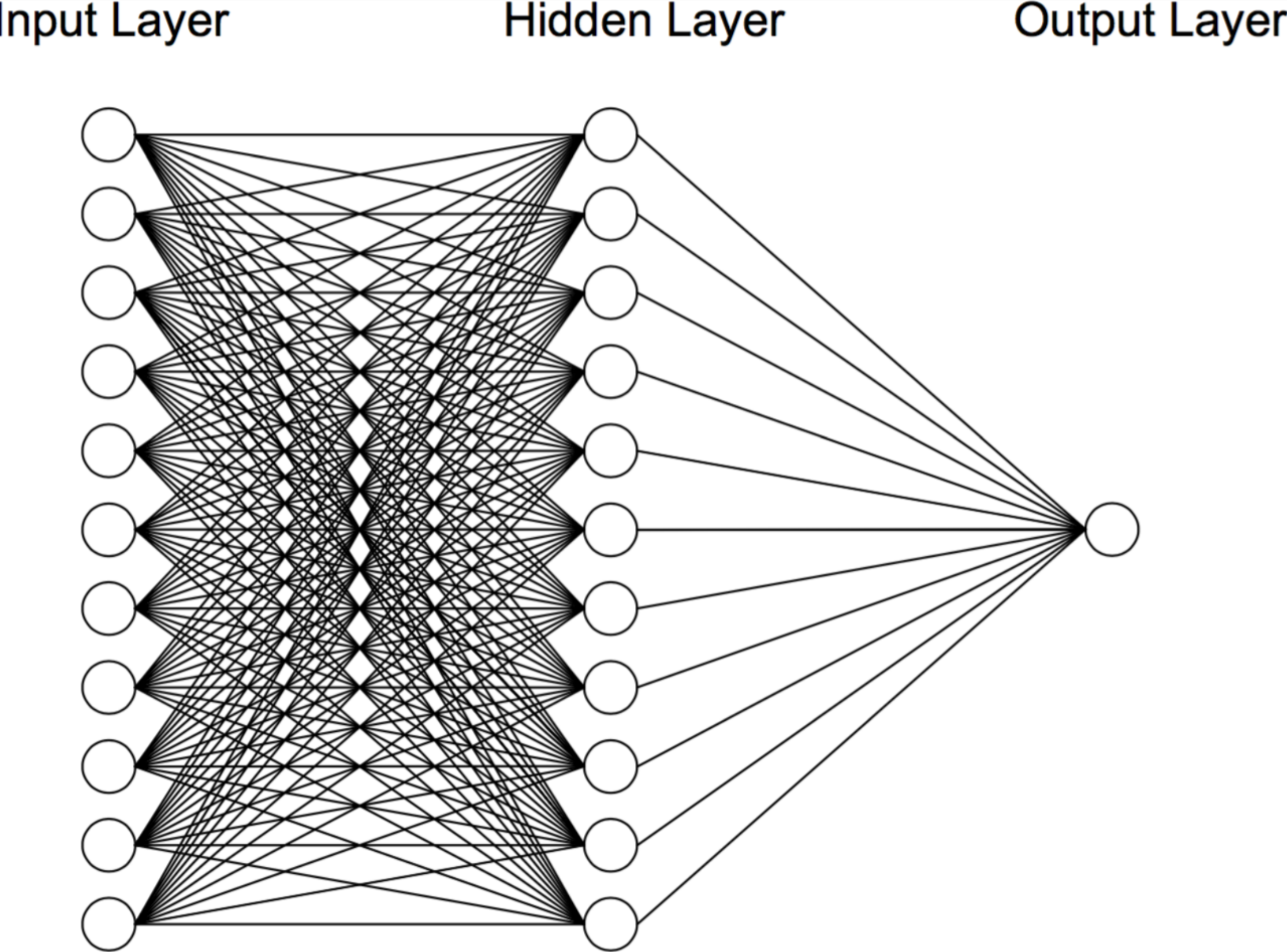}
\end{center}
\caption{ \label{fig:vanilla_nn} A one layer feed forward neural network.}
\end{figure}

A one layer feed forward neural network was built as shown in Figure~\ref{fig:vanilla_nn} to predict vibration given other parameters within the same second. However, the results were poor, with significant noise in the predictions. This imposed a question about the quality of the chosen parameters using this method and due to this, another method of parameter-selection was sought. A potential cause for such misleading cross-correlation chosen parameters was that some flight configuration parameters like spoilers/slats/flaps positions, pitch angle and main-landing-gear position do not change but few times during the flight, which can translate into high correlation with the vibration.

\subsubsection{Aerodynamics/Turbo-machinery Parameter Selection}
\label{subsec:aero_select}

A second subset of the FDR parameters were then chosen based on the likelihood of their contribution to the vibration based on aerodynamics/turbo-machinery expert knowledge. Again, a one layer feed forward neural network with a structure similar to the one shown in Figure~\ref{fig:vanilla_nn}, except the number of input and hidden nodes was equal to the number of chosen parameters, was applied and these results were encouraging enough to use these parameters for predicting vibration in future.

Some parameters, such as Inlet Guide Vans Configuration, Fuel Flow, Spoilers Configuration (this was preliminarily considered because of the special position of the engine mount), High Pressure Valve Configuration and Static Air Temperature were excluded because it was found that they generated more noise than positively contributing to the vibration prediction.

The final chosen parameters were:

\begin{enumerate}
\item Altitude [ALT]
\item Angle of Attack [AOA]
\item Bleed Pressure [BPRS]
\item Turbine Inlet Temperature [TIT]
\item Mach Number [M]
\item Primary Rotor/Shaft Rotation Speed [N1]
\item Secondary Rotor/Shaft Rotation Speed [N2]
\item Engine Oil pressure [EOP]
\item Engine Oil Quantity [EOQ]
\item Engine Oil Temperature [EOT]
\item Aircraft Roll [Roll]
\item Total Air Temperature [TAT]
\item Wind Direction [WDir]
\item Wind Speed [WSpd]
\item Engine Vibration [Vib]
\end{enumerate}

\subsection{Recurrent Neural Network Design}
\label{sec:methodology}

\begin{figure*}
\begin{center}
\includegraphics[width=0.95\textwidth]{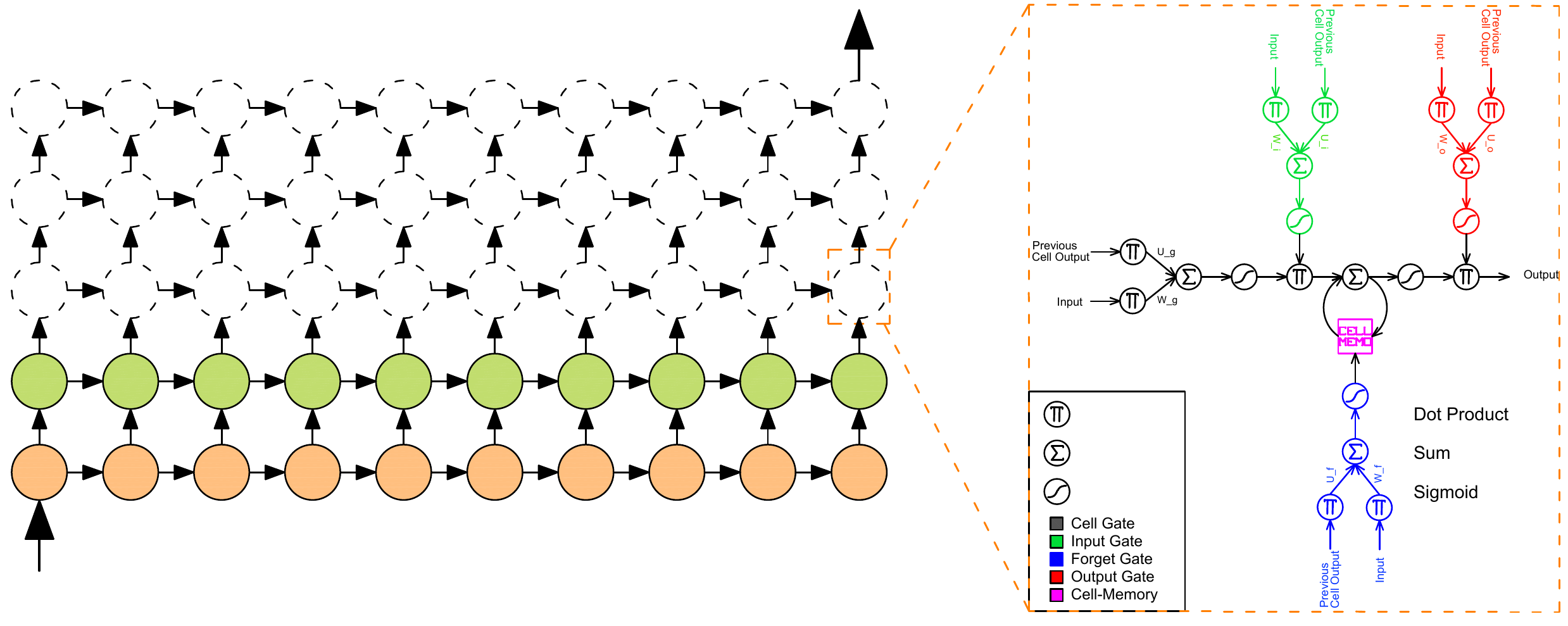}
\end{center}
\caption{ \label{fig:LSTM_cell_enlarge} A generic overview of the design of a LSTM RNN.}
\end{figure*}

Three LSTM RNN architectures were designed to predict engine vibration 5 seconds, 10 seconds, and 20 seconds in the future. Each of the 15 selected FDR parameters is represented by a node in the inputs of the neural network and an additional node is used for a bias. Each neural network in the three designs consists of LSTM cells that receive both an initial input of flight data at some time in the past or the output from a cell in the lower layer, and the output of the previous cell in the same layer, as inputs (see Figure~\ref{fig:LSTM_cell_enlarge}). Each cell has three gates to control the flow of information through the cell and accordingly, the output of the cell. Each cell also has a cell-memory which is the core of the LSTM RNN design. The cell-memory allows the flow of information from the previous states into the current predictions.

\begin{figure}
\begin{center}
\includegraphics[width=0.48\textwidth]{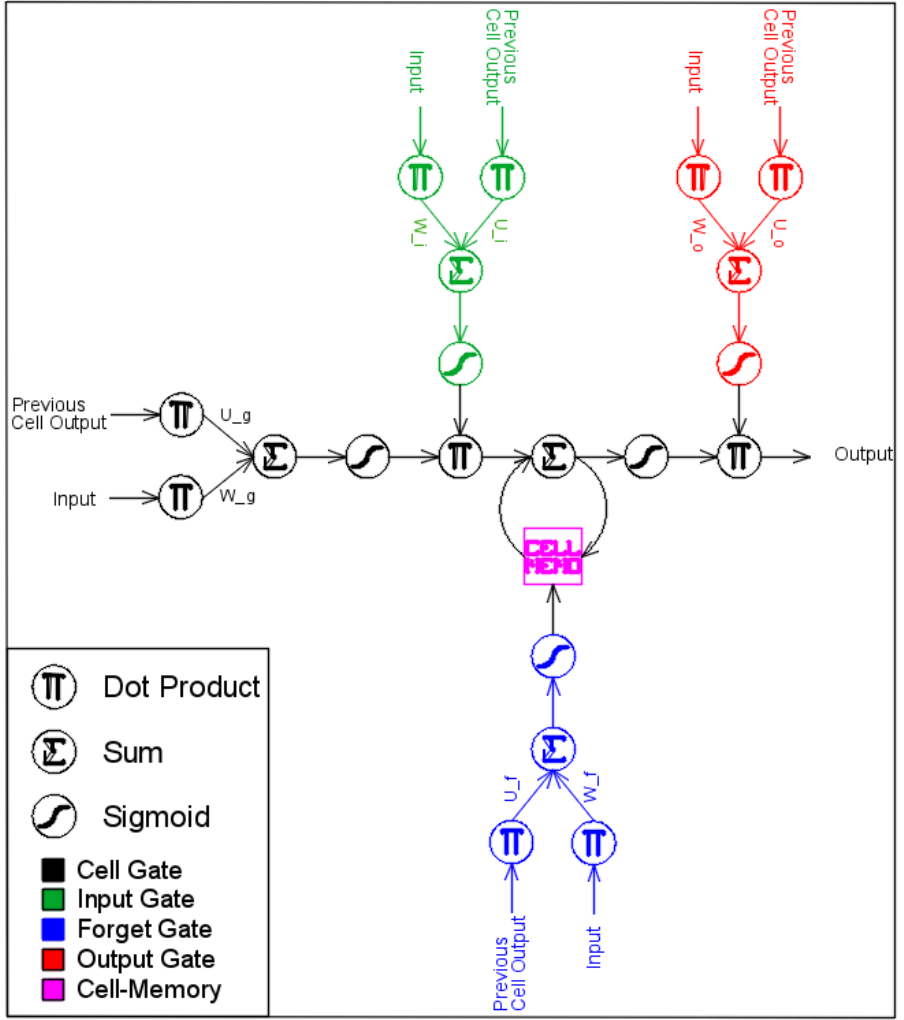}
\end{center}
\caption{ \label{fig:LSTM_cell} LSTM cell design}
\end{figure}

The gates that control the flow are shown in Figure~\ref{fig:LSTM_cell}. They are: {\it i)} the \emph{input gate}, which controls how much information will flow from the inputs of the cell, {\it ii)} the \emph{forget gate}, which controls how much information will flow from the cell-memory, and {\it iii)} the \emph{output gate}, which controls how much information will flow out of the cell. This design allows the network to learn not only about the target values, but also about how to tune its controls to reach the target values.

\begin{figure}
\begin{center}
\includegraphics[width=.98\textwidth, height=.3\textheight]{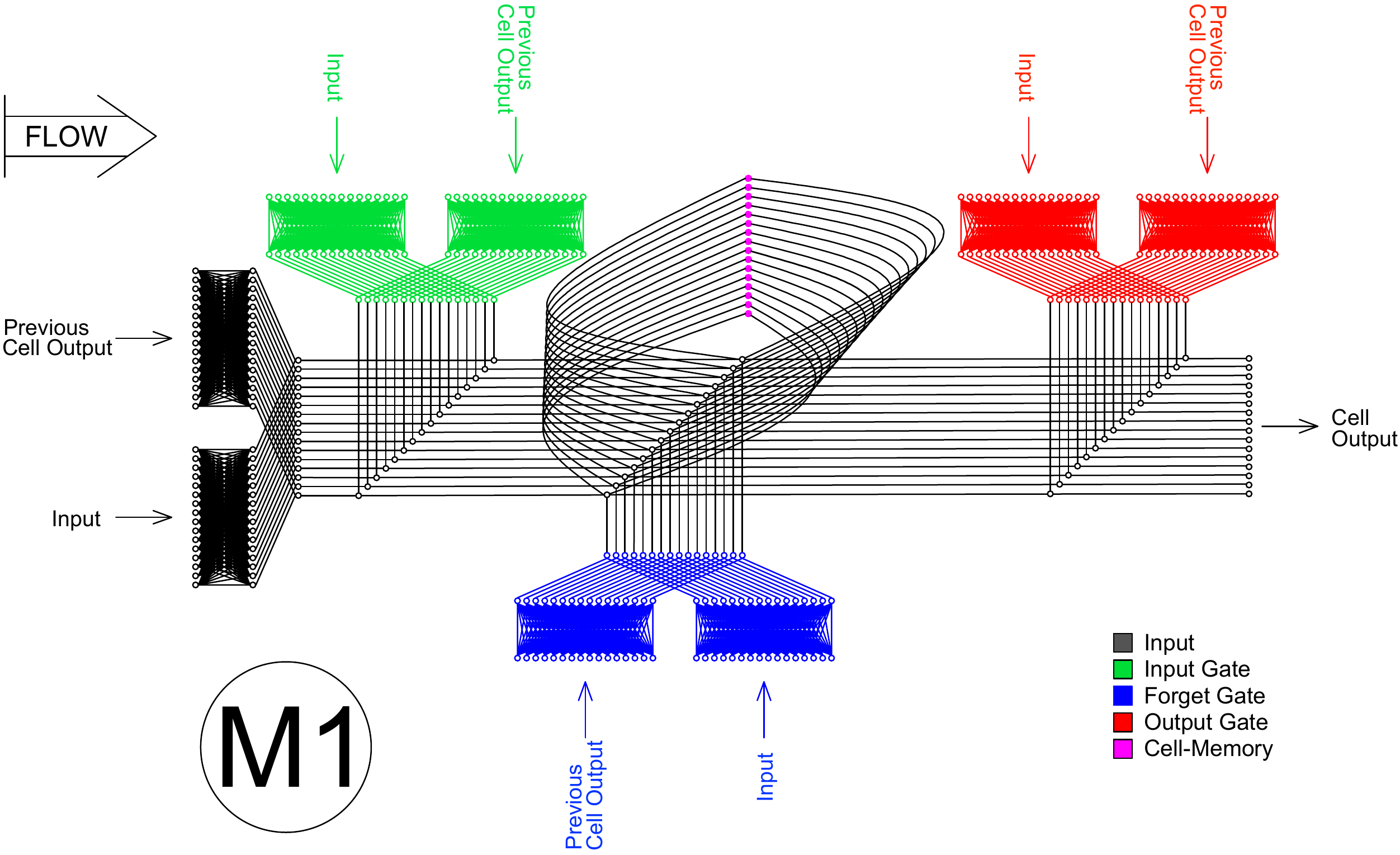}
\end{center}
\caption{ \label{fig:M1} Level 1 LSTM cell design}
\end{figure}

\begin{figure}
\begin{center}
\includegraphics[width=.98\textwidth, height=.3\textheight]{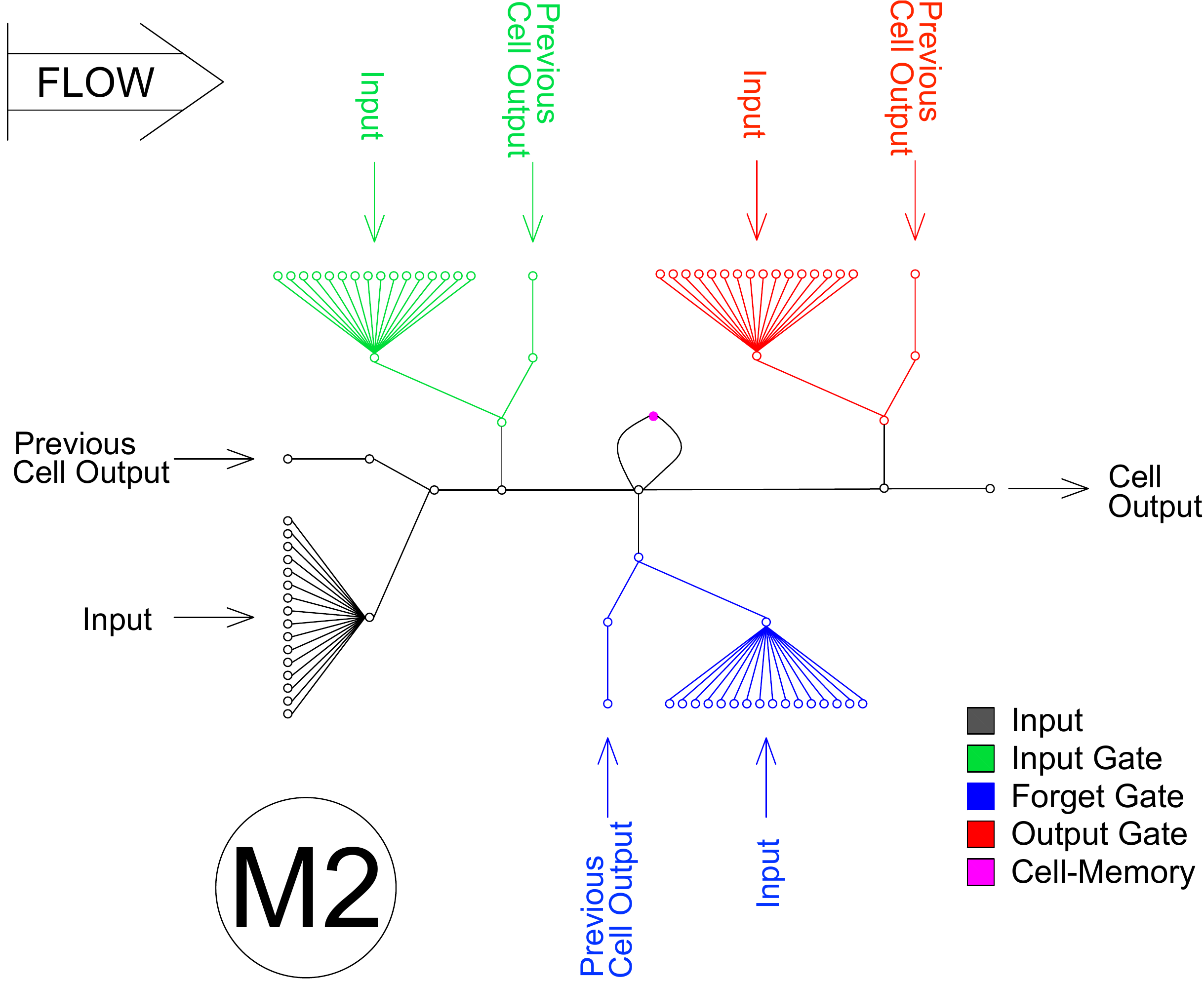}
\end{center}
\caption{ \label{fig:M2} Level 2 LSTM cell design}
\end{figure}

All the utilized architectures follow the common LSTM RNN designs shown in Figure~\ref{fig:LSTM_cell_enlarge} and~\ref{fig:LSTM_cell}. However, there are two variations of this common design used in the utilized architectures, shown in Figures~\ref{fig:M1} and~\ref{fig:M2}, with the difference being the number of inputs from the previous cell. Cells that take an initial number of inputs and output the same number of outputs are denoted by {\it`M1'} cells. As input nodes are needed to be reduced through the neural network, the design of the cells are different. Cells which perform a reduction on the inputs are denoted by {\it`M2'} cells.

\subsubsection{LSTM RNN Forward Propagation Equations}

The equations used in the forward propagation through the neural network are:

{
\begin{equation}
    i_t = Sigmoid(w_i \bullet x_t + u_i \bullet  a_{t-1} + bais_i)
    \label{eq:input-gate}
\end{equation}

\begin{equation}
    f_t = Sigmoid(w_f \bullet x_t + u_f \bullet  a_{t-1} + bais_f)
    \label{eq:forget-gate}
\end{equation}

\begin{equation}
    o_t = Sigmoid(w_o \bullet x_t + u_o \bullet  a_{t-1} + bais_o)
    \label{eq:output-gate}
\end{equation}

\begin{equation}
    g_t = Sigmoid(w_g \bullet x_t + u_g \bullet  a_{t-1} + bais_g)
    \label{eq:gate}
\end{equation}

\begin{equation}
    c_t = f_t \bullet c_{t-1} + i_t \bullet g_t
    \label{eq:cell_memo}
\end{equation}

\begin{equation}
    a_t = o_t \bullet Sigmoid(c_t)
    \label{eq:out}
\end{equation}
}
where (see Figure~\ref{fig:LSTM_cell}):

\hspace{7mm $i_t$: input-gate output}

\hspace{7mm $f_t$: forget-gate output}

\hspace{7mm $o_t$: output-gate output}

\hspace{7mm $g_t$: input's sigmoid}

\hspace{7mm $c_t$: cell-memory output}

\hspace{7mm $w_i$: weights associated with input and input-gate}

\hspace{7mm $u_i$: weights associated with previous output and input-gate}

\hspace{7mm $w_f$: weights associated with input and forget-gate}

\hspace{7mm $u_f$: weights associated with previous output and forget-gate}

\hspace{7mm $w_o$: weights associated with input and output-gate}

\hspace{7mm $u_o$: weights associated with previous output and the output-gate}

\hspace{7mm $w_g$: weights associated with the cell input}

\hspace{7mm $u_g$: weights associated with previous output and the cell input}

\vspace{5mm}

and the formula of the sigmoid function is:
\begin{equation}
	Sigmoid(\alpha)= \frac{1}{1+e^{-\alpha}}
\end{equation}

\subsection{LSTM RNN Architectures}
The three architectures are as follows, with the dimensions of the weights of these architectures shown in Table~\ref{table:wghts} and the total number of weights shown in Table~\ref{table:wghts_elmts}.

\begin{figure}
\centering
\subfloat[Architecture I]{\includegraphics[width=0.95\textwidth, height=.30\textheight]{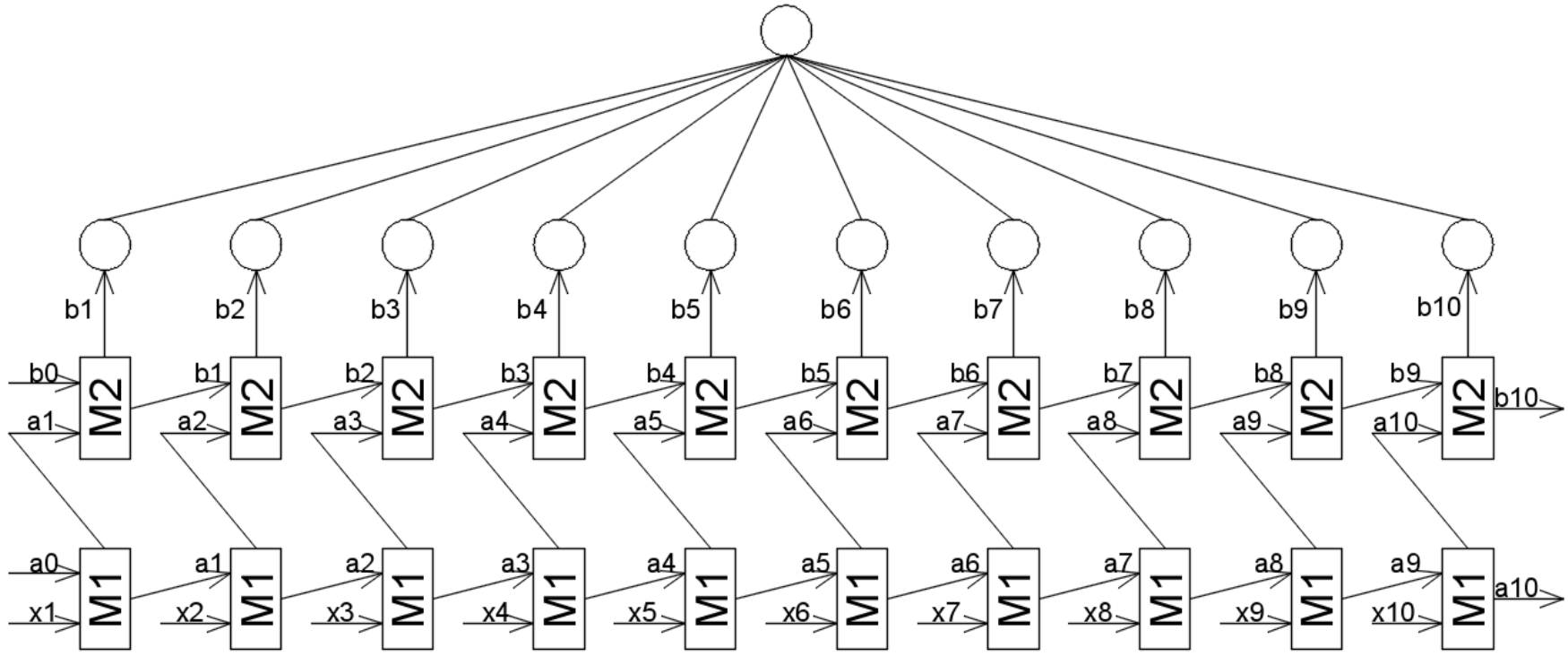}
\label{fig:art_1}}
\\
\subfloat[Architecture II]{\includegraphics[width=0.95\textwidth, height=.30\textheight]{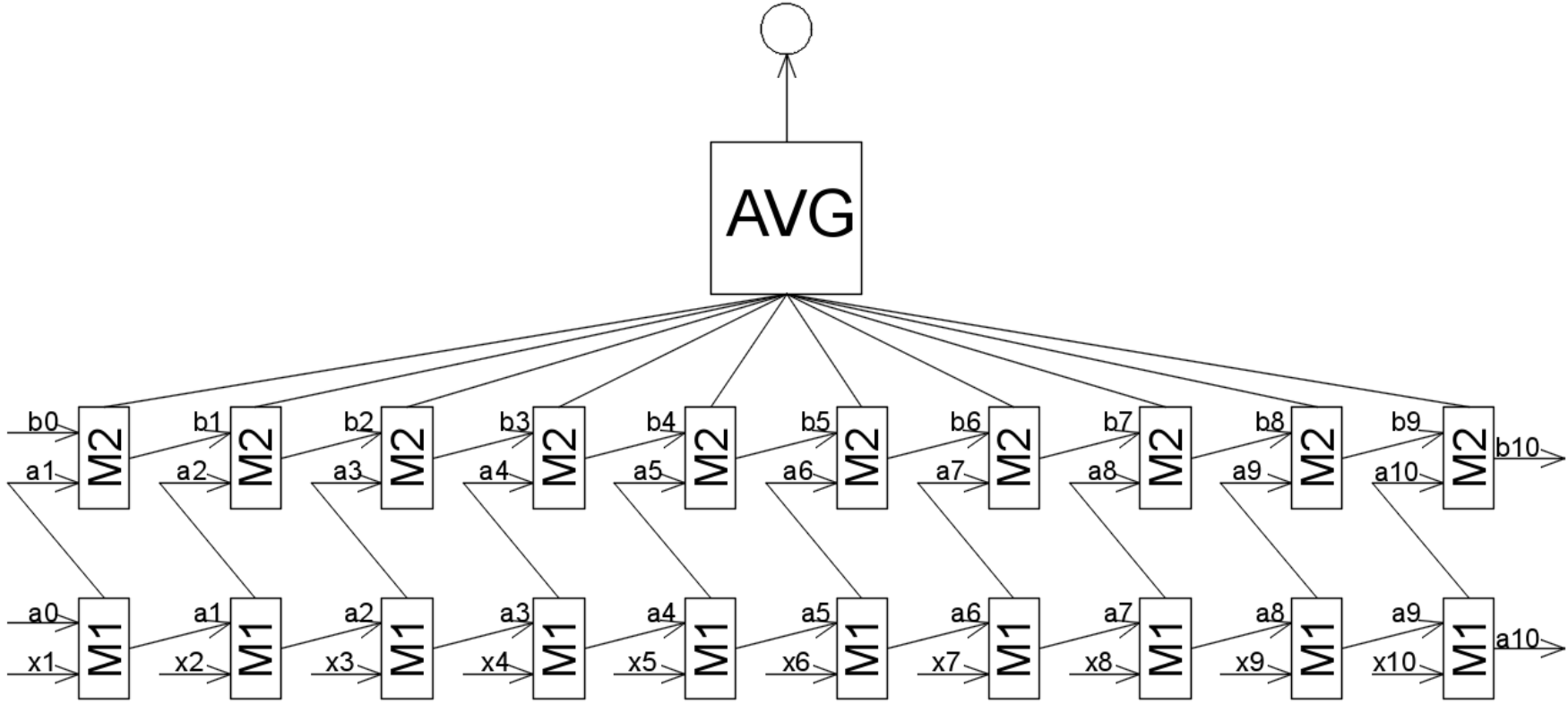}
\label{fig:art_2}}
\\
\subfloat[Architecture III]{\includegraphics[width=0.95\textwidth, height=.30\textheight]{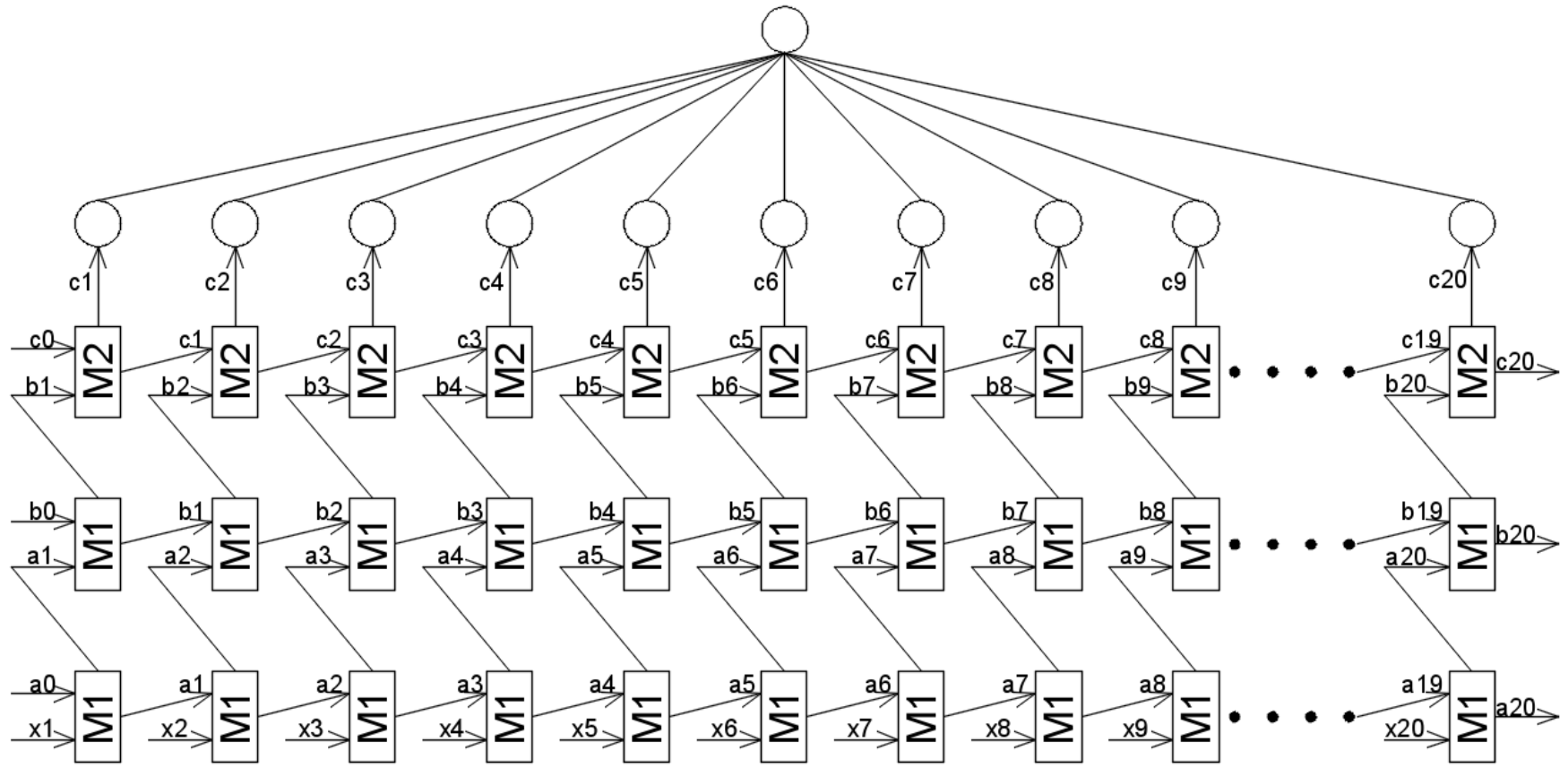}
\label{fig:art_3}}
\caption{Used LSTM RNNs Architectures}
\label{fig_arts}
\end{figure}

\vspace{.3cm}
\noindent
\textbf{Architecture I} \\
\label{sec:arci}
As shown in Figure~\ref{fig:art_1}, the first level of the architecture takes inputs from ten time series (the current time instant and the past nine). It then feeds the second level of the neural network with the output of the first level. The output of the first level of the neural network is considered the first hidden layer. The second level of the neural network then reduces the number of nodes fed to it from 16 nodes (15 input nodes + bias) per cell to only one node per cell. The output of the second level of the neural network is considered the second hidden layer. Finally, the output of the second level of the neural network would be only 10 nodes, a node from each cell. These nodes are fed to a final neuron in the third level to compute the output of the whole network. 

\begin{figure*}
\begin{center}
\includegraphics[width=\textwidth, height=.6 \textheight]{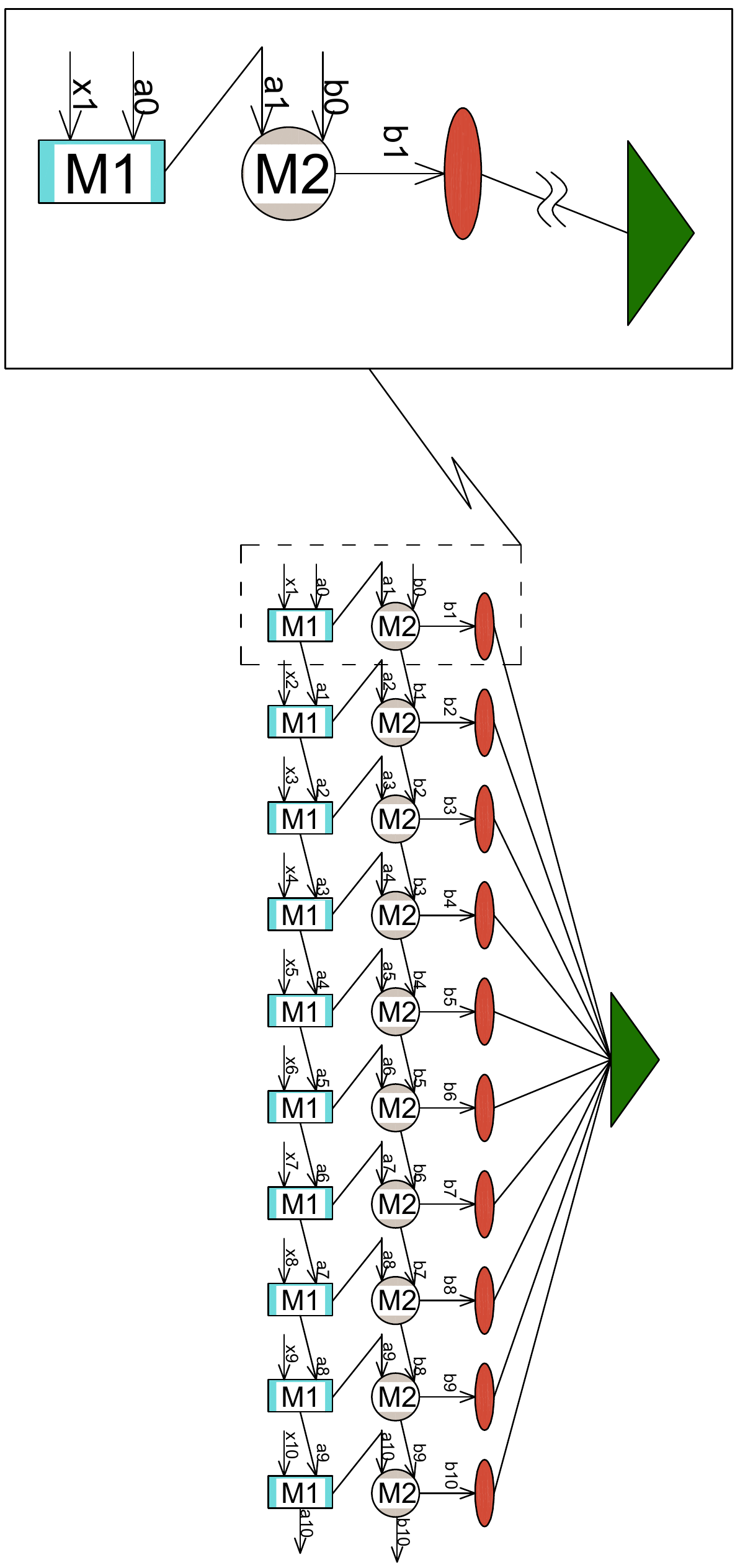}
\end{center}
\center{\caption{\label{fig:art1_one_step} One time step of Architecture I}}
\end{figure*}

\begin{figure*}
\begin{center}
\includegraphics[width=\textwidth, height=.97 \textheight]{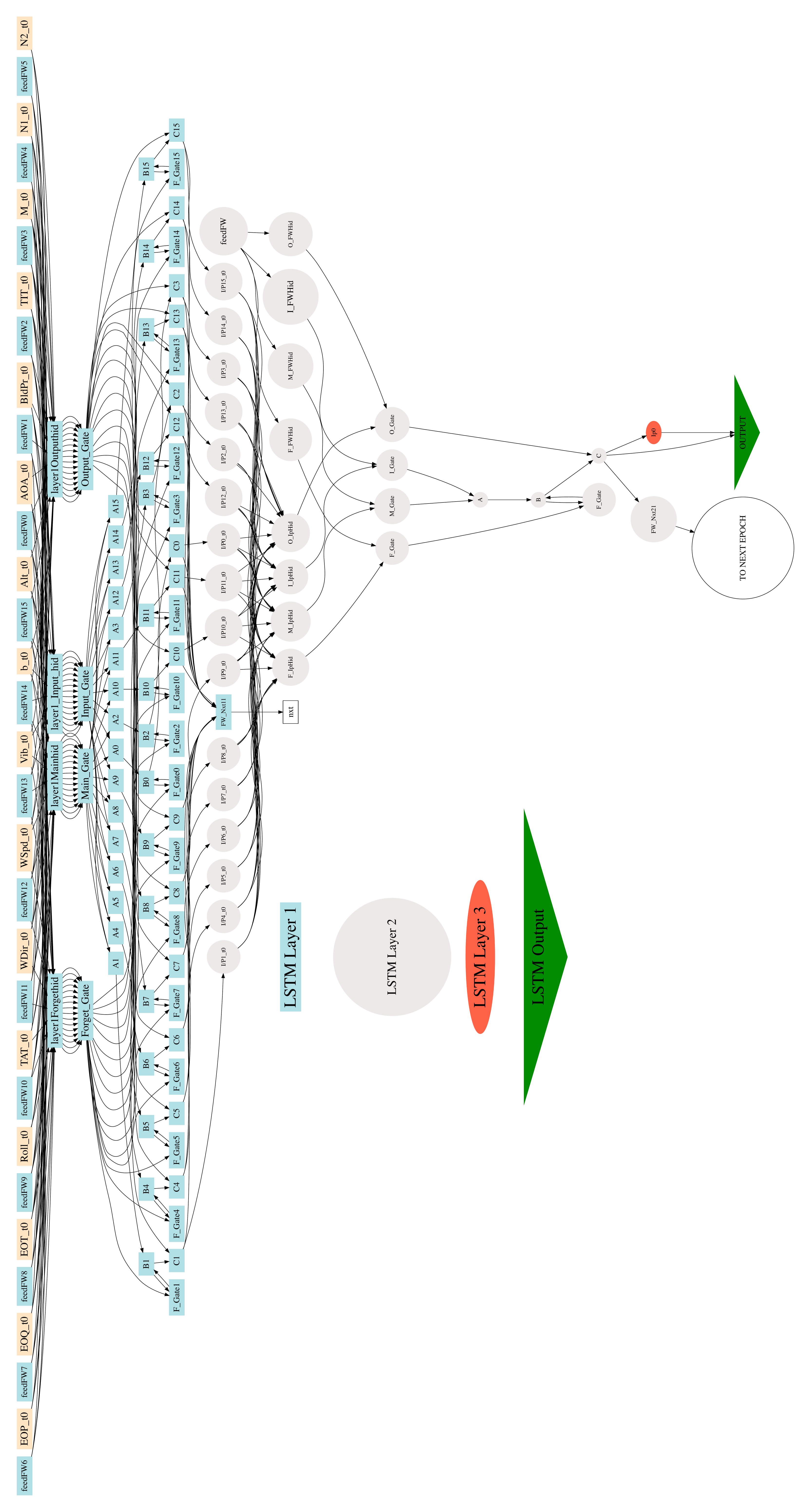}
\end{center}
\center{\caption{\label{fig:art1_full} One time step of the Architecure I Full Structure}}
\end{figure*}

The dimensions of the weights matrices and vectors of this architecture are shown in Table~\ref{table:wghts}. The total number of weights are is shown in Table~\ref{table:wghts_elmts}. Figures~\ref{fig:art1_one_step} and~\ref{fig:art1_full} provide an overview of architecture I, as it has a large number of connections (21,170). Figure~\ref{fig:art1_one_step} shows the overall design of how the LSTM cells are connected, and then Figure~\ref{fig:art1_full} displays all the connections within a single time step of the full LSTM RNN. As a whole, there are 10 different instances of Figure~\ref{fig:art1_one_step}, each connected as specified in Figure~\ref{fig:art1_full}. 

\vspace{.3cm}
\noindent
\textbf{Architecture II}\\
\label{sec:arcii}
As shown in Figure~\ref{fig:art_2}, this architecture is almost the same as the previous one except that it does not have the third level. Instead, the output of the second level is averaged to compute the output of the whole network.

The dimensions of the weights matrices and vectors of this architecture are shown in Table~\ref{table:wghts}. The total number of weights are shown in Table~\ref{table:wghts_elmts}.

\vspace{.3cm}
\noindent
\textbf{Architecture III}\\
\label{sec:arciii}
Figure~\ref{fig:art_3} presents a deeper neural network architecture. In this design, the neural network takes inputs from twenty time series (the current time instant and the past nineteen) as the first level. It feeds the second level of the neural network with the output from the first level. The second level does the same procedure as first level giving a chance for more abstract decision making. The output of the second level of the neural network is considered the first hidden layer and the output of the second level is considered the second hidden layer. The third level of the neural network then reduces the number of nodes fed to it from 16 nodes (15 input nodes + bias) per cell to only one node per cell. The output of the third level of the neural network is considered the third hidden layer. Finally, the output of the third level of the neural network is twenty nodes, a node from each cell. These nodes are fed to a final neuron in the fourth level to compute the output of the whole network.

The Dimensions of the weights matrices and vectors of this architecture are shown in Table~\ref{table:wghts}. The total number of weights are shown in Table~\ref{table:wghts_elmts}.

\begin{table}
\centering
\caption{\small Architectures Weights-Matrices Dimensions}
\label{table:wghts}
\resizebox{0.48\textwidth}{!}
{%
\begin{tabular}{lcccccllcccllccccccllcccc}
	\toprule
\\&

 \multicolumn{8}{c}{\bf \small Architecture I}         \\
                             & $w_i$ & $u_i$ & $w_f$ & $u_f$ & $w_o$ & $u_o$ & $w_g$ & $u_g$ \\
{\bf \small Level 1} & {\tiny 16$\times$16} & {\tiny 16$\times$16} & {\tiny 16$\times$16} & {\tiny 16$\times$16} & {\tiny 16$\times$16} & {\tiny 16$\times$16} & {\tiny 16$\times$16} & {\tiny 16$\times$16} &
\\
{\bf \small Level 2} & {\tiny 16$\times$1} & {\tiny 1$\times$1} & {\tiny 16$\times$1} & {\tiny 1$\times$1} & {\tiny 16$\times$1} & {\tiny 1$\times$1} & {\tiny 16$\times$1} & {\tiny 1$\times$1} &
\\ 
\cmidrule[.1pt]{2-9} 
{\bf \small Level 3} &  \multicolumn{8}{c}{\tiny 16$\times$1}  & 
\\ &
\\ \cmidrule[.3pt]{1-10} &
\\&
\multicolumn{8}{c}{\bf \small Architecture II}         \\
                             & $w_i$ & $u_i$ & $w_f$ & $u_f$ & $w_o$ & $u_o$ & $w_g$ & $u_g$ \\
{\bf \small Level 1} & {\tiny 16$\times$16} & {\tiny 16$\times$16} & {\tiny 16$\times$16} & {\tiny 16$\times$16} & {\tiny 16$\times$16} & {\tiny 16$\times$16} & {\tiny 16$\times$16} & {\tiny 16$\times$16} &
\\
{\bf \small Level 2} & {\tiny 16$\times$1} & {\tiny 1$\times$1} & {\tiny 16$\times$1} & {\tiny 1$\times$1} & {\tiny 16$\times$1} & {\tiny 1$\times$1} & {\tiny 16$\times$1} & {\tiny 1$\times$1} &
\\ & 
\\ \cmidrule[.3pt]{1-10} &
\\&
 \multicolumn{8}{c}{\bf \small Architecture III}         \\
                             & $w_i$ & $u_i$ & $w_f$ & $u_f$ & $w_o$ & $u_o$ & $w_g$ & $u_g$ \\
{\bf \small Level 1} & {\tiny 16$\times$16} & {\tiny 16$\times$16} & {\tiny 16$\times$16} & {\tiny 16$\times$16} & {\tiny 16$\times$16} & {\tiny 16$\times$16} & {\tiny 16$\times$16} & {\tiny 16$\times$16} &
\\

{\bf \small Level 2} & {\tiny 16$\times$16} & {\tiny 16$\times$16} & {\tiny 16$\times$16} & {\tiny 16$\times$16} & {\tiny 16$\times$16} & {\tiny 16$\times$16} & {\tiny 16$\times$16} & {\tiny 16$\times$16} &
\\

{\bf \small Level 3} & {\tiny 16$\times$1} & {\tiny 1$\times$1} & {\tiny 16$\times$1} & {\tiny 1$\times$1} & {\tiny 16$\times$1} & {\tiny 1$\times$1} & {\tiny 16$\times$1} & {\tiny 1$\times$1} &
\\ 
\cmidrule[.1pt]{2-9}
{\bf \small Level 4} &  \multicolumn{8}{c}{\tiny 16$\times$1} \\
\bottomrule
\end{tabular}
}
\end{table}

\begin{table}
\centering
\caption{Architectures Weights Matrices' Total Elements}
\label{table:wghts_elmts}
\resizebox{0.48\textwidth}{!}
{%
\begin{tabular}{ccc}
    \toprule
    \multicolumn{1}{c}{\bfseries Architecture I} &
    \multicolumn{1}{c}{\bfseries Architecture II} &
    \multicolumn{1}{c}{\bfseries Architecture III}\\
    \midrule
	21,170 & 21,160 & 83,290\\
    \bottomrule
\end{tabular} 
}\\
\end{table}

\subsection{Forward Propagation}
The following is a general description for the forward propagation path. This example uses Architecture I as an example but similar steps are taken in the other architectures with minor changes apparent in their diagrams. With Figure~\ref{fig:art_1} presenting an overview of the structure of the whole network and considering Figure~\ref{fig:M1} as an overview of the structure of the cells in {\it Level 1} (M1) and Figure~\ref{fig:M2} as an overview of the structure of the cells in {\it Level 2} (M2) -- the input at each iteration consists of 10 seconds of time series data of the 15 input parameters and 1 bias ({\it Input} in Figure~\ref{fig:M1}) in one vector ($x_t$ in Figure~\ref{fig:art_1}) and the output of the previous cell ({\it Previous Cell Output} in Figure~\ref{fig:M1}) in another vector ($a_{t-1}$ in Figure~\ref{fig:art_1}). Each second of time series input is fed to the corresponding cell (\ie, the first seconds' 15 parameters and 1 bias are fed to first cell, the second seconds' 15 parameters and 1 bias are fed to second cell, ...) into the {\it cell gate} (shown in black color), {\it input gate} (shown in green color), {\it forget gate} (shown in blue color) and the {\it output gate} (shown in red color). If the gates ({\it input gate}, {\it forget gate} and, {\it output gate}) are seen as valves that control how much of the data flow through it, the outputs of these gates ($i_t$, $f_t$ and, $o_t$) are considered as how much these valves are opened or closed. 

First, at the {\it cell gate}, $x_t$ is dot multiplied by its weights matrix $w_g$ and $a_{t-1}$ is dot multiplied by its weights matrix $u_g$. The output vectors are summed and an activation function is applied to it as in Equation~\ref{eq:gate}. The output is called $g_t$.

Second, at the {\it input gate}, $x_t$ is dot multiplied by its weights matrix $w_i$ and $a_{t-1}$ is dot multiplied by its weights matrix $u_i$. The output vectors are summed and an activation function is applied to it as in Equation~\ref{eq:input-gate}. The output is called $i_t$.

Third, at the {\it forget gate}, $x_t$ is dot multiplied by its weights matrix $w_f$ and $a_{t-1}$ is dot multiplied by its weights matrix $u_f$. The output vectors are summed and an activation function is applied to it as in Equation~\ref{eq:forget-gate}. It controls how much of the {\it cell memory} Figure~\ref{fig:art_1} (saved from previous time-step) should pass. The output is called $f_t$.

Fourth, at the {\it output gate}, $x_t$ is dot multiplied by its weights matrix $w_o$ and $a_{t-1}$ is dot multiplied by its weights matrix $u_o$. The output vectors are summed and an activation function is applied to it as in Equation~\ref{eq:output-gate}. The output is called $o_t$.

Fifth, the contribution of the cell input {\it Input} $g_t$ and {\it cell memory} $c_{t-1}$ is decided in Equation~\ref{eq:cell_memo} by dot multiplying them by $f_t$ and $i_t$ respectively. The output of this step is the new {\it cell memory} $c_{t}$.

Sixth, cell output is also regulated by the output gate (valve). This is done by applying the sigmoid function to the {\it cell memory} $c_{t}$ and dot multiplying it by $o_t$ as shown in Equation~\ref{eq:out}. The output of this step is the final output of the cell at the current time-step $a_t$. $a_t$ is fed to the next cell in the same level and also fed to the cell in the above level as an {\it Input} $a_t$. 

The same procedure is applied at {\it Level 2} but with different weight vectors and different dimensions. Weights at {\it Level 2} have smaller dimensions to reduce their input detentions from vectors with 16 dimensions to vectors with one dimension. The output from {\it Level 2} is a one dimensional vector from each cell of the 10 cells in {\it Level 2}. These vectors are fed as one 10 dimensional vector to a simple neuron collection shown in Figure~\ref{fig:art_1} at {\it Level 3} to be dot multiplied by a weight vector to reduce the vector to a single scalar value: the final output of the network at the time-step.

\section{Evolving LSTM RNN Cells using Ant Colony Optimization}
\label{sec:algorithm}

Although the results from architecture I are promising, there is still room for further optimization in that the network may have excessive connections which confound accurate predictions and that the structure could be further optimized. A particular concern was that some connections could cause noise in the obtained results and ultimately would drift the results from their most optimum values, as this had been shown in the initial one layer feed forward neural networks with certain input parameters.  The goal of using the ant colony optimization strategy is to evolve the structure of the LSTM cells, encouraging more diverse networks and selecting the topologies that give the best performance.

The ACO algorithm operates on the fully connected inputs to the M1 and M2 cells, as shown in Figures~\ref{fig:M1} and~\ref{fig:M2}. Each M1 cell has eight 16x16 input gates, four of which take the input from the previous cell in the same layer, and four of which take the input from the time series or the cell in the lower layer.  Each M2 cell has eight 8x1 input gates, four of which receive input from the previous cell in the same layer and four from the cell in the lower layer.  

\begin{figure}
\begin{center}
\includegraphics[width=.55\textwidth]{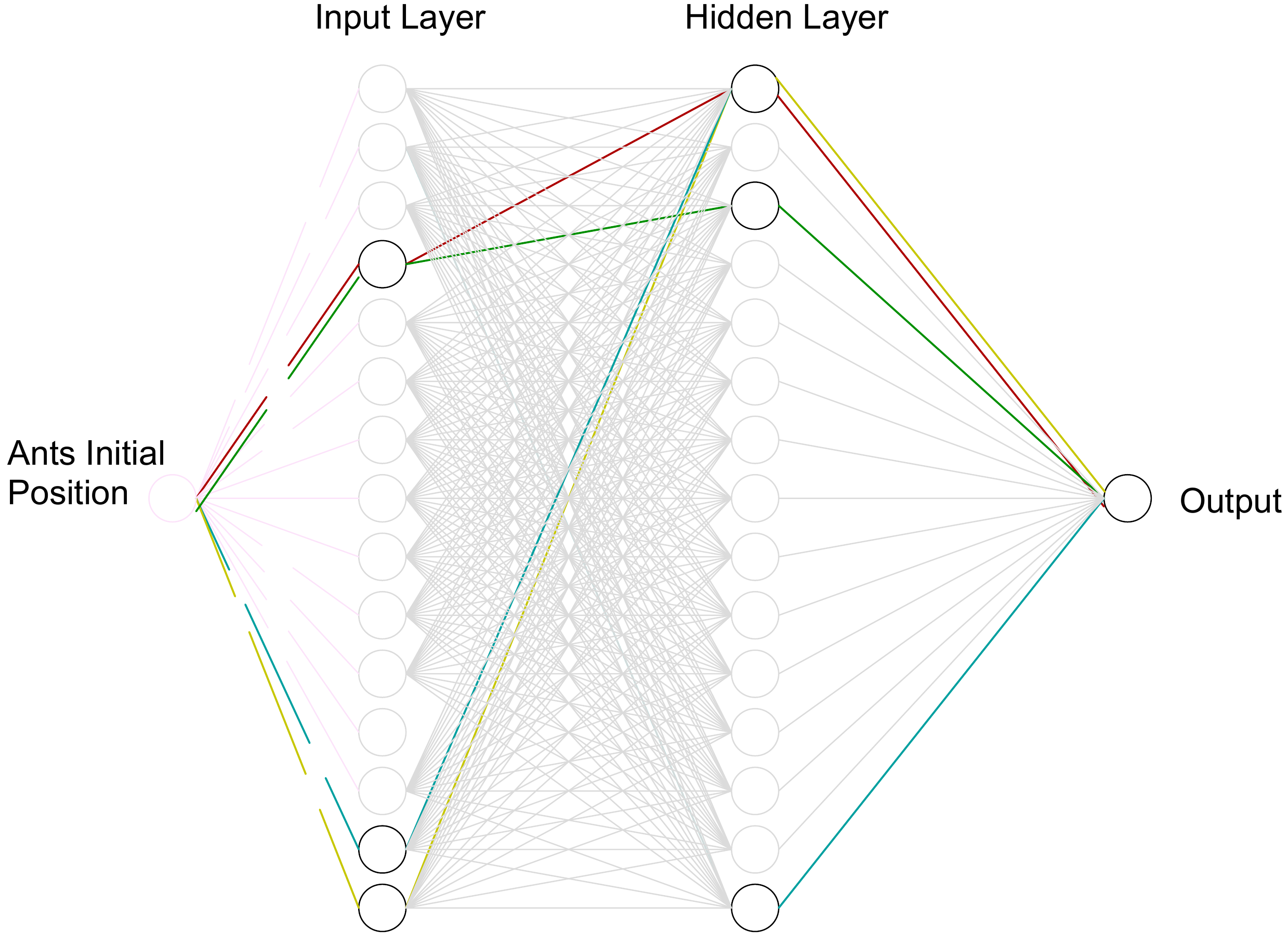}
\end{center}
\center{\caption{\label{fig:schem_ant_col} Schematic of Neural Network Structure after AOC}}
\end{figure}

The algorithm begins with a fully connected gate that will be used by the ants each time to generate new paths for new network designs. Paths are selected by the ants based on pheromones -- each connection in the network has a pheromone value that determine its probability to be chosen as a path. Given a number of ants, each one will select one path from the fully connected network. All the paths selected from all the ants are then collected, duplicated paths are removed and a design network is generated based on the new cell topology. Figure~\ref{fig:schem_ant_col} shows an example on an M1 cell, assuming four ants choosing their paths on an input gate to an M1 cell, which generates a subgraph from the potentially fully connected input gate.  The same ACO generated topology is used for each of these 8 input gates.  Figure~\ref{fig:m1_optimized} provides an example of the best found ACO optimized M1 cell.

In detail, the paths generated by ACO are used in the connections between the {\it``Input''} and the hidden layer neurons that follow it, and the {\it``Previous Cell Output''} and the hidden layer neurons that follow it. The connections between the {\it``Input''} and the hidden layer neurons that follow it are shown in the first level cells' (Figure~\ref{fig:M1} ``M1'') in  {\color{black} BLACK color}, {\color{green} GREEN color}, {\color{blue} BLUE color}, and {\color{red} RED color} at the gates of the cell. Once a hidden node in first level cell is reached by an ant, the connection between this node and the output node shown in second level cells' Figure~\ref{fig:M2} ``M2'' in {\color{black} BLACK color}, {\color{green} GREEN color}, {\color{blue} BLUE color}, and {\color{red} RED color}, will automatically be part of the evolved mesh because the ant will not have any other option to reach the output node except through that single connection.

The same generated mesh is used at all the gates: {\color{black} Main Gate}, {\color{green} Input Gate}, {\color{blue} Forget Gate}, and {\color{red} Output Gate} at the ``M1''cells and ``M2'' cells at all the time-steps in the LSTM RNN Architecture I as shown in Figure~\ref{fig:art_1}. In other words, regardless the LSTM RNN time-step, whenever there is a transition \underline{without data reduction}: the first set of connections in the generated mesh is used, and whenever there is a transition \underline{with data reduction}: the second set of connections in the generated mesh is used.

\subsection{Distributed ACO Optimization}

Evolving large LSTM RNNs is a computationally expensive process. Even training a single LSTM RNN is extremely time consuming (approximately 8.5-9 hours to train one architecture), and applying the ACO algorithm requires running the training process on each evolved topology. This significantly raises the computational requirements in time and resources necessary to process and evolve better LSTM networks. For that reason, the ant colony algorithm was parallelized using the message passing interface (MPI) for Python~\cite{dalcin2008mpi} to allow for it to be run utilizing high performance computing resources.

The distributed algorithm utilizes an asynchronous master worker approach, which has been shown to provide performance and scalability over iterative approaches in evolutionary algorithms~\cite{szymanski-ags-ppam-2007,desell-analysis-massive-eas-2010}. This approach provides an additional benefit in that it is automatically load balanced -- workers request and receive new LSTM RNNs when they have completed training previous ones, without blocking on results from other workers. The master process can generate a new LSTM RNN to be trained from whatever is currently present in its population.

The algorithm is defined in Algorithm~\ref{antcol_code}. In detail, the algorithm beings with the master process generating an initial set of network designs randomly (given a user defined number of ants), and sending these to the worker processes. When the worker receives a network design, it creates an LSTM RNN architecture by creating the LSTM cells with the according input gates and cell memory.  The generated structure is then trained on different flight data records using the backpropagation algorithm and the resulting fitness (test error) is evaluated and sent back along with the LSTM cell paths to the master process.

The master process then compares the fitness of the evaluated network to the other results in the population, inserts it into the population, and will reward the paths of the best performing networks by increasing the pheromones by 15\% of their original value if it was found that the result was better than the best in the population. However, the pheromones values are not allowed to exceed a fixed threshold of 20. The networks that did not out perform the best in the population are not penalized by reducing the pheromones along their paths. 

\begin{algorithm}
	\footnotesize
    \caption{Ant Colony Algorithm}\label{antcol_code}
    \begin{algorithmic}[1]
        \Variables
            \LineComment{A user specified number of ants.}
            \State $\textit{N\_ANTS}$

            \LineComment{A user specified maximum number of pheromones.}
            \State $\textit{N\_PHEROMONES}$

            \LineComment{The number of input parameters (15) + 1 for bias.}
            \State $\textit{N\_INPUTS} \gets 16$
        \EndVariables

        \Function{generate\_paths}{}
            \State $paths = \textbf{new}~Paths$
            \LineComment{path arrays all initialized to 0. 1 indicates a connection.}
            \State $paths.input = \textbf{array}[16]$
            \State $paths.m1 = \textbf{array}[16][16]$
            \State $paths.m2 = \textbf{array}[16]$

            \For {$ant \gets 1 \dots n\_ants$}
                \LineComment{select input path probabilistically according to pheromones}
                \State $pheromone\_sum \gets \textbf{sum}(pheromones.input)$
                \State $r \gets \textbf{uniform\_random}(0, pheromone\_sum - 1)$
                \State $input\_path \gets 0$

                \While {$r > 0$}:
                    \If {$r < pheromones.input[input\_path]$}
                        \State $paths.input\_paths[input\_path] \gets 1$
                        \State \textbf{break}
                    \Else
                        \State $r \gets r - pheromones.input[input\_path]$
                        \State $input\_path \gets input\_path + 1$
                    \EndIf
                \EndWhile

                \LineComment{select hidden path probabilistically according to pheromones}
                
                \State $pheromone\_sum \gets \textbf{sum}(pheromones.m1[input\_path])$
                \State $r \gets \textbf{uniform\_random}(0, pheromone\_sum - 1)$
                \State $hidden\_path \gets 0$
%
                \While {$r > 0$}
                    \If {$r < pheromones.m1[input\_path][hidden\_path]$}
                        \State $paths.m1\_paths[input\_path][hidden\_path] \gets 1$
                        \State $paths.m2\_paths[hidden\_path] \gets 1$
                        \State \textbf{break}
                    \Else
                        \State $r \gets r - pheromones.m1[input\_path][hidden\_path]$
                        \State $hidden\_path \gets hidden\_path + 1$
                    \EndIf
                \EndWhile
            \EndFor

            \Return {paths}
        \EndFunction

        \Function{update\_pheromones}{pheromones, paths}
            \For {$i \gets 1 \dots paths.input.length$}
                \If {$paths.input[i] = 1$}
                    \State $pheromones.input[i] \gets$ \par
                    \hskip\algorithmicindent $min(pheromones.input[i] * 1.15, MAX\_PHEROMONE)$
                \EndIf
            \EndFor

            \For {$i \gets 1 \dots paths.m1.length$}
                \For {$j \gets 1 \dots paths.m1[i].length$}
                    \If {$paths.m1[i][j] = 1$}
                        \State $pheromones.m1[i][j] \gets$ \par
                        \hskip\algorithmicindent $min(pheromones.m1[i][j] * 1.15, MAX\_PHEROMONE)$
                    \EndIf
                \EndFor
            \EndFor

        m\EndFunction

        \Procedure{Master}{}
        \LineComment{\textit{pheromones all initialized to 1}}
        \State $pheromones = \textbf{new}~Pheromones$
        \State $pheromones.input \gets array[16]$
        \State $pheromones.m1 \gets array[16][16]$
        \State $pheromones.m2 \gets array[16]$

        \State $population = \textbf{List}()$
        \algstore{myalg2}
    \end{algorithmic}
\end{algorithm}

\begin{algorithm}
	\footnotesize
    \begin{algorithmic}[1]
        \algrestore{myalg2}
        \Repeat
            \State $worker,message \gets get\_next\_message()$

            \If {$message ~\textbf{is}~ request\_paths$}
                \Return{generate\_paths()}
            \ElsIf {$message ~\textbf{is}~ report\_fitness$}
                \State $fitness, paths \gets message.get\_arguments()$
                \State $rank \gets population.inorder\_insert( ~\{fitness, paths\}~ )$
                \If {$rank = 0$}
                    \State $update\_pheromones(pheromones, paths)$
                \EndIf
            \EndIf
        \Until{finished}
        \EndProcedure

        \Procedure{Worker}{}
        \Repeat
            \State $paths \gets master.request\_paths()$
            \State $fitness \gets LSTM\_RNN(paths).backpropagate()$
            \State $master.report\_fitness(fitness, paths)$
        \Until{finished}
        \EndProcedure
    \end{algorithmic}
\end{algorithm}

\section{Implementation}
\label{sec:impelemtation}

\subsection{Programming Language}
Python's Theano Library~\cite{2016arXiv160502688short} was used to implement the neural networks.  It was chosen due to four major advantages: {\it i)} it will compile the most, if not all, of functions coded using it to C and CUDA providing fast performance, {\it ii)} it will perform the weights updates for backpropagation with minimal overhead, {\it iii)} Theano can compute the gradients of the error (cost function output) with respect to the weights, saving significant effort and time needed to manually derive the gradients, coding and debugging them (which is particularly challenging in regards to LSTM neurons), and finally, {\it iv)} it can utilize GPUs for further increased performance.

\subsection{Data Processing}
The flight data parameters used were normalized between 0 and 1. The sigmoid function was used as an activation function over all the gates and inputs/outputs. The ArcTan activation function was tested on the data, however it gave distorted results and sigmoid function provided significantly better performance.

\subsection{Machine Specifications}
The algorithm was implemented in Python using MPI for Python~\cite{dalcin2008mpi} and was run on the University of North Dakota's high performance computing cluster. The cluster is running the Red Hat Enterprise Linux (RHEL) 7.2 operating system with 31 nodes, each with 8 cores for 248 in total, 64GBs RAM per node for a total 1948 GB, and it is using InfiniBand 10 gigabit (GB) for interconnect. The same number of epochs (575) were used before to train the LSTM Network as in previous work for comparison purposes.

\subsection{Using GPUs for LSTM RNN Training}

The neural networks' weight matrices for a LSTM cell are repeated at a given time-step at a given layer. Thus, the computational cost increases if the output if these gates is computed separately, one gate at a time, as the data input/output consumes CPU cycles. This case is also obvious if a GPU is utilized for high performance computing as the cost of sending data forward and backward between the CPU (host) and GPU (device). For that, the input of a cell at a given layer is dot multiplied by a matrix that holds all of the gates weights concatenated one after the other. Then, the outputs; $g$ Equation~\ref{eq:gate}, $i$ Equation~\ref{eq:input-gate}, $f$ Equation~\ref{eq:forget-gate} and, $o$ Equation~\ref{eq:output-gate}, can be extracted from the dot product output matrix. Equation~\ref{eq:concat_mat} is an example of combining (concatenating) the weights matrices for the LSTM cells' gates of level one in Architecture I. By this, all weights are transferred between the CPU and the GPU as one data structure, which would theoretically boost the performance.

These measures were followed when using the Theano library for GPU computing to manage the GPU threads, blocks and grids as well as the data transfer between the CPU and GPU. However, the performance was slower when compared to the pure CPU version. For Architecture I as an example, one iteration through the network during the learning process, it took the GPU version more than twenty minutes while it took slightly more than two minutes for the pure CPU version. 

A further effort was made to overcome the data transfer penalty between the CPU and the GPU. The whole input data set was sent to the GPU as one data structure to avoid the data transfer through the iterations at every time series in the data and to perform those iterations on the GPU. Unfortunately, this also did not help with the performance. Ultimately, a conclusion was reached that the subject matrices are not large enough to overcome the data transfer overhead. Further study is required to determine if it is possible to achieve good performance for these types of LSTM RNNs on GPUs.

	\begin{equation}
		\label{eq:concat_mat}
		\resizebox{0.48\textwidth}{!}
		{%
		$\begin{bmatrix}
		    w_{g_{1,1}} & w_{g_{1,2}} & w_{g_{1,3}} & \dots & w_{g_{1,16}} \\
		     \vdots & \vdots & \vdots & \ddots & \vdots \\
		    w_{g_{16,1}} & w_{g_{16,2}} & w_{g_{16,3}} & \dots & w_{g_{16,16}} \\
	
			\hdashline
	
		    w_{i_{1,1}} & w_{i_{1,2}} & w_{i_{1,3}} & \dots & w_{i_{1,16}} \\
		     \vdots & \vdots & \vdots & \ddots & \vdots \\
		    w_{i_{16,1}} & w_{i_{16,2}} & w_{i_{16,3}} & \dots & w_{i_{16,16}} \\
	
			\hdashline
	
		    w_{f_{1,1}} & w_{f_{1,2}} & w_{f_{1,3}} & \dots & w_{f_{1,16}} \\
		     \vdots & \vdots & \vdots & \ddots & \vdots \\
		    w_{f_{16,1}} & w_{f_{16,2}} & w_{f_{16,3}} & \dots & w_{f_{16,16}} \\
	
			\hdashline
	
		    w_{o_{1,1}} & w_{o_{1,2}} & w_{g_{1,3}} & \dots & w_{o_{1,16}} \\
		    \vdots & \vdots & \vdots & \ddots & \vdots \\
		    w_{o_{16,1}} & w_{o_{16,2}} & w_{o_{16,3}} & \dots & w_{o_{16,16}}    
		\end{bmatrix}
		\bigodot
		\begin{bmatrix}
		    x_{11} \\
			x_{12} \\
			x_{13} \\
			\vdots  \\
			x_{16}
		\end{bmatrix}
		=
		\begin{bmatrix}
		    out_{g_{1}} \\
			\vdots   \\
			out_{g_{16}} \\
			\hdashline
		    out_{i_{1}} \\
			\vdots   \\
			out_{i_{16}} \\
			\hdashline
		    out_{f_{1}} \\
			\vdots   \\
			out_{f_{16}} \\
			\hdashline
		    out_{o_{1}} \\
			\vdots   \\
			out_{o_{16}} \\
		\end{bmatrix}$%
		}
	\end{equation}

\section{Results}
\label{sec:results}

\subsection{Previous Training Results}
Training process results from the original three architectures are shown in Table~\ref{table:training_results}. These results are directly proportional to the testing results as will be shown in the results section. The errors shown are mean squared error. 

\begin{table}
\centering
\caption{\small{Previous Training Results}}
\label{table:training_results}
\resizebox{0.48\textwidth}{!}
{%
\begin{tabular}{lrrrr}
    \toprule
    \multicolumn{1}{c}{} &
    \multicolumn{4}{c}{\bfseries Prediction Error } \\
    \multicolumn{1}{c}{} &
	\multicolumn{1}{c}{\bfseries 1 seconds} &
    \multicolumn{1}{c}{\bfseries 5 seconds} &
    \multicolumn{1}{c}{\bfseries 10 seconds} &
    \multicolumn{1}{c}{\bfseries 20 seconds}\\
    \midrule
	Architecture I & 0.000154 & 0.000398 & 0.000972 & 0.001843\\
	Architecture II & 0.001239 & 0.001516 & 0.001962 & 0.002870\\
	Architecture III & 0.000133 & 0.000409 & 0.000979 & 0.001717\\
    \bottomrule
\end{tabular}
}
\end{table}

\subsection{Cost Function}
\begin{figure}
\centering
\subfloat[Cost Plots @ 1 SEC]{\includegraphics[width=.48\textwidth, height=.20\textheight]{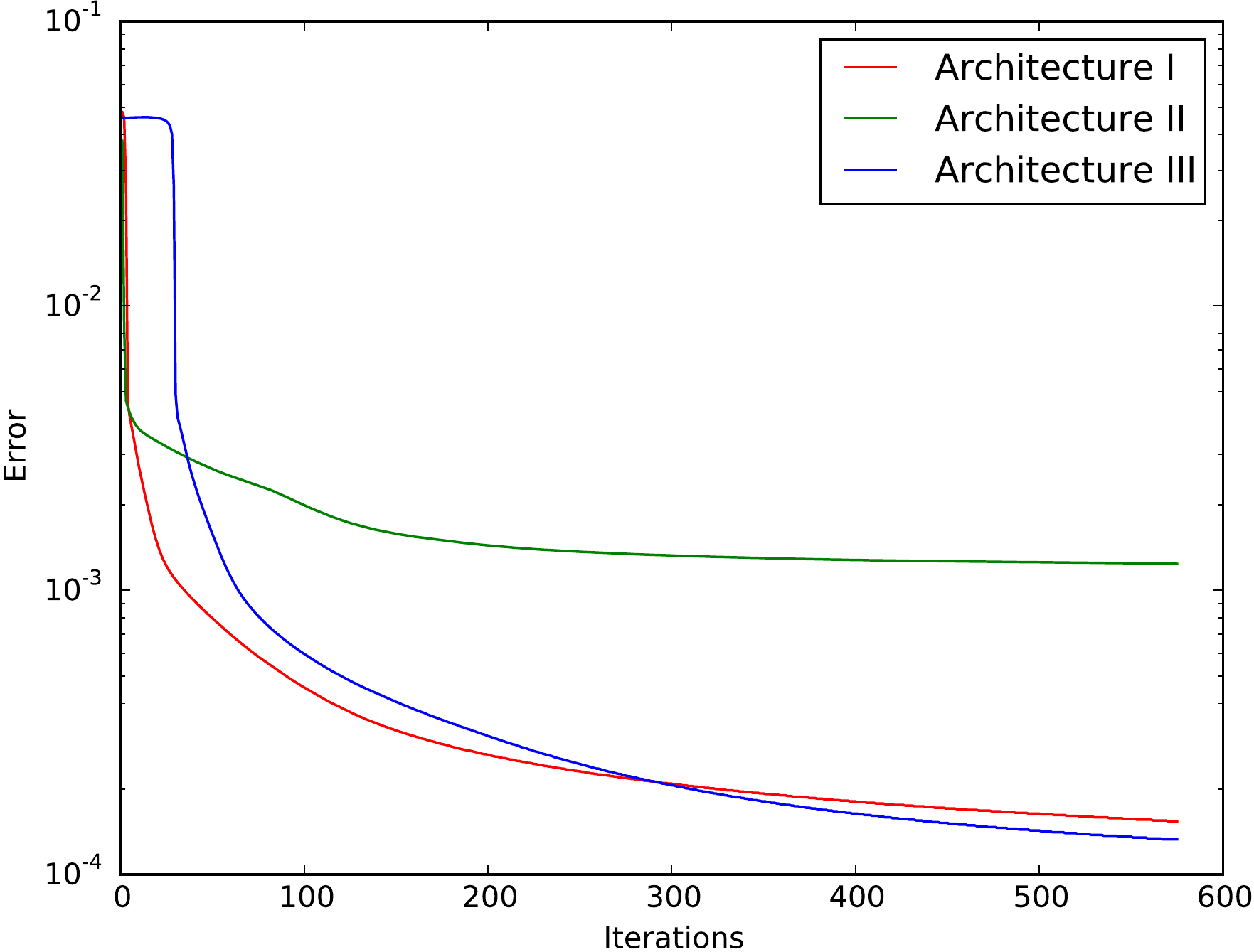}
\label{fig:art_all_cost_1sec}}
\subfloat[Cost Plots @ 5 SEC]{\includegraphics[width=.48\textwidth, height=.20\textheight]{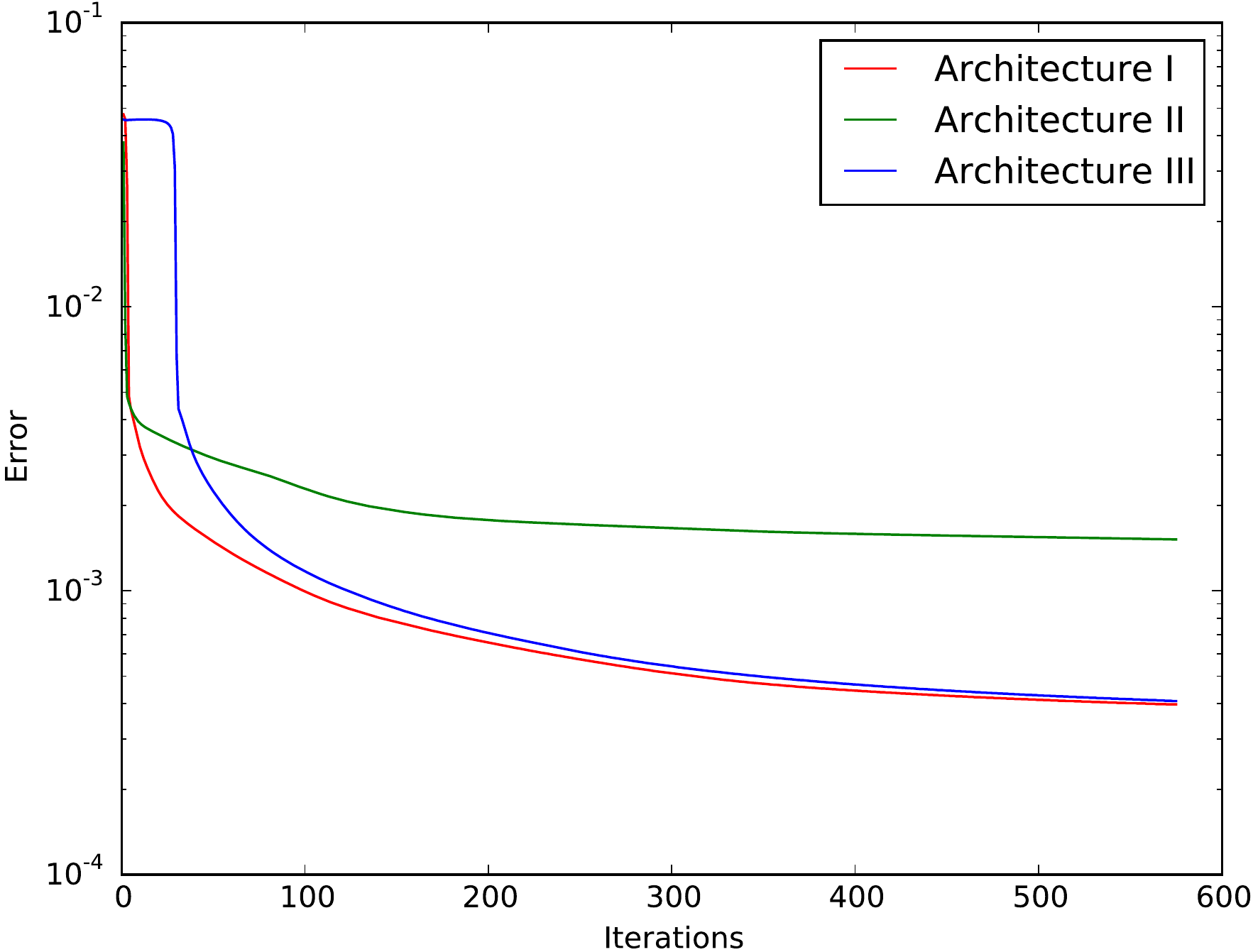}
\label{fig:art_all_cost_5sec}}
\\
\subfloat[Cost Plots @ 10 SEC]{\includegraphics[width=.48\textwidth, height=.20\textheight]{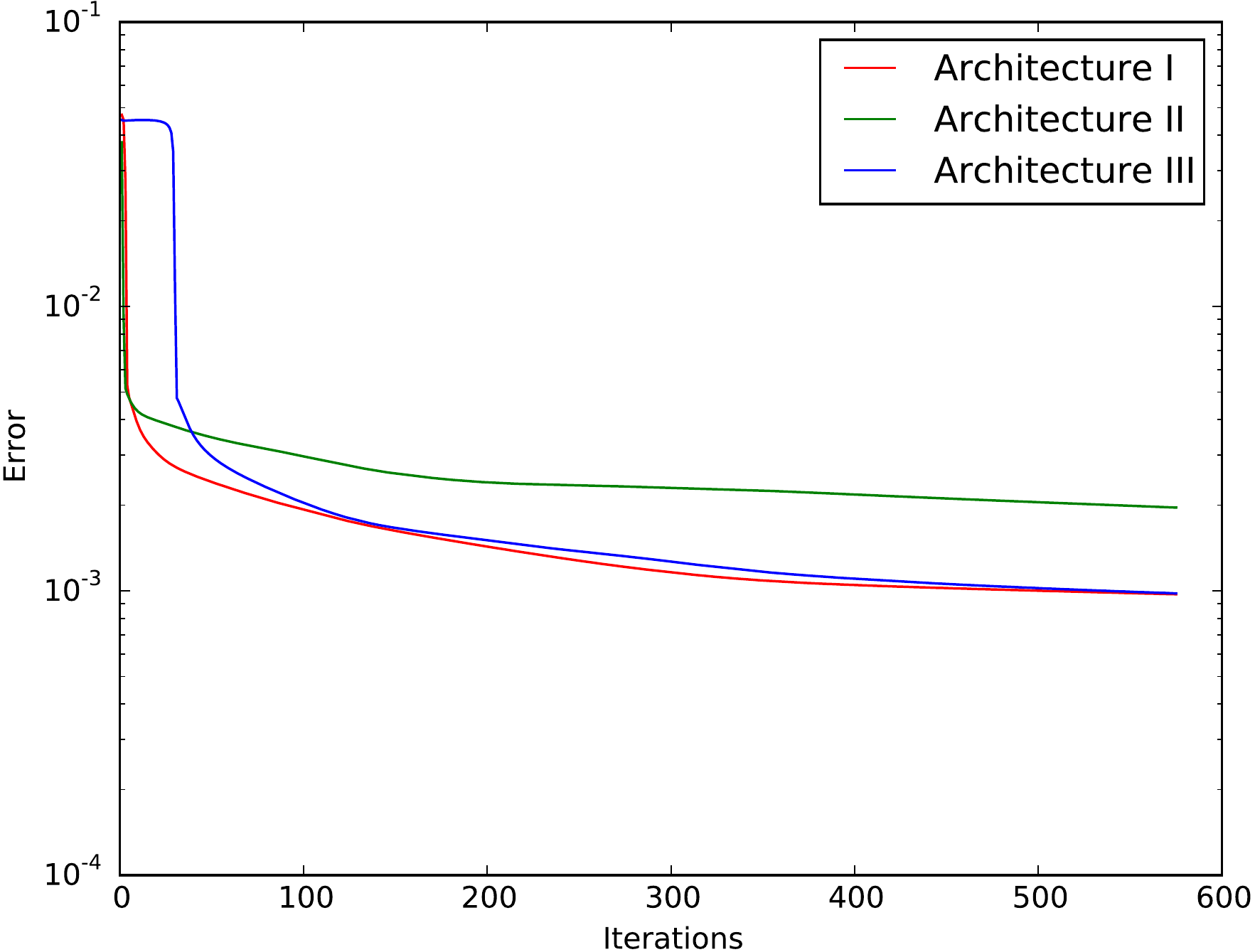}
\label{fig:art_all_cost_10sec}}
\subfloat[Cost Plots @ 20 SEC]{\includegraphics[width=.48\textwidth, height=.20\textheight]{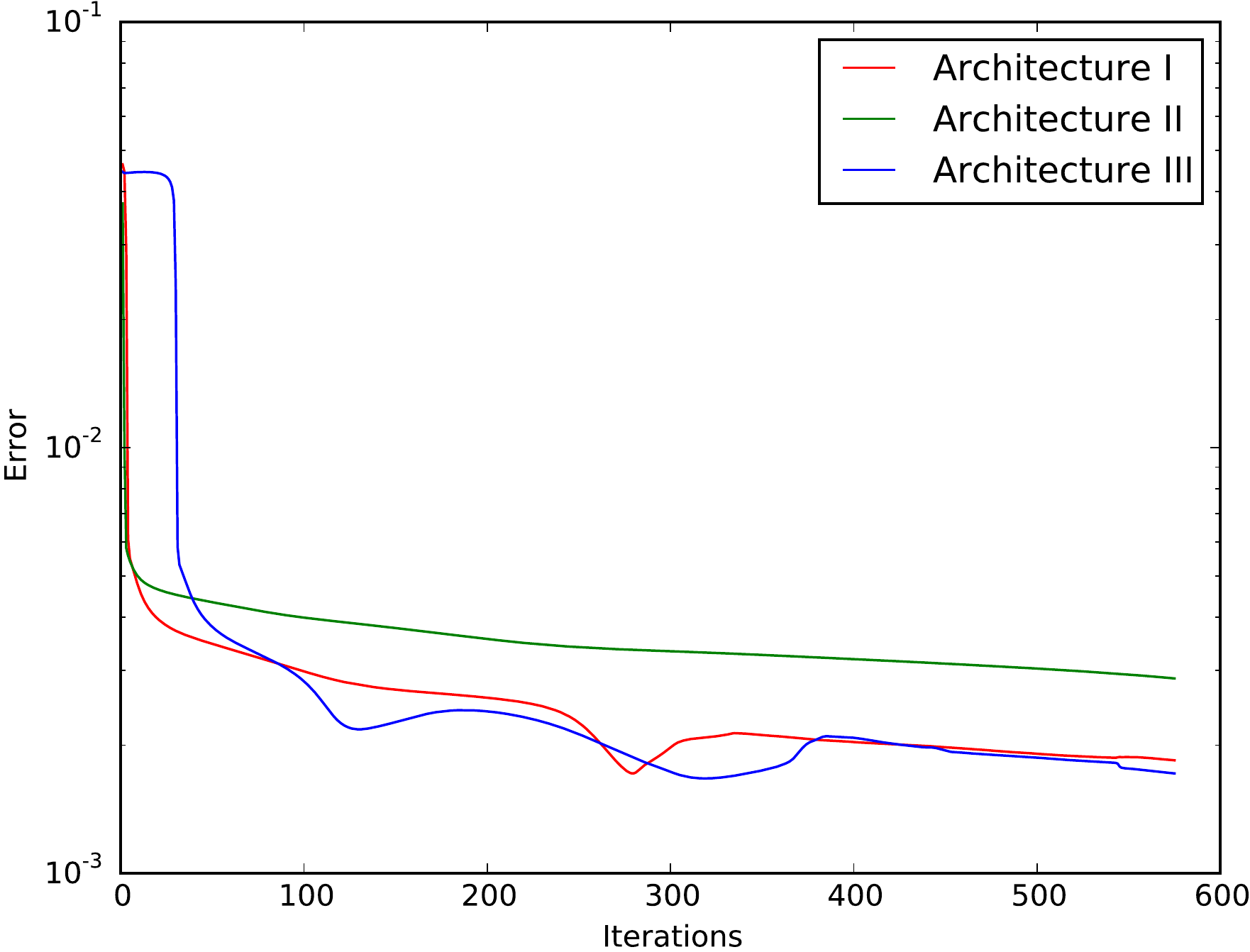}
\label{fig:art_all_cost_20sec}}
\caption{Mean squared error during the training process for the three architectures predicting vibration in 1, 5, 10, and 20 future sec.}
\label{fig:costs_1sec}
\end{figure}

Mean squared error was used to train the neural networks as it provides a smoother optimization surface for backpropagation than mean average error. The cost function output for predicting 1 sec, 5 sec, 10 sec and, 20 sec is shown in Figures~\ref{fig:art_all_cost_1sec}, ~\ref{fig:art_all_cost_5sec}, ~\ref{fig:art_all_cost_10sec} and, ~\ref{fig:art_all_cost_20sec} respectively. Results are shown in logarithmic scale. 

\subsection{Previous Architecture Results}

Mean Squared Error (MSE) (shown in Equation~\ref{eq:err}) was used as an error measure to train the three architectures, which resulted in values shown in Table~\ref{table:results}. Mean Absolute Error (MAE) (shown in Equation~\ref{eq:new_err}) was used as a final measure of accuracy for the three architectures, with results shown in Table~\ref{table:new_results}. As the parameters were normalized between 0 and 1, the MAE is also the percentage error.

\begin{equation}
    Error= \frac{0.5 \times \sum(Actual\:Vib - Predicted\:Vib)^2}{Testing\:Seconds}
    \label{eq:err}
\end{equation}

\begin{equation}
    Error= \frac{\sum[ABS(Actual\:Vib - Predicted\:Vib)]}{Testing\:Seconds}
    \label{eq:new_err}
\end{equation}

\begin{table}
\centering
\caption{\small{Previous Testing Process Mean Squared Error}}
\label{table:results}
\resizebox{0.48\textwidth}{!}
{%
\begin{tabular}{lrrrr}
    \toprule
    \multicolumn{1}{c}{} &
    \multicolumn{4}{c}{\bfseries Prediction Error} \\
    \multicolumn{1}{c}{} &
    \multicolumn{1}{c}{\bfseries 1 seconds} &
	\multicolumn{1}{c}{\bfseries 5 seconds} &
    \multicolumn{1}{c}{\bfseries 10 seconds} &
    \multicolumn{1}{c}{\bfseries 20 seconds}\\
    \midrule
	Architecture I   & 0.000792 & 0.001165 & 0.002926 & 0.010427\\
	Architecture II  & 0.010311 & 0.009708 & 0.009056 & 0.012560\\
	Architecture III & 0.000838 & 0.002386 & 0.004780 & 0.041417\\
    \bottomrule
\end{tabular}
}
\end{table}

\begin{table}
\centering
\caption{\small {Previous Testing Process Mean Absolute Error}}
\label{table:new_results}
\resizebox{0.48\textwidth}{!}
{%
\begin{tabular}{lrrrr}
    \toprule
    \multicolumn{1}{c}{} &
    \multicolumn{4}{c}{\bfseries Prediction Error} \\
    \multicolumn{1}{c}{} &
    \multicolumn{1}{c}{\bfseries 1 seconds} &
	\multicolumn{1}{c}{\bfseries 5 seconds} &
    \multicolumn{1}{c}{\bfseries 10 seconds} &
    \multicolumn{1}{c}{\bfseries 20 seconds}\\
    \midrule
	Architecture I   & 0.028407 & 0.033048 & 0.055124 & 0.101991\\
	Architecture II  & 0.098357 & 0.097588 & 0.096054 & 0.112320\\
	Architecture III & 0.027621 & 0.048056 & 0.070360 & 0.202609\\
    \bottomrule
\end{tabular}
}
\end{table}

\begin{figure}
\centering
\subfloat[Architecture I Results Plot @ 05 SEC]{\includegraphics[width=.48\textwidth, height=.20\textheight]{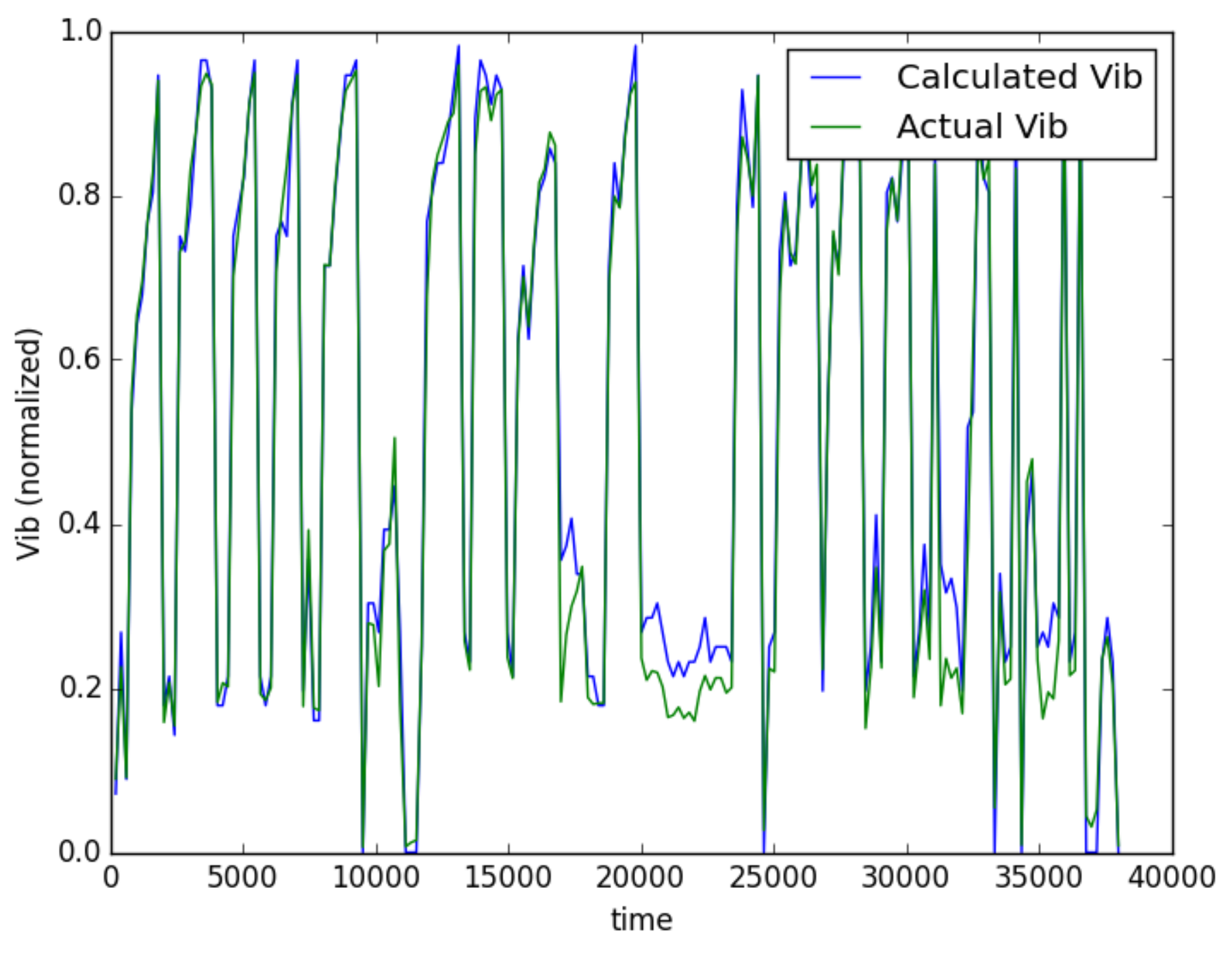}
\label{fig:res_arti_05}}
\subfloat[Architecture II Results Plot @ 05 SEC]{\includegraphics[width=.48\textwidth, height=.20\textheight]{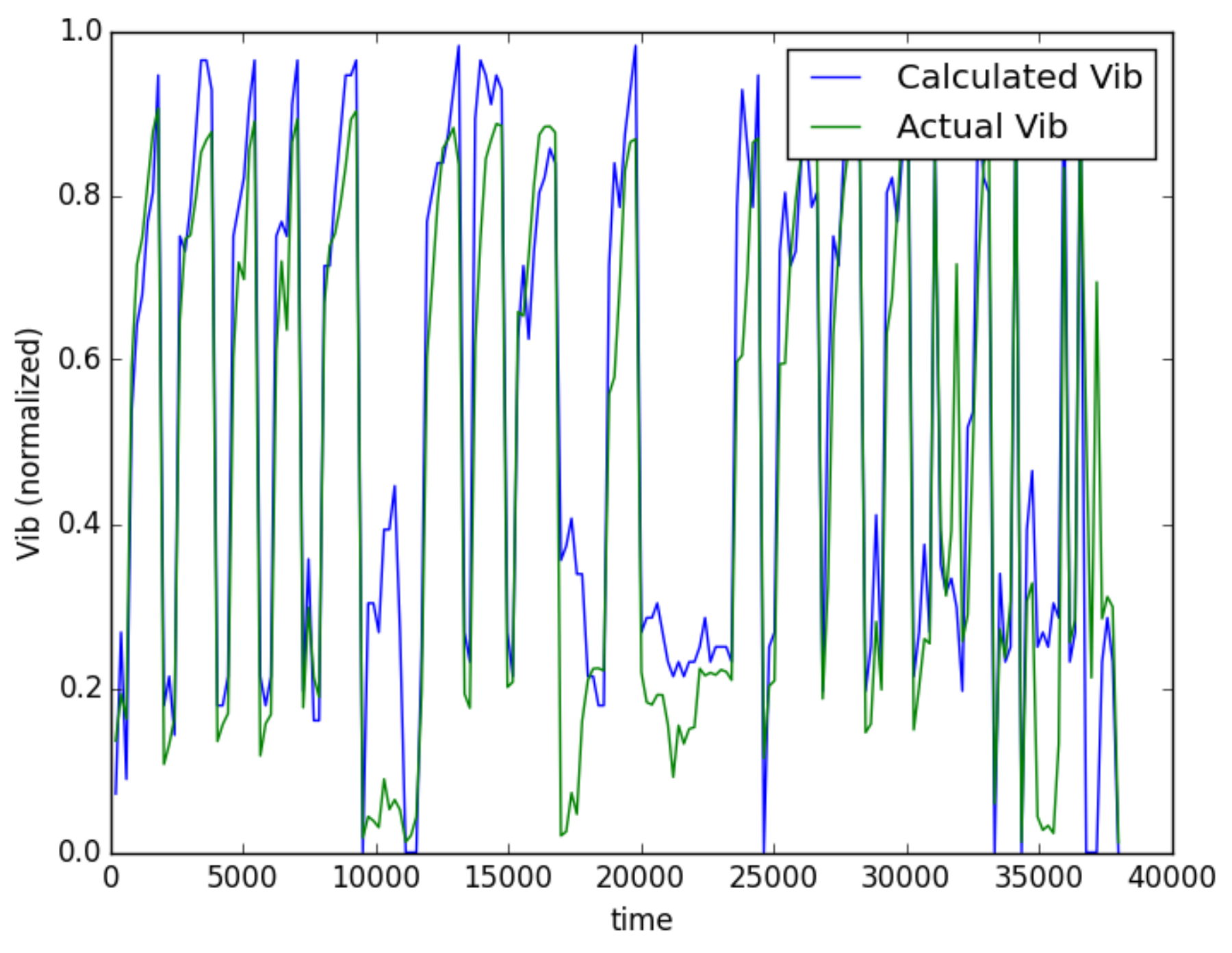}
\label{fig:res_artii_05}}
\\
\subfloat[Architecture I Results Plot @ 10 SEC]{\includegraphics[width=.48\textwidth, height=.20\textheight]{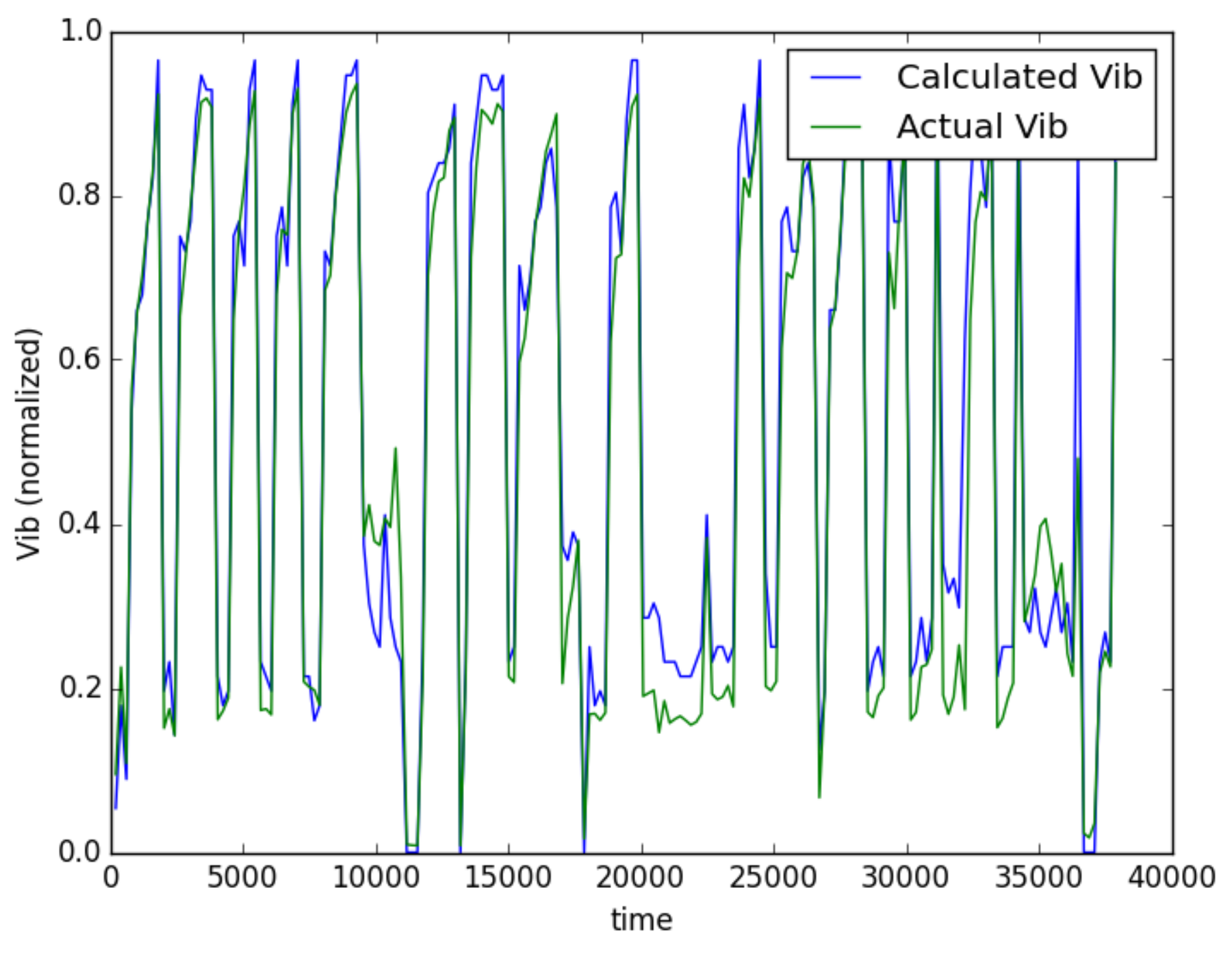}
\label{fig:res_arti_10}}
\subfloat[Architecture II Results Plot @ 10 SEC]{\includegraphics[width=.48\textwidth, height=.20\textheight]{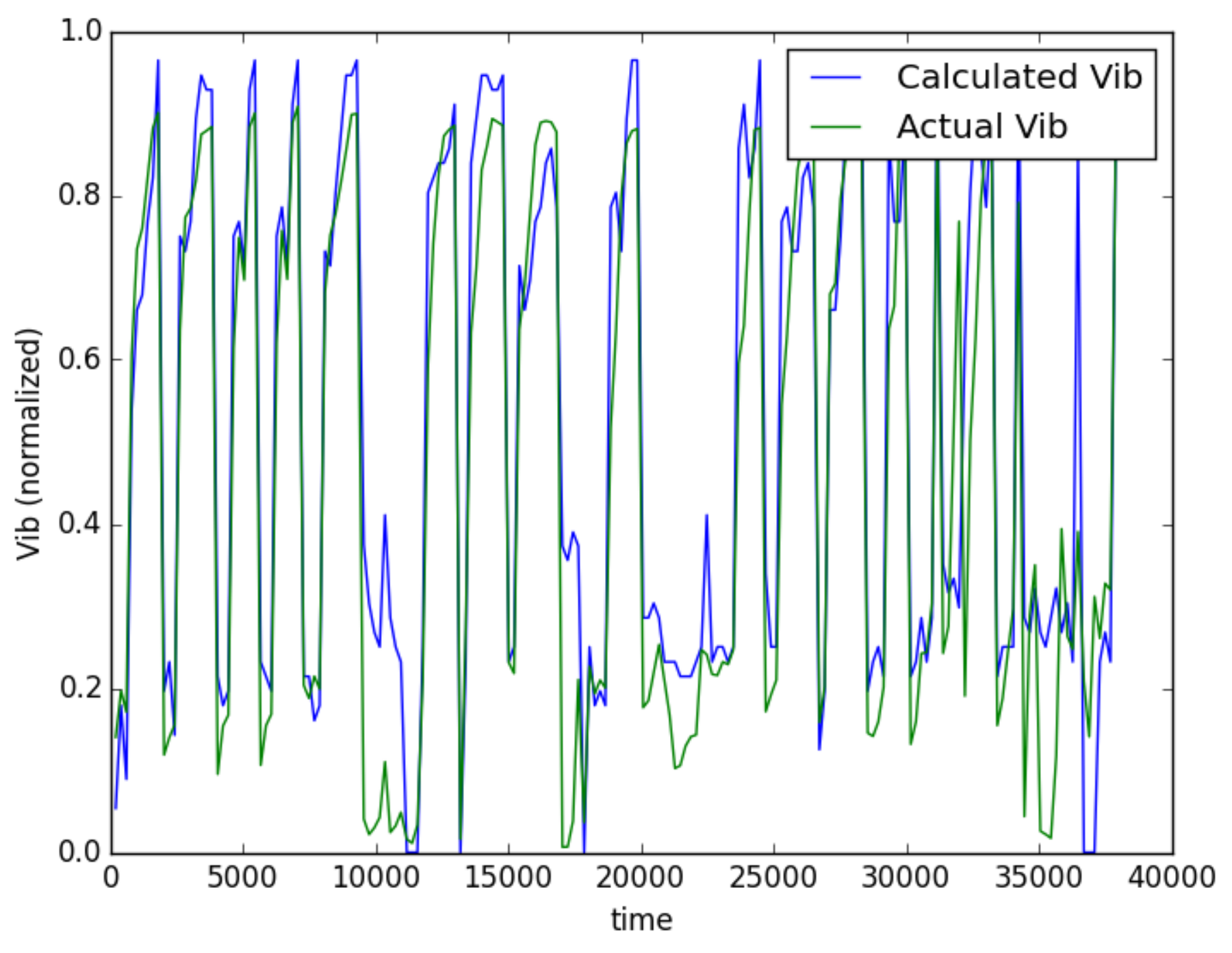}
\label{fig:res_artii_10}}
\\
\subfloat[Architecture I Results Plot @ 20 SEC]{\includegraphics[width=.48\textwidth, height=.20\textheight]{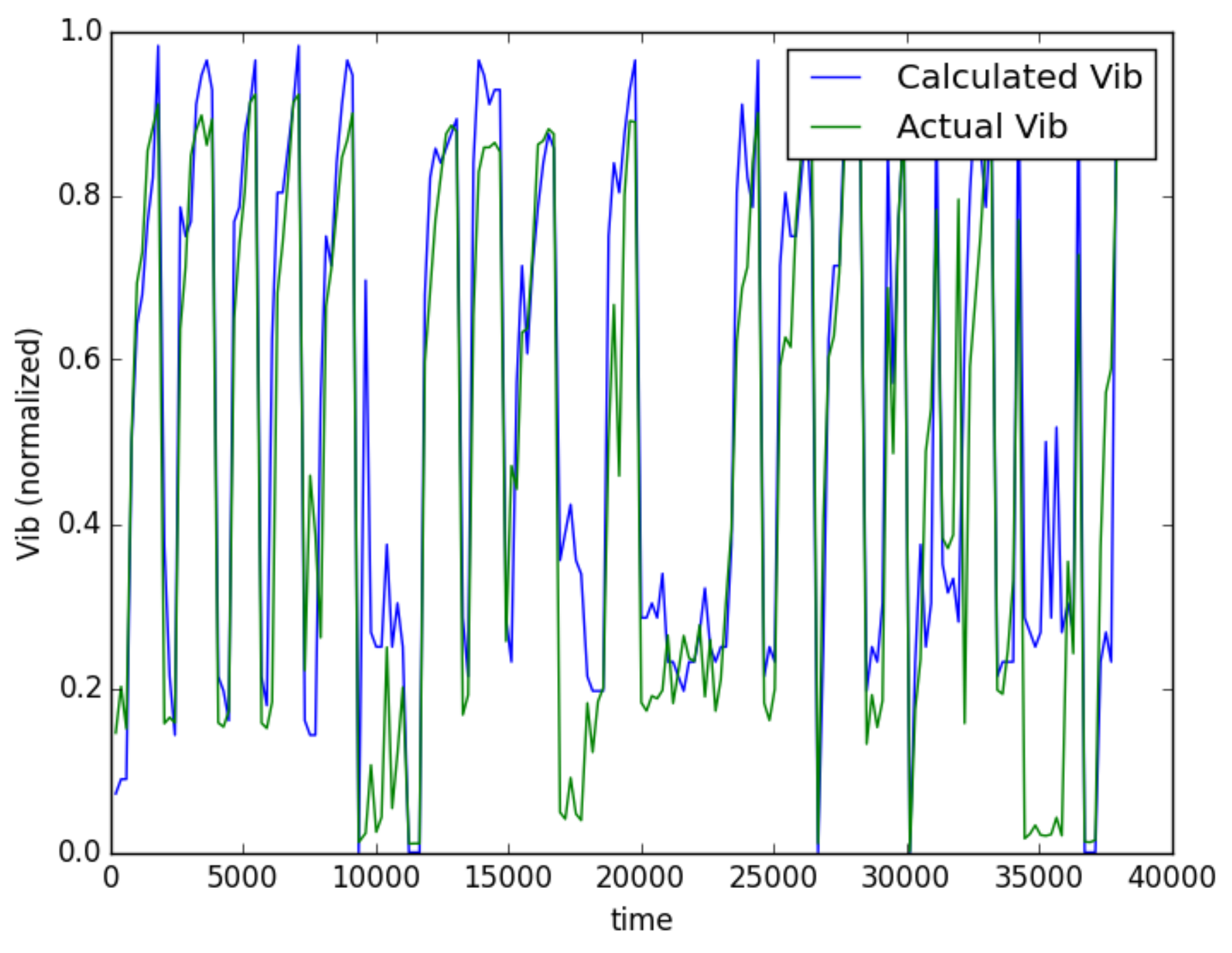}
\label{fig:res_arti_20}}
\subfloat[Architecture II Results Plot @ 20 SEC]{\includegraphics[width=.48\textwidth, height=.20\textheight]{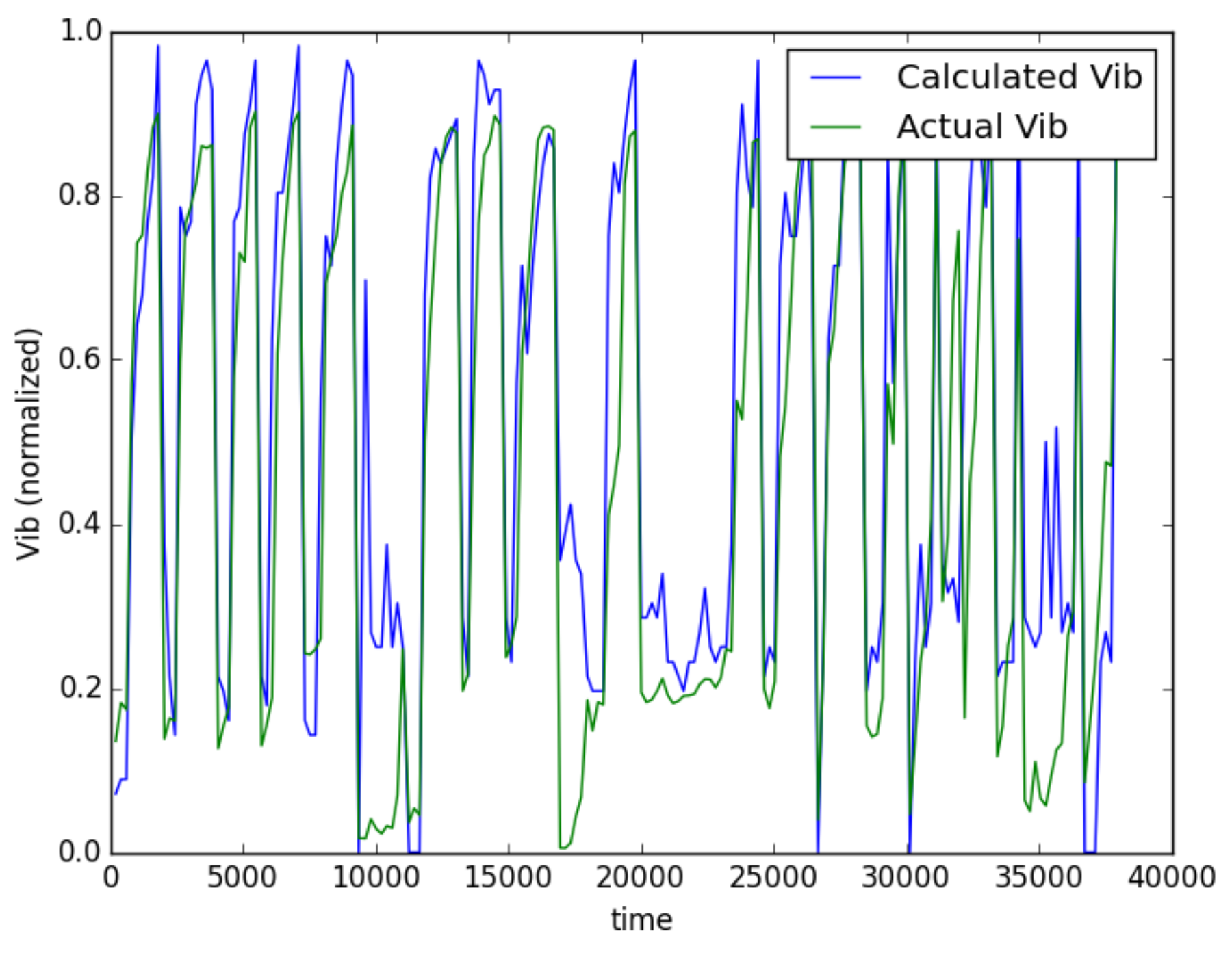}
\label{fig:res_artii_20}}
\caption{Plotted results for Architectures I and II for 5, 10 and 20 seconds in the future.}
\label{fig:res_arti_artii}
\end{figure}

\begin{figure}
\centering
\subfloat[{\footnotesize Architecture I Results Plot @ 1 SEC (all test flight)}]{\includegraphics[width=.48\textwidth, height=.20\textheight]{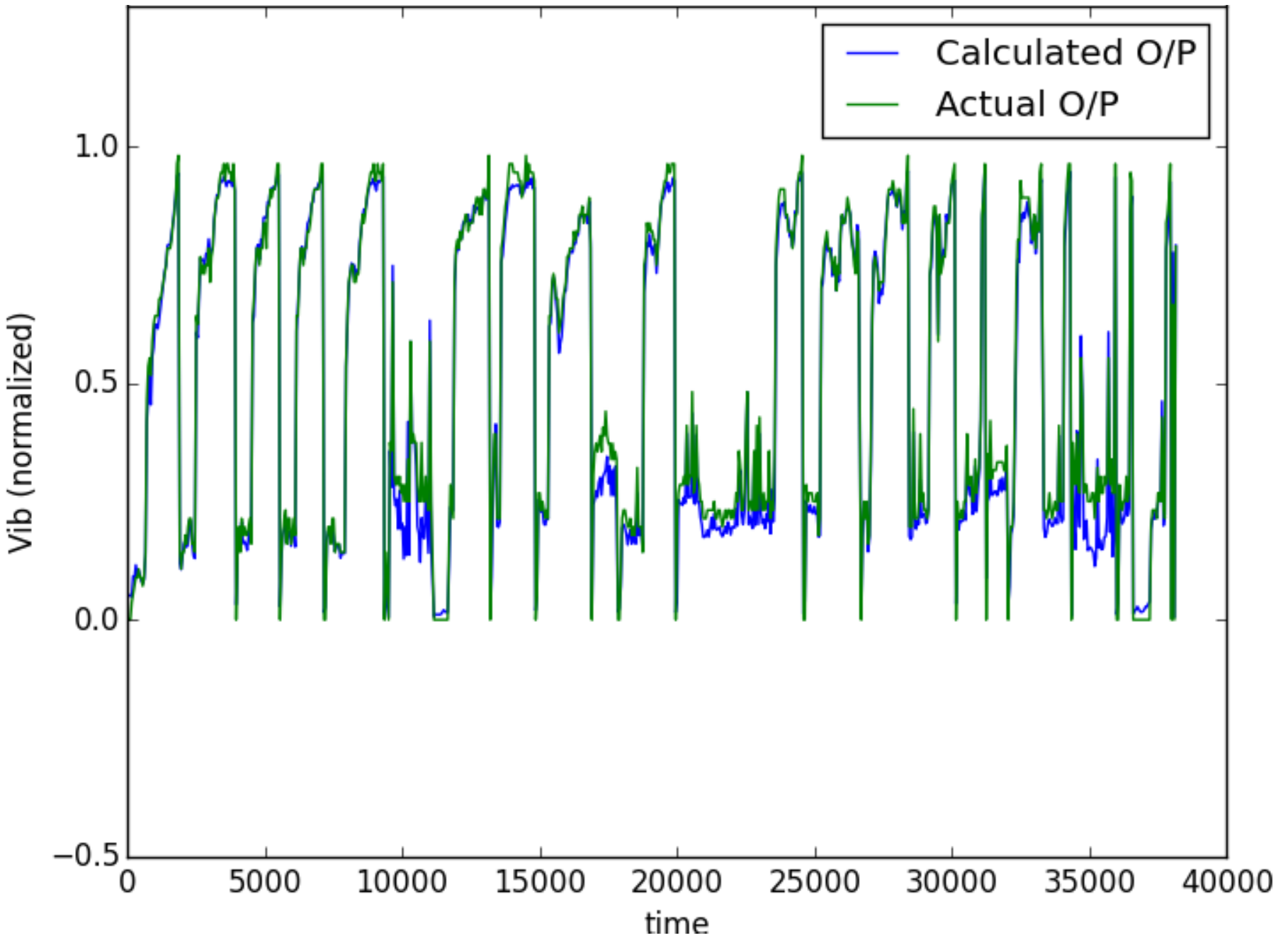}
\label{fig:arti_1sec}}
\subfloat[{\footnotesize{Architecture III Results Plot @ 05 SEC (all test flight)}}]{\includegraphics[width=.48\textwidth, height=.20\textheight]{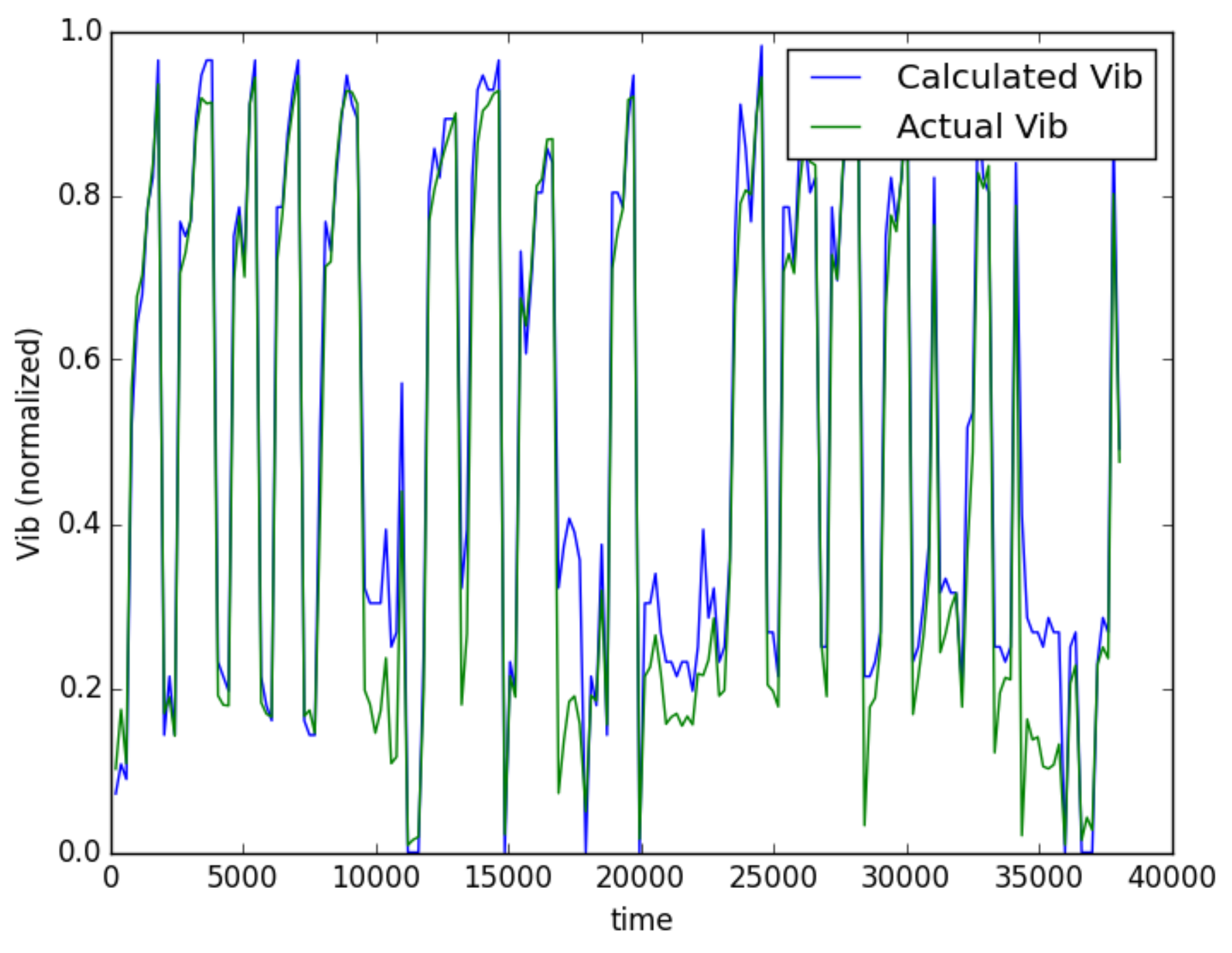}
\label{fig:res_artiii_05}}
\\
\subfloat[{\footnotesize Architecture II Results Plot @ 1 SEC (all test flight)}]{\includegraphics[width=.48\textwidth, height=.20\textheight]{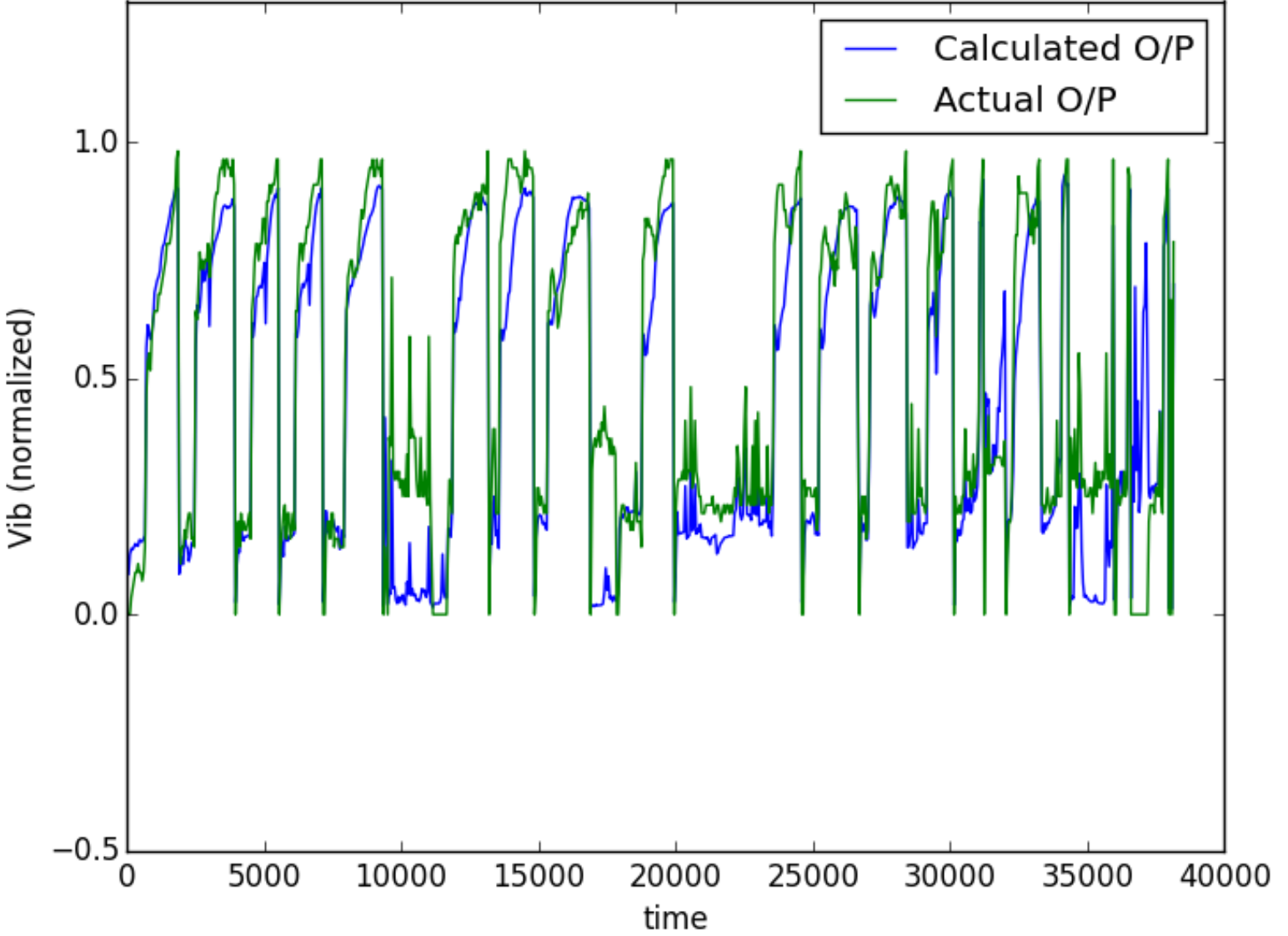}
\label{fig:artii_1sec}}
\subfloat[{\footnotesize Architecture III Results Plot @ 10 SEC (all test flight)}]{\includegraphics[width=.48\textwidth, height=.20\textheight]{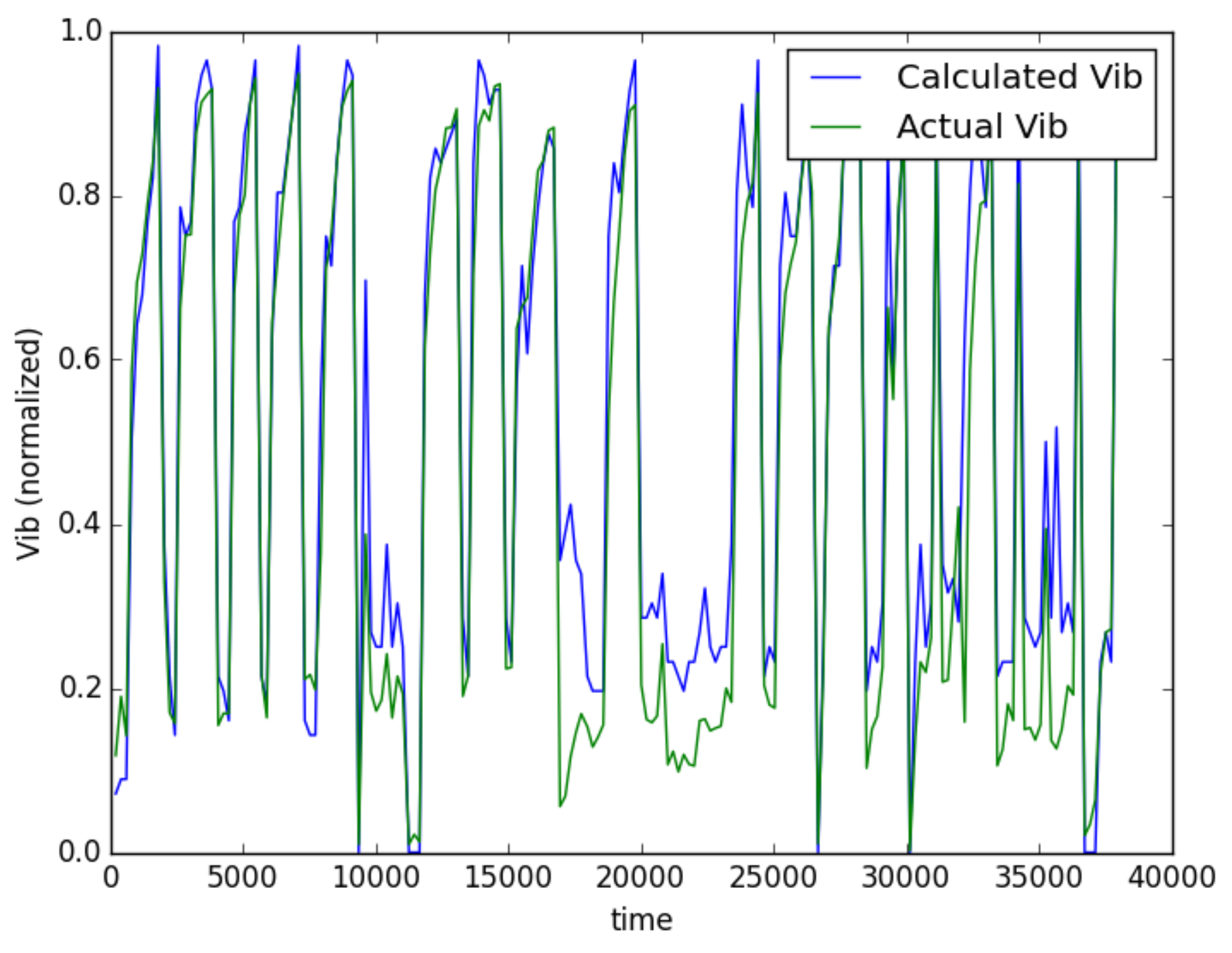}
\label{fig:res_artiii_10}}
\\
\subfloat[{\footnotesize Architecture III Results Plot @ 1 SEC (all test flight)}]{\includegraphics[width=.48\textwidth, height=.20\textheight]{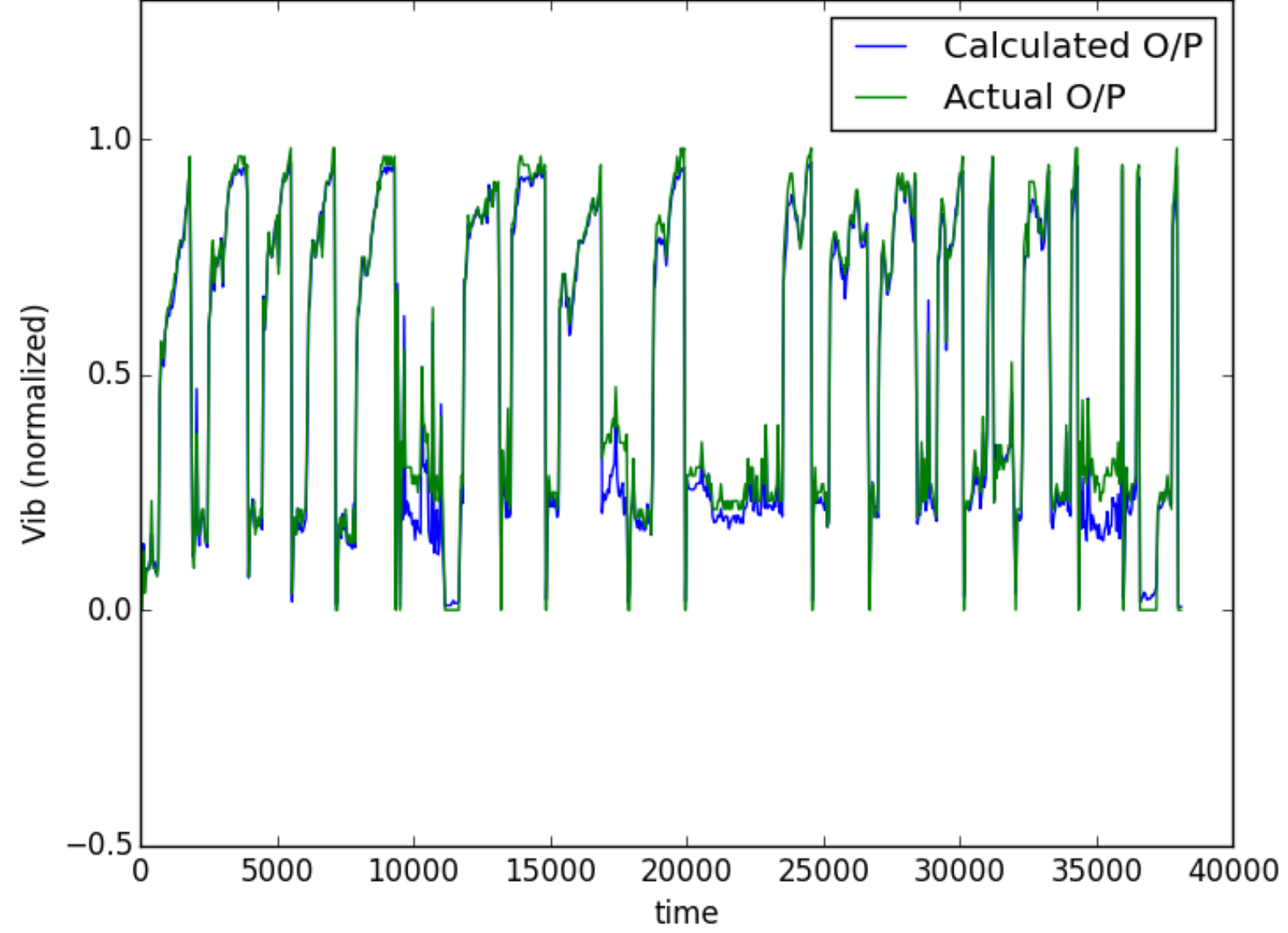}
\label{fig:artiii_1sec}}
\subfloat[{\footnotesize Architecture III Results Plot @ 20 SEC (all test flight)}]{\includegraphics[width=.48\textwidth, height=.20\textheight]{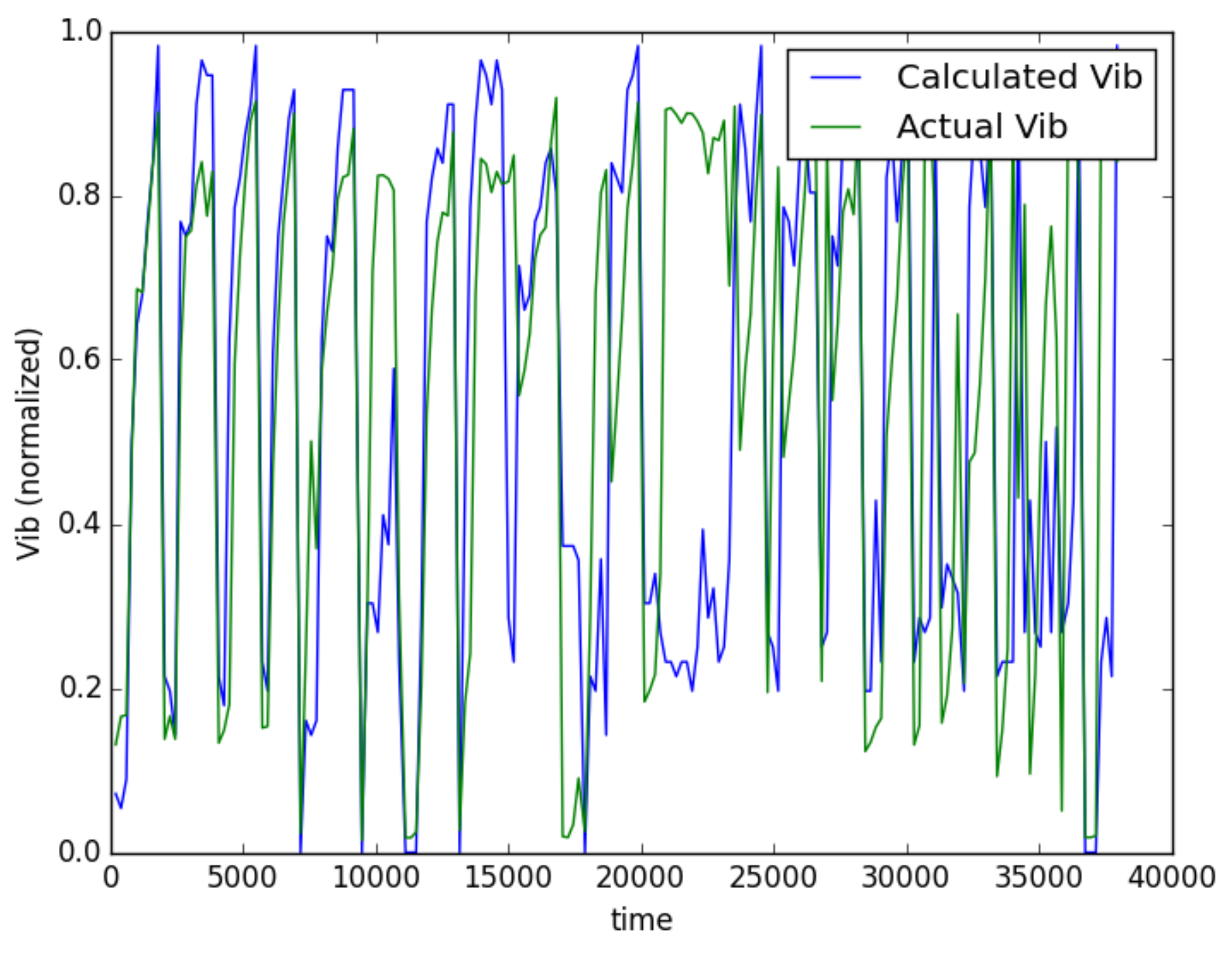}
\label{fig:res_artiii_20}}
\\
\subfloat[{\footnotesize Architecture I @ 1 SEC (one flight)}]{\includegraphics[width=.48\textwidth, height=.20\textheight]{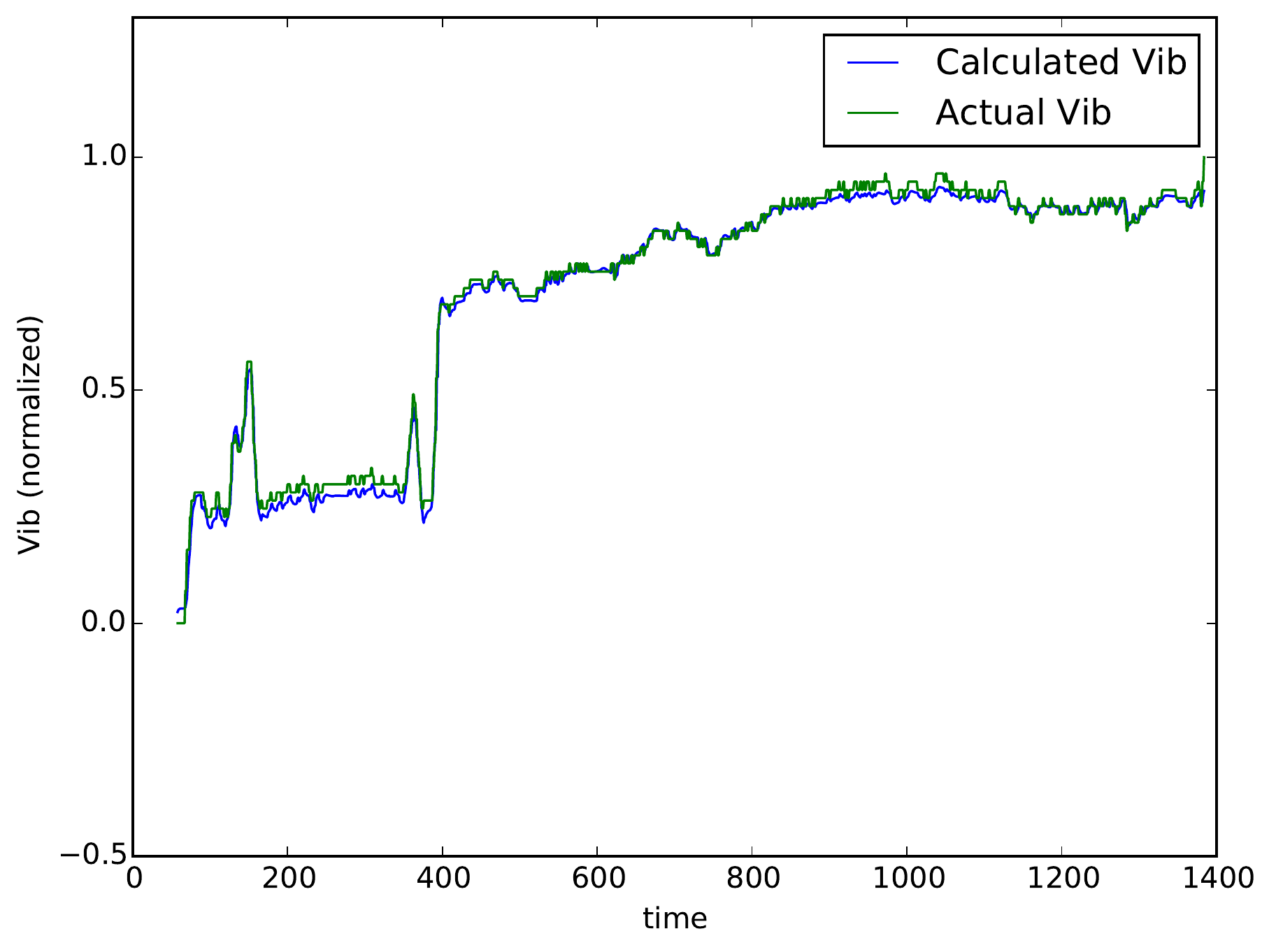}
\label{fig:arti_single_flight_1}}
\subfloat[{\footnotesize Architecture III @ 1 SEC (one flight)}]{\includegraphics[width=.48\textwidth, height=.20\textheight]{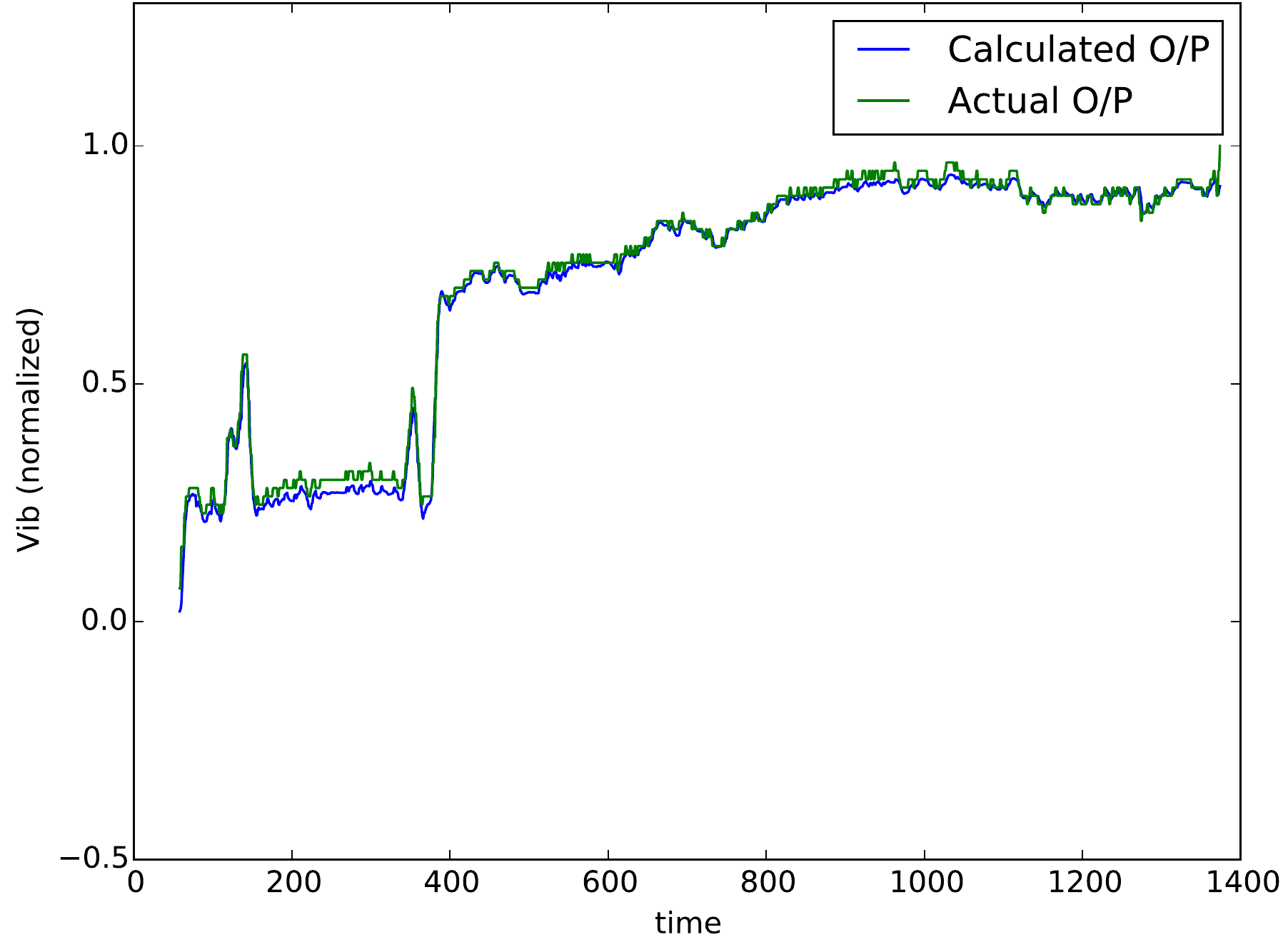}
\label{fig:artiii_single_flight_1}}
\caption{Plotted results (cont.).}
\label{fig:res_cnt}
\end{figure}

Figures~\ref{fig:res_arti_artii}, and Figures~\ref{fig:res_artiii_05}, \ref{fig:res_artiii_10}, \ref{fig:res_artiii_20} present the predictions for all the test flights condensed on the same plot. The time shown on the x-axis is the total time for all the test flights. Each flight ends when the vibration reaches the max critical value (normalized to 1) and then the next flight in the test set begins. Figures~\ref{fig:arti_artiii-single_flight} provides an uncompressed example of Architecture I and Architecture III predicting vibration 5, 10 and, 20 seconds in the future over a single flight from the testing data.

\subsubsection{Results of Architecture I}
The results of this architecture, shown in Table~\ref{table:results}, came out to be the best results regarding the overall accuracy of the vibration prediction. There is more misalignment between the actual and calculated vibration values as predictions are made further in the future, as shown in Figures~\ref{fig:res_arti_05}, \ref{fig:res_arti_10}, \ref{fig:res_arti_20}, however this is to be expected as it is more challenging to predict further in the future. Also, it can be seen that the prediction of higher peaks is more accurate than the prediction of lower peaks, as if the neural network is tending to learn more about the max critical vibration value, which is favorable for this project.

To test this architecture further, the architecture was trained and tested on the same data set but for predicting vibration just one second in future. As expected, the results showed improvement in mean absolute error over all the test flights by about 0.5\% compared to the results of the same architecture predicting five seconds in the future. A plot of the test data prediction for this experiment is shown in Figure~\ref{fig:arti_1sec}. Also, for comparison, a plot for the same flights plotted in Figure~\ref{fig:arti_artiii-single_flight} is shown in Figure~\ref{fig:arti_single_flight_1}.



\subsubsection{Results of Architecture II}
The results of this architecture in Table~\ref{table:results} came out to be the least successful in vibration prediction. While it managed to predict much of the vibration, its performance was weak at the peaks (either low or high) compared to the other architectures, as shown in Figures~\ref{fig:res_artii_10}, \ref{fig:res_artii_10}, \ref{fig:res_artii_10}. However, it is also worth mentioning that somehow the lower peaks were better at some positions on the curve of this architecture, compared to the other architectures. A potential reason for the poor performance of this architecture is due to using the average of values from the LSTM second layer output, the other two architectures can weight the values from the LSTM second layer output for more accuracy.

\subsubsection{Results of Architecture III}
\begin{figure}
\centering
\subfloat[ART I predicting vibration 5 seconds in the future for one flight.]{\includegraphics[width=.48\textwidth, height=.20\textheight]{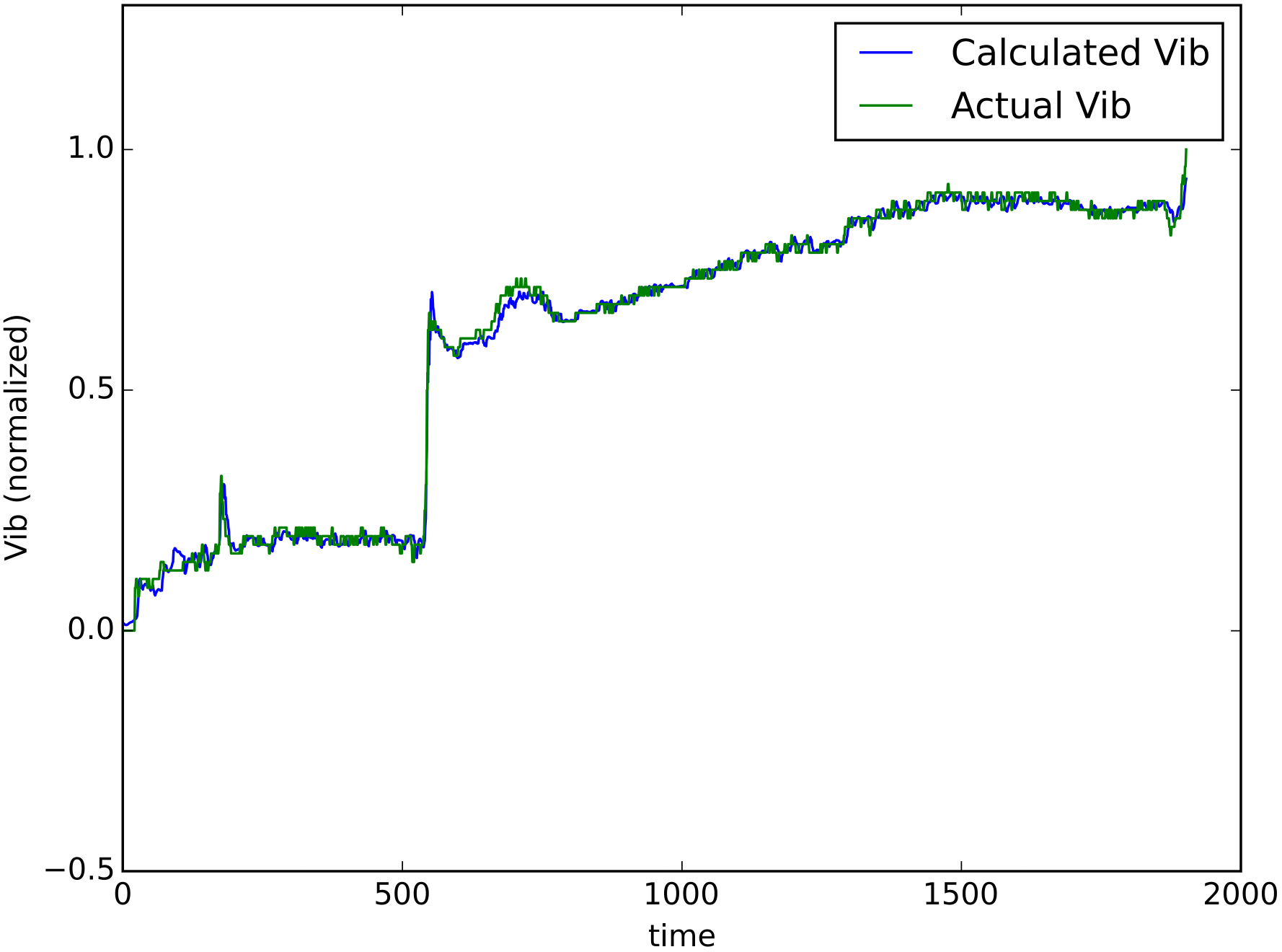}
\label{fig:arti-single_flight_05}}
\subfloat[ART III predicting vibration 5 seconds in the future for one flight.]{\includegraphics[width=.48\textwidth, height=.20\textheight]{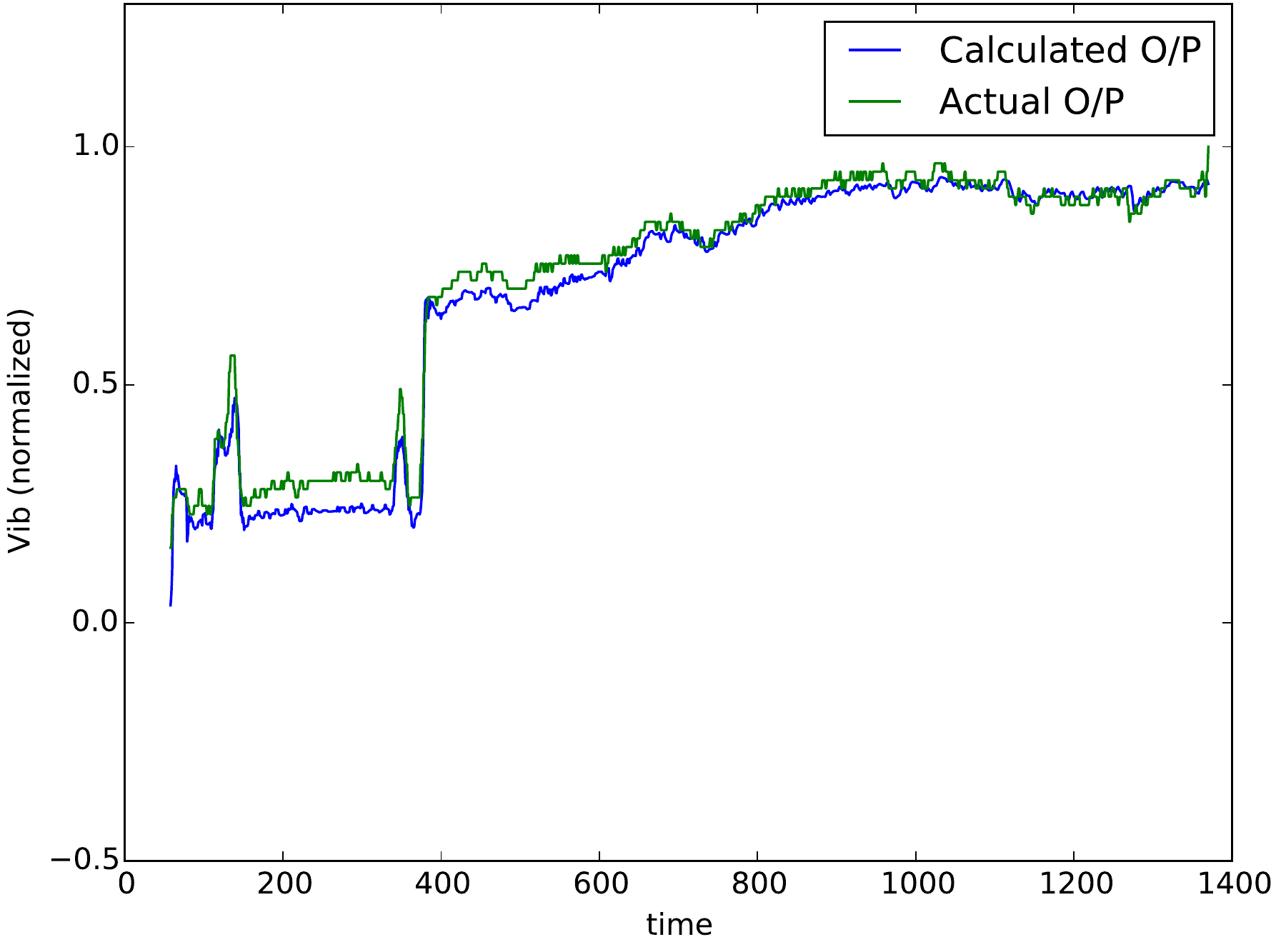}
\label{fig:artiii-single_flight_5}}
\\
\subfloat[ART I predicting vibration 10 seconds in the future for one flight.]{\includegraphics[width=.48\textwidth, height=.20\textheight]{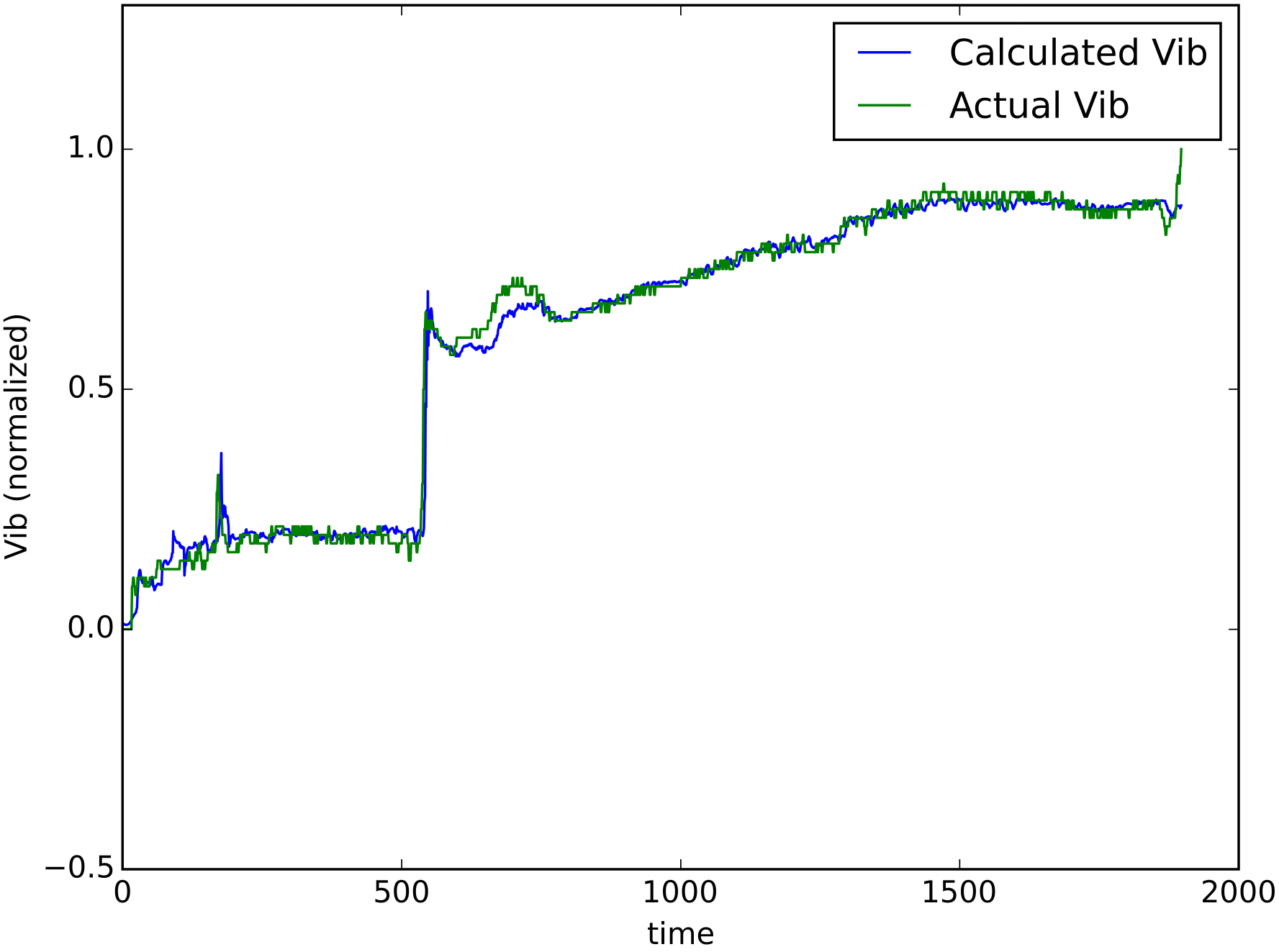}
\label{fig:arti-single_flight_10}}
\subfloat[ART III predicting vibration 10 seconds in the future for one flight.]{\includegraphics[width=.48\textwidth, height=.20\textheight]{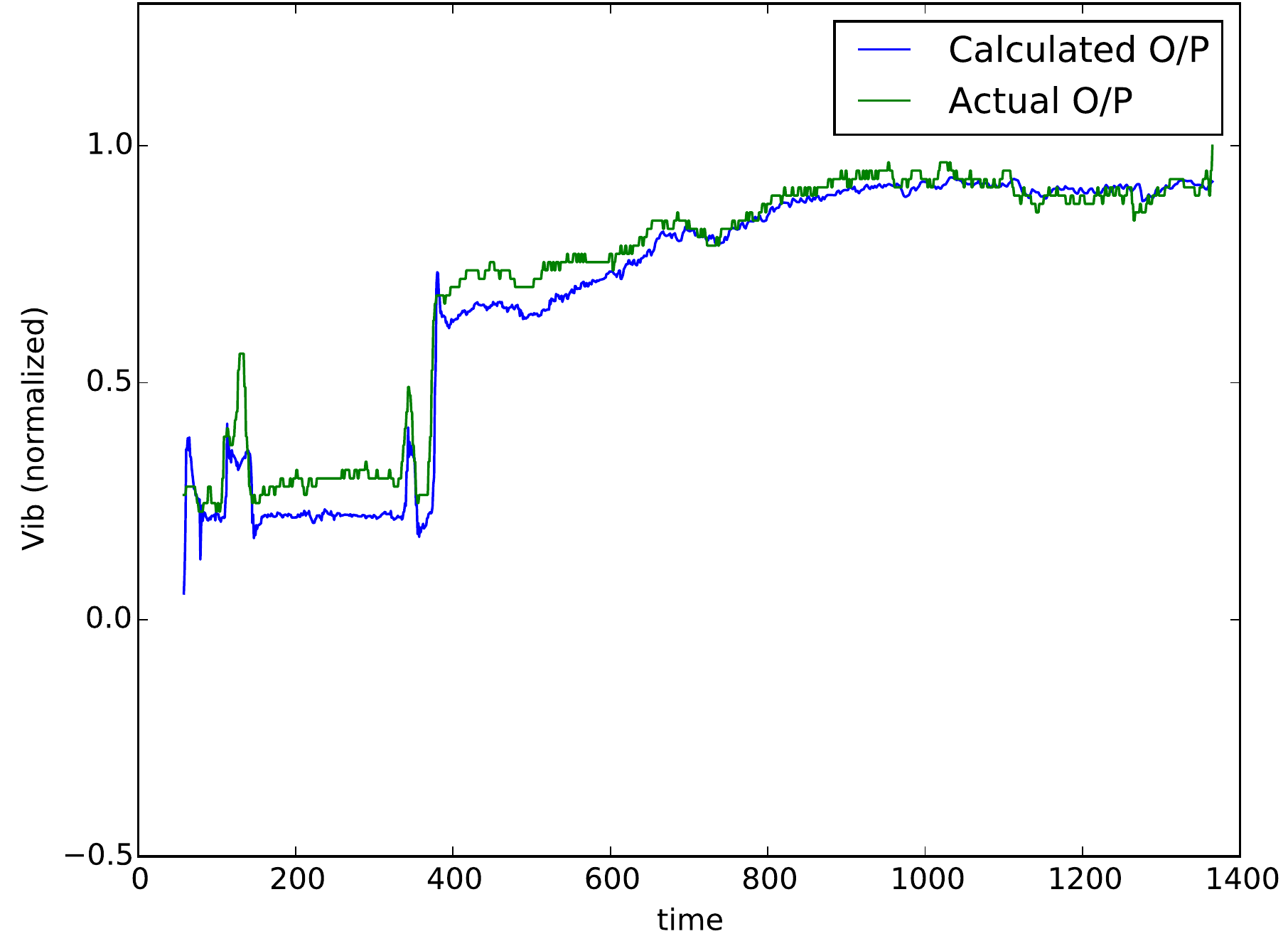}
\label{fig:artiii-single_flight_10}}
\\
\subfloat[ART I predicting vibration 20 seconds in the future for one flight.]{\includegraphics[width=.48\textwidth, height=.20\textheight]{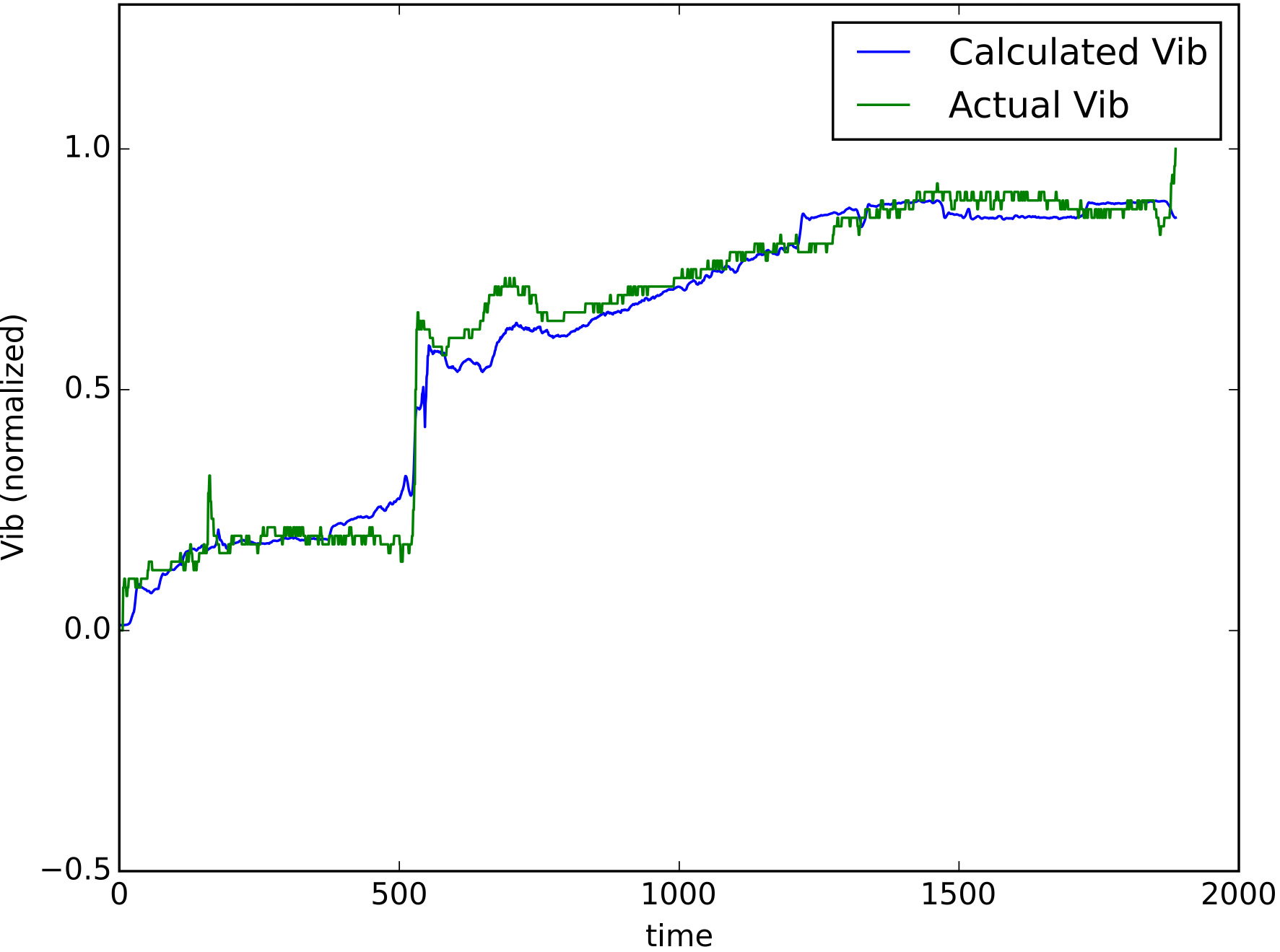}
\label{fig:arti-single_flight_20}}
\subfloat[ART III predicting vibration 20 seconds in the future for one flight.]{\includegraphics[width=.48\textwidth, height=.20\textheight]{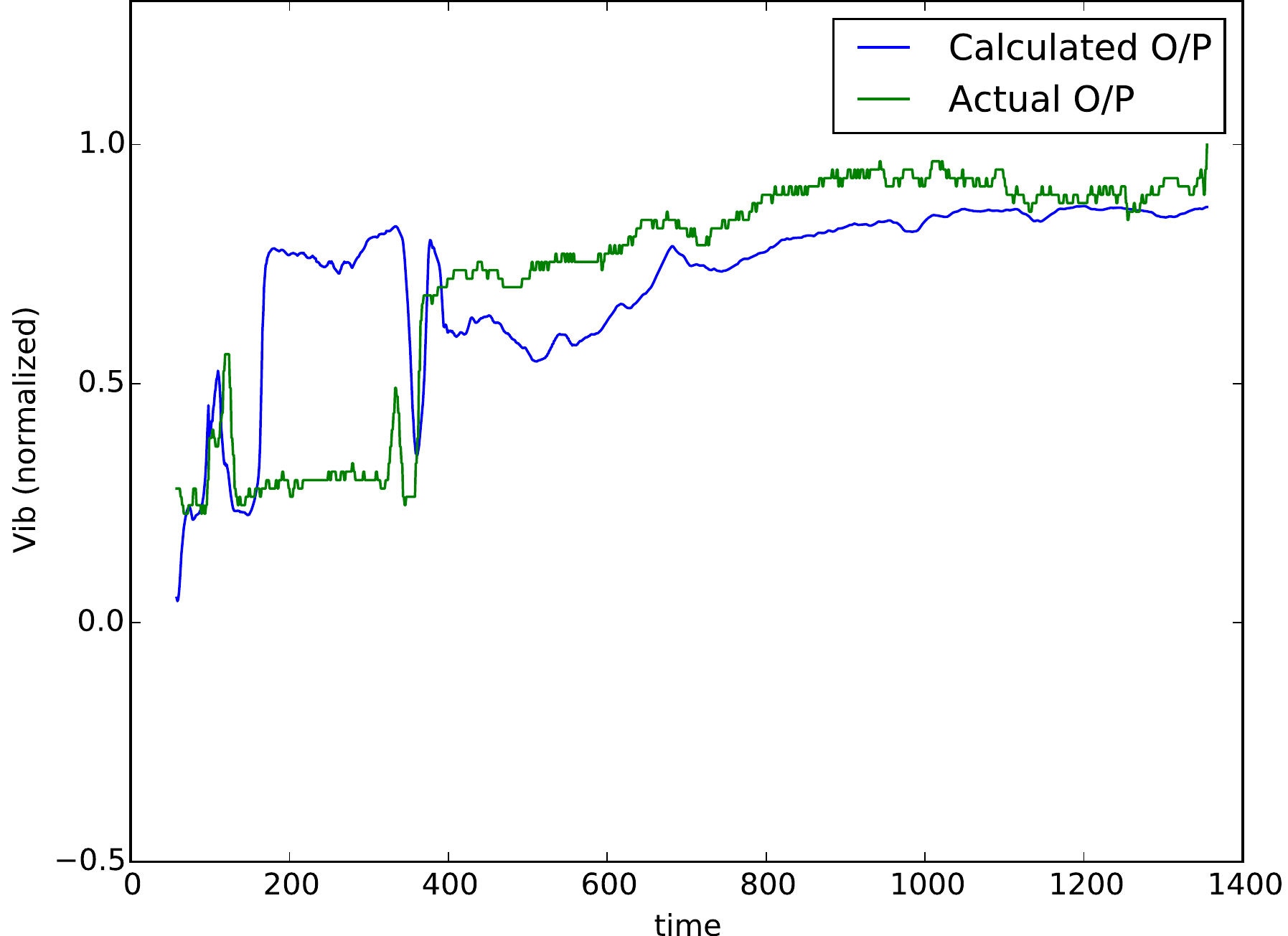}
\label{fig:artiii-single_flight_20}}
\caption{Architectures I and III predicting vibration for one flight.}
\label{fig:arti_artiii-single_flight}
\end{figure}

This LSTM RNN was one layer deeper and also had 20 seconds memory from the past which was not available for the other two LSTM RNNs used. Although it was the most computationally expensive and had the most chance for deeper learning, the results of this architecture were not as good as expected, as shown in Figure~\ref{fig:res_cnt}. Figures~\ref{fig:artiii-single_flight_5}, \ref{fig:artiii-single_flight_10}, \ref{fig:artiii-single_flight_20} provide an uncompressed example of Architecture III predicting vibration 5, 10 and, 20 seconds in the future over a single flight from the testing data. The results of this architecture in Table~\ref{table:results} show that the prediction accuracy for this architecture was less than the more simple Architecture I. While this came counter to the benefits of deeper learning, it does opens door for investigating about how to further tune more complicated LSTM RNNs. 

The overall error in Table~\ref{table:results} for the prediction at 20 future seconds was relatively high. Looking at Figure~\ref{fig:res_artiii_20} between time 10,000-15,0000, 20,000-25,000 and 35,000-40,000, it can be seen that the calculated curve got very much higher than the actual vibration curve. This strange behavior is unique as it can be seen that the calculated vibration would rarely exceed the actual vibration for all the curves plotted for all the architectures at all scenarios, and it would be for relatively small value if occurred. 

This network could potentially gain further improvement if trained for more epochs over the other simpler architectures since it was deeper. This was tried, giving the neural network about double the number of training epochs. However, a significant improvement in the prediction was not achieved. Nonetheless, it is of note that the plots of the cost function of this architecture were not smooth while trained to predict for 20 seconds in the future. This could potentially be a result of under-training or other issues in the training process. 

Initially, the training epochs were fixed at 575 for all the architecture as a standard for performance comparison. Further, the performance of this architecture (the mean absolute error) was slightly better than the other architectures when predicting for 1 second in the future. This result supports the believe that this architecture can perform better if given the chance to train for more epochs.

\subsection{ACO Results}
As Architecture I gave the most promising results, it was chosen as the initial candidate for the ACO.  The ACO code was run for 1000 iterations with a chosen number of 200 ants. Each run took approximately 4 days. The neural networks were run trained against flights that suffered from the excessive vibration.  They were then evaluated against different set of test flights, which also suffered from the same problem. There were 28 flights in the training set, with a total of 41,431 seconds of data. There were 29 flights in the testing set, with a total of 38,126 seconds of data. The networks were allowed to train for 575 epochs to learn and for the cost function output curve to flatten. The minimum value for the pheromones were 1 and the maximum was 20. The pheromones increased only when the network was found to give better fitness than the best fitness in the ACO generated population. The population size was equal to number number of iterations in the ACO process, \ie, the population size was also 1000.

\begin{figure}
\centering
\subfloat[Unoptimized: all test flights]{\includegraphics[width=.48\textwidth, height=.20\textheight]{./{plt_4.1.3_10}.pdf}
\label{fig:ant_all_flt_opt}}
\subfloat[Unoptimized: one test flights]{\includegraphics[width=.43\textwidth, height=.20\textheight]{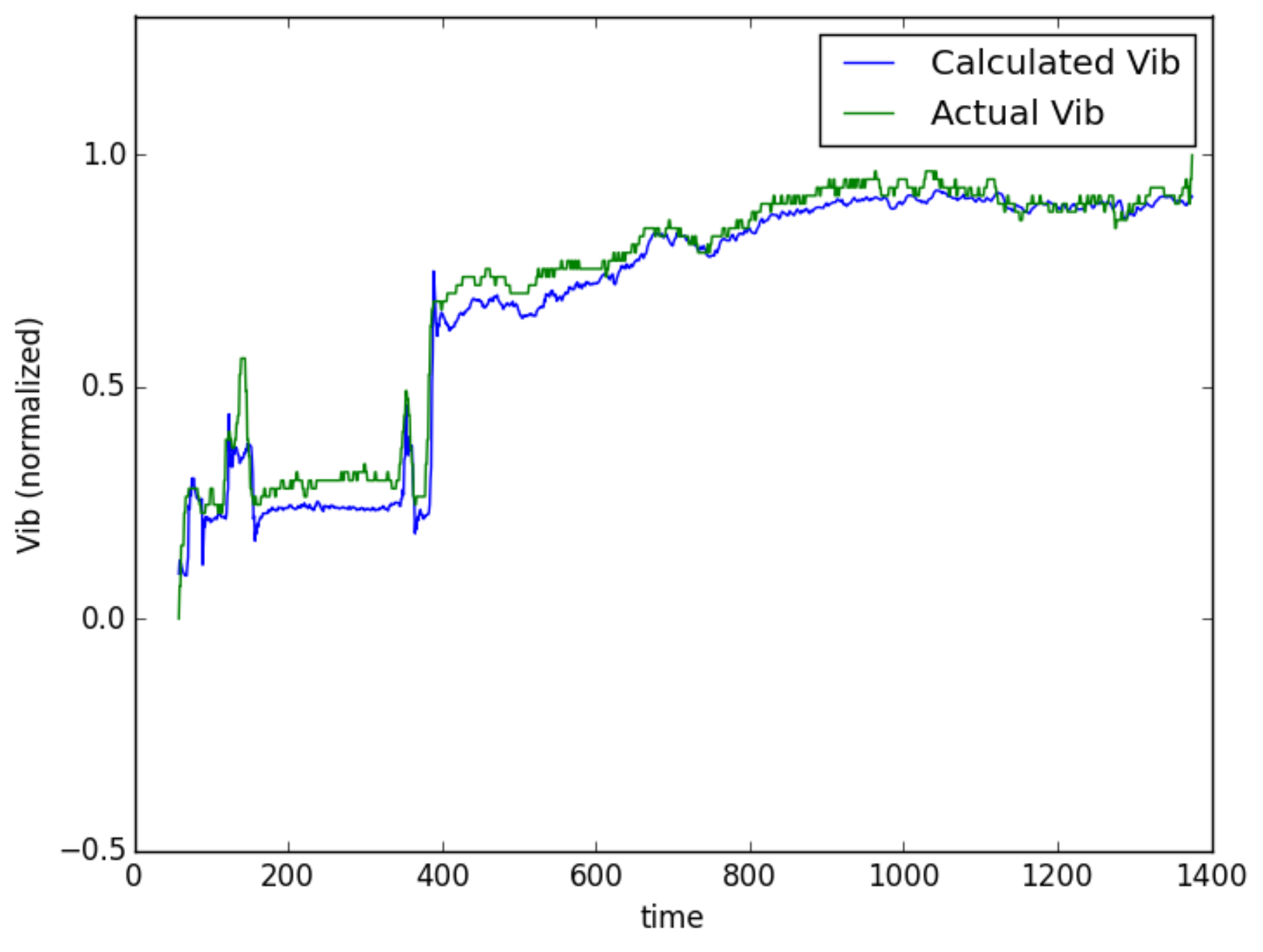}
\label{fig:ant_one_flt_opt}}
\\
\subfloat[Optimized: all test flights]{\includegraphics[width=.48\textwidth, height=.20\textheight]{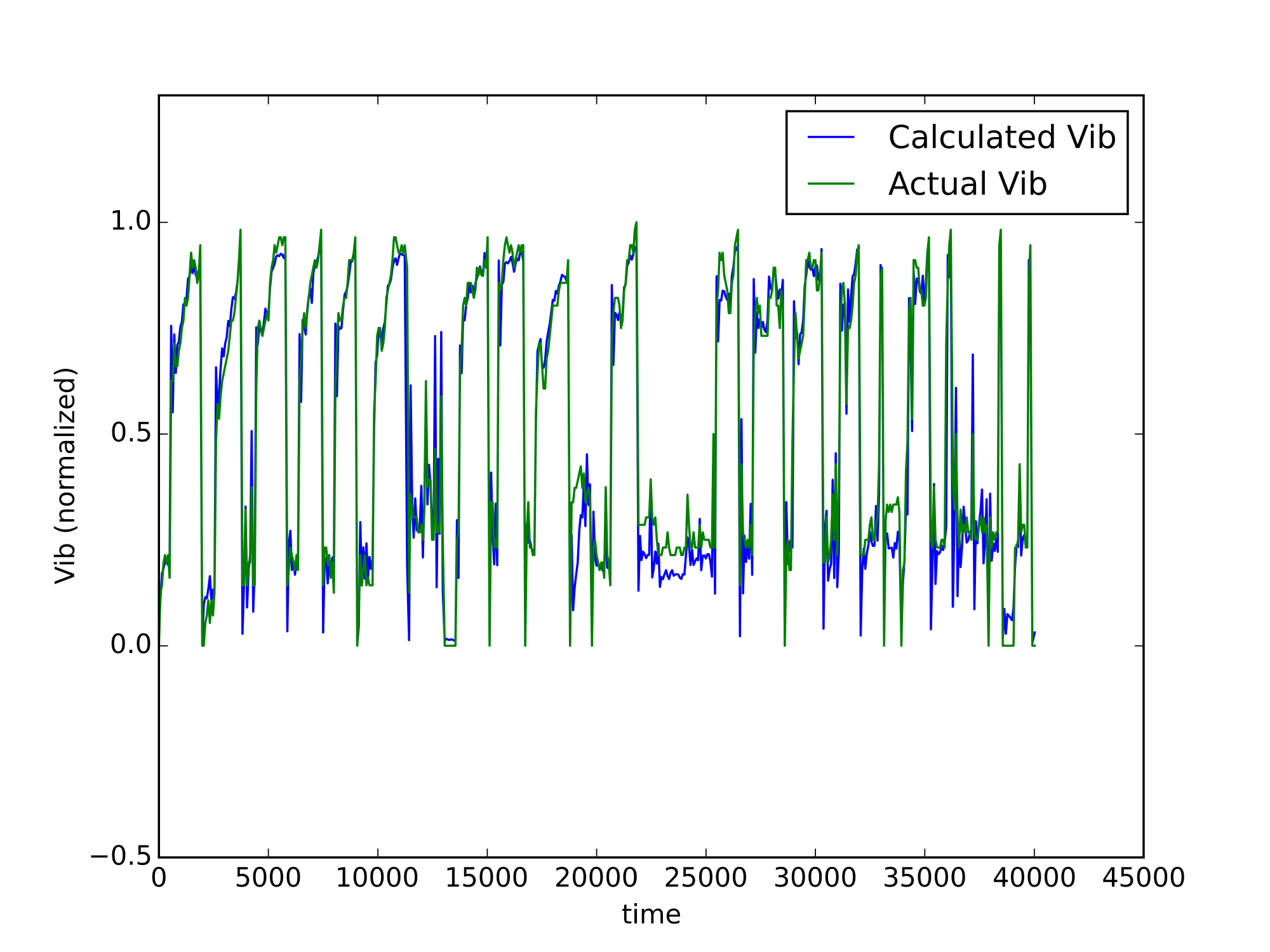}
\label{fig:ant_all_flt_un_opt}}
\subfloat[Optimized: one test flights]{\includegraphics[width=.48\textwidth, height=.20\textheight]{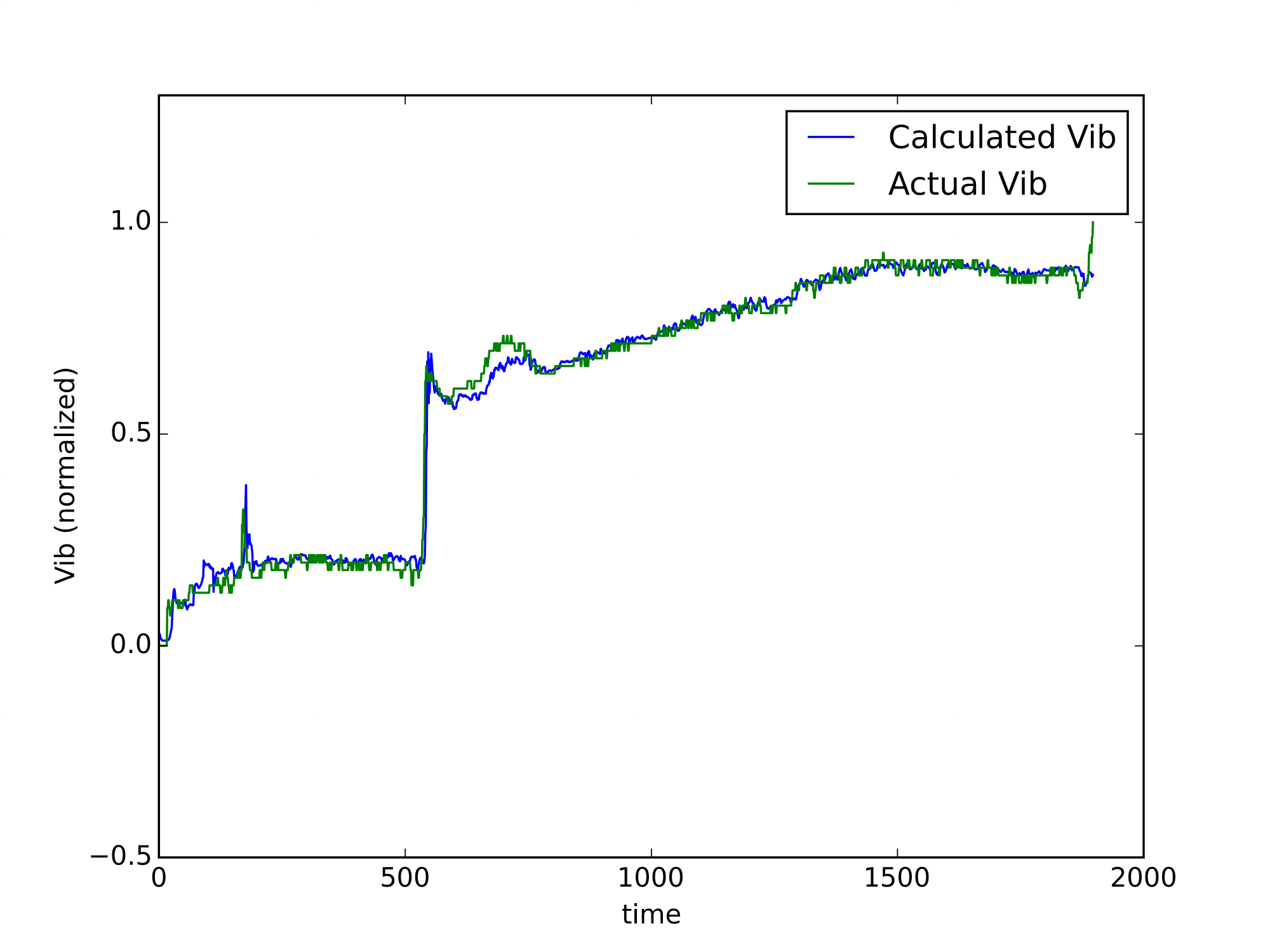}
\label{fig:ant_one_flt_un_opt}}
\caption{Plotted results for predicting ten seconds in the future.}
\label{fig:ant_res}
\end{figure}

\begin{figure}
\begin{center}
\includegraphics[width=.85\textwidth, height=.20\textheight]{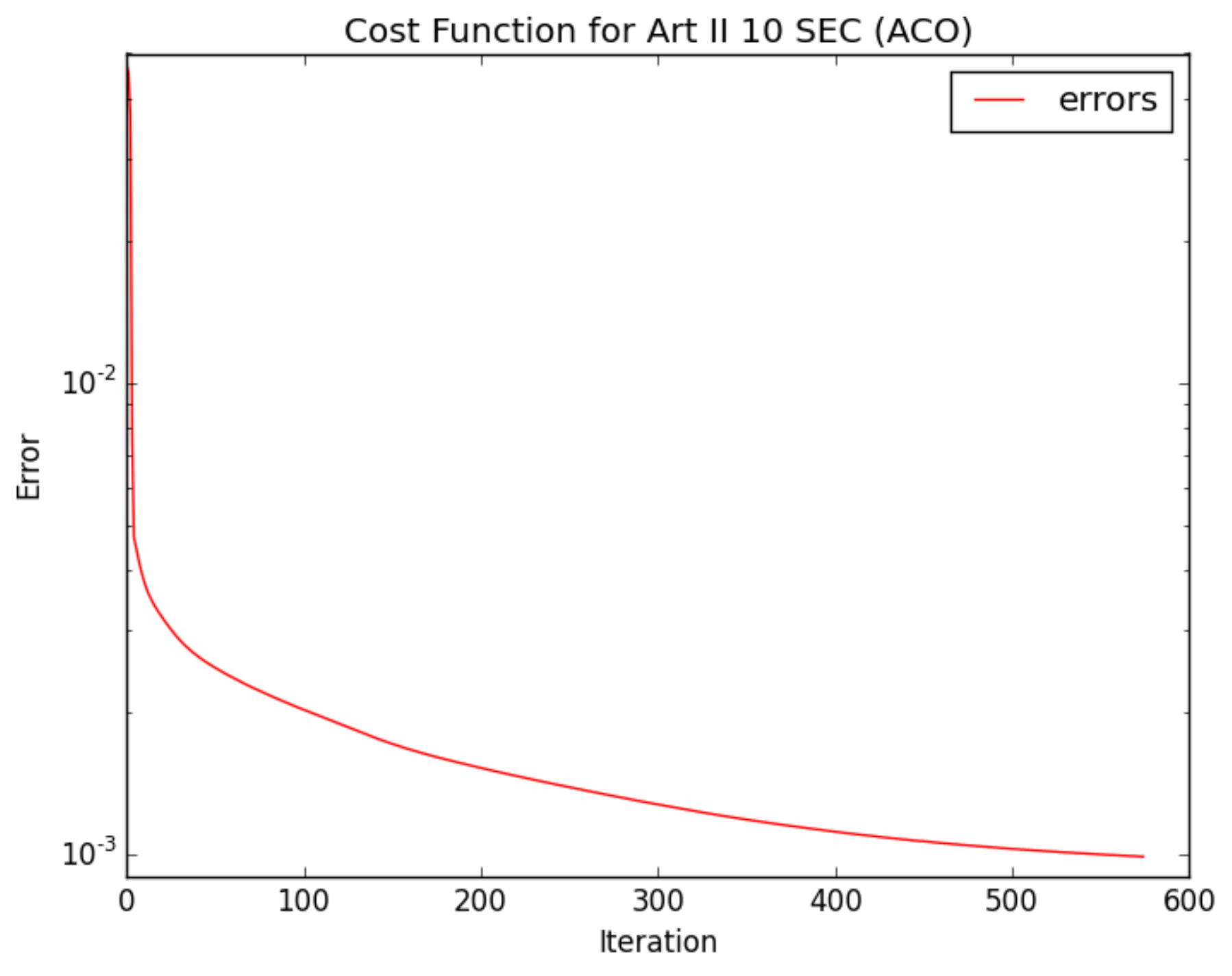}
\end{center}
\caption{ \label{fig:ant_cost} Cost function plot for ACO optimized ART I predicting vibration in 10 future sec}
\end{figure}

The best version of Architecture I evolved with ACO showed an improvement of 1.35\% for predictions 10 seconds in the future, reducing prediction error from 5.51\% to 4.17\% compared to the architecture's performance before the ACO. Results of the ACO are shown in Figure~\ref{fig:ant_one_flt_opt} for Architecture I for a single test flight and they are compared to the performance of the same architecture also for a single test flight. Figures~\ref{fig:ant_all_flt_un_opt}, \ref{fig:ant_all_flt_opt} show the results for the all the test flight for the architecture, before and after the ACO. The plot of the cost function of the training process of the best evolved network is shown in Figure~\ref{fig:ant_cost}.

Returning to an initial question of how the number of the connections in the network affects the soundness of the results, Table~\ref{table:top_three_evolve} shows the top thirty evolved networks with respect to the fitnesses they provide. 
The table also shows the total number of connections in both $mesh\_1$ (first set of connections in the generated mesh) and $mesh\_2$ (second set of connections in the generated mesh), and the total number weights (connections) in the networks. Comparing these values to the total number of weights in a fully connected Architecture I type network, as shown in Table~\ref{table:wghts_elmts}, it is found that total number of weights were reduced by 42\% to 45\% in the top 30 networks.

The ACO generated mesh (as defined in Algorithm~\ref{antcol_code}) used to generate this topology is shown in the matrices in Equations~\ref{eq:mesh_1} and~\ref{eq:mesh_2}. It is worth stressing that this topology is not the complete LSTM RNN used in the utilized Architecture I, but rather applies to the individual gates in each cell. Equation~\ref{eq:mesh_1} is used for any fully connected process and Equation~\ref{eq:mesh_2} is used for any data-reduction process (This is discussed in details in Section~\ref{sec:algorithm}).

\begin{figure}
	\centering
	\includegraphics[width=.98\textwidth]{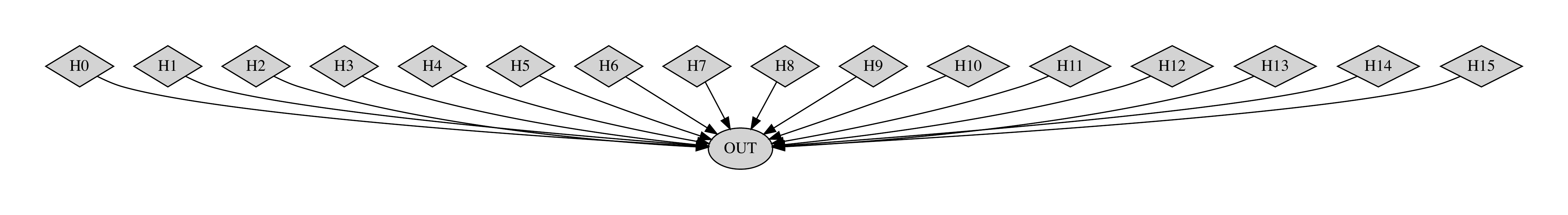}
	\caption{\label{fig:sub2_ant_art_second_best} ACO Architecture I Best Fitness Topology's mesh (Equation~\ref{eq:mesh_2}) at ``M2'' (Figure~\ref{fig:M2}) LSTM cells: 1000 Iterations, and 200 Ants.}
\end{figure}

The topology of the design of the networks' cells are shown in Figures~\ref{fig:ant_best_fit} and~\ref{fig:sub2_ant_art_second_best}. Figures~\ref{fig:ant_art_first_best},~\ref{fig:ant_art_second_best}, and~\ref{fig:ant_art_third_best} show the ACO optimized $mesh\_1$, which were used within the ``M1'' LSTM cells (see Figure~\ref{fig:M1}) in the top three evolved LSTM RNNs.
Equation~\ref{eq:mesh_1} represents $mesh\_1$ of the best evolved neural network which was used to generate Figure~\ref{fig:ant_art_first_best}.

\begin{figure*}
\centering
\subfloat[ACO Architecture I First Best Fitness Network: LSRM M1 cell.]{\includegraphics[width=.45\textwidth, height=.95\textheight]{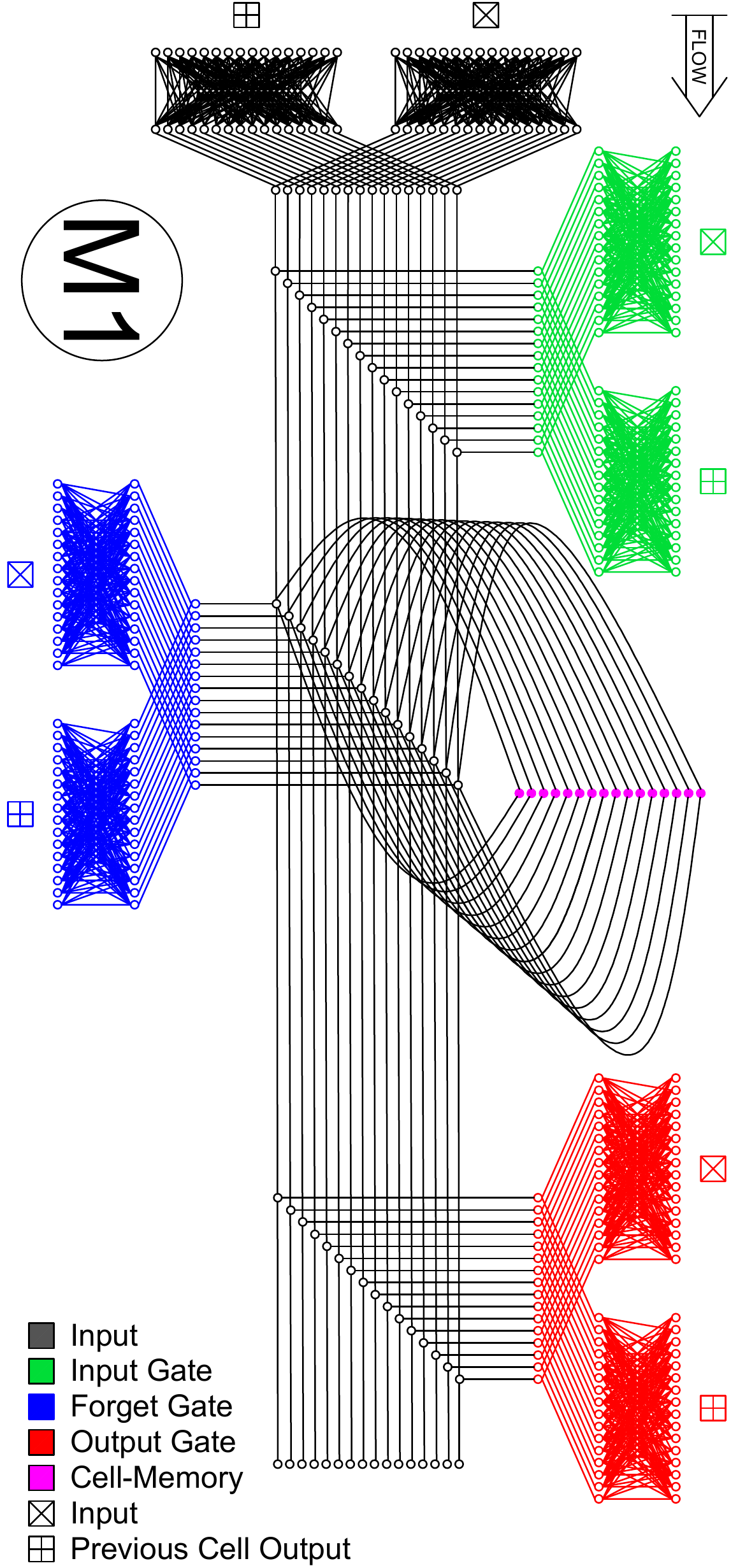}
\label{fig:m1_optimized}}
\hspace{.2cm}
\vline
\hspace{.2cm}
\subfloat[Meshes before (up) and after (down) optimization.]{\includegraphics[width=.45\textwidth, height=.95\textheight]{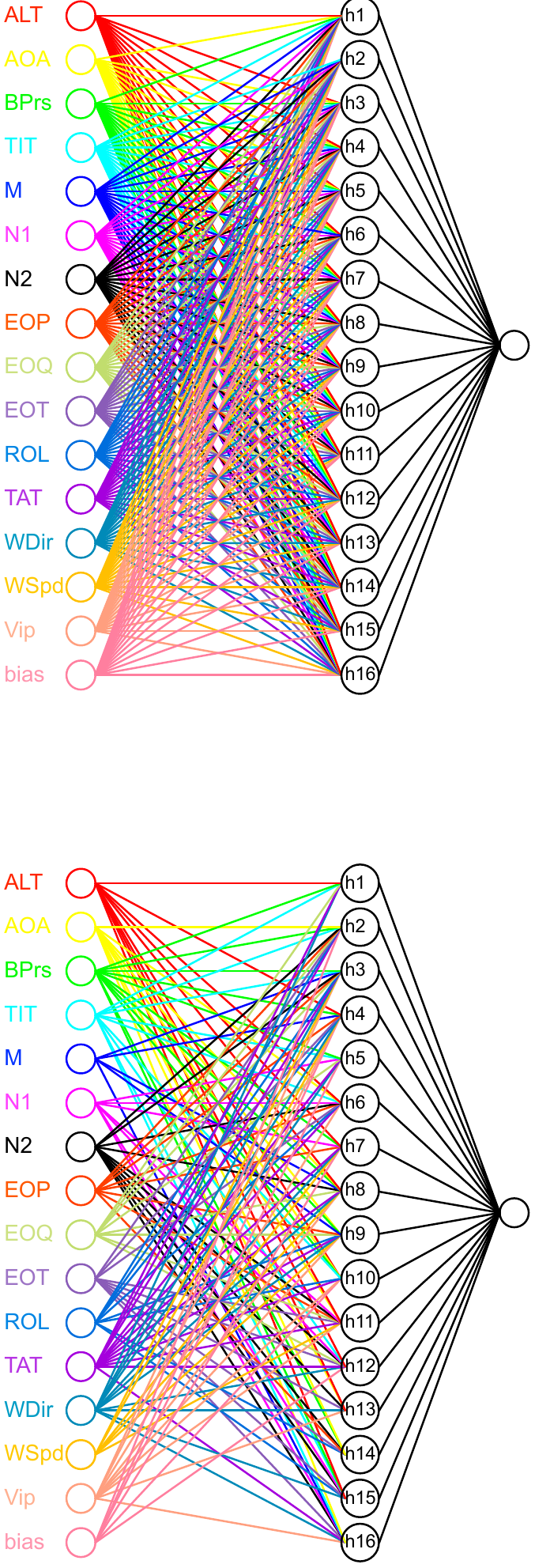}
\label{fig:meshes_opt_nonopt}} 
\end{figure*}

The evolved networks retained all the elements of $mesh\_2$ , represented by Equation~\ref{eq:mesh_2}, for use in the ``M2'' LSTM cells (see Figure~\ref{fig:M2}). Figure~\ref{fig:sub2_ant_art_second_best} is used to show this part of the evolved mesh. For clarity, Figure~\ref{fig:meshes_opt_nonopt} shows the differences between the M1 cells before and after ACO optimization. Figure~\ref{fig:m1_optimized} is simply a LSTM  cell ``M1'' that have its gates' meshes (shown in Figure~\ref{fig:meshes_opt_nonopt}, Up) substituted with the ACO meshes (shown in Figure~\ref{fig:meshes_opt_nonopt}, Down). ``M2'' did not change from its original topology as shown in Figure~\ref{fig:M2} since all the elements in $mesh\_2$ after the optimization remained ones (Equation~\ref{eq:mesh_2}).

The colored nodes in Figure~\ref{fig:ant_art_first_best} are the input nodes (first line of nodes) at the {\color{black} Main}, {\color{green} Input}, {\color{blue} Forget}, and {\color{red} Output} gates at the ``M1'' cells (Figure~\ref{fig:M1}). The diamond nodes in Figure~\ref{fig:ant_art_first_best} are the hidden layer nodes (second line of nodes) at the {\color{black} Main}, {\color{green} Input}, {\color{blue} Forget}, and {\color{red} Output} gates at the ``M1'' cells (Figure~\ref{fig:M1}). The diamond nodes are also the input nodes at the {\color{black} Main}, {\color{green} Input}, {\color{blue} Forget}, and {\color{red} Output} gates at the``M2'' cells (Figure~\ref{fig:M2}). The last single node in Figure~\ref{fig:sub2_ant_art_second_best} is the output of the gates in the ``M2'' cells.

\begin{equation}
\label{eq:mesh_1}
\tiny
\textit{mesh\_1}= 
\begin{array}{@{}*{19}{r}@{}}
i1\\
i2\\
i3\\
i4\\
i5\\
i6\\
i7\\
i8\\
i9\\
i10\\
i11\\
i12\\
i13\\
i14\\
i15\\
bias\\
\end{array} 
\left| \begin{array}{@{}*{19}{c}@{}}
{{h1}} & {h2} & {h3} & {h4} & {h5} & {h6} & {h7} & {h8} & {h9} & {h10} & {h11} & {h12} & {h13} & {h14} & {h15} & {h16} \\
0 & 1 & 0 & 1 & 1 & 0 & 1 & 0 & 0 & 1 & 1 & 1 & 1 & 0 & 1 & 0 \\
0 & 0 & 0 & 1 & 0 & 1 & 1 & 0 & 0 & 1 & 1 & 0 & 1 & 1 & 0 & 1 \\
0 & 0 & 0 & 0 & 0 & 1 & 1 & 0 & 1 & 1 & 0 & 0 & 0 & 1 & 1 & 0 \\
0 & 1 & 0 & 1 & 0 & 0 & 1 & 0 & 0 & 1 & 0 & 0 & 0 & 0 & 1 & 1 \\
1 & 0 & 0 & 1 & 1 & 0 & 1 & 1 & 0 & 1 & 1 & 0 & 1 & 0 & 0 & 1 \\
0 & 1 & 1 & 0 & 1 & 0 & 1 & 1 & 1 & 1 & 1 & 0 & 0 & 0 & 1 & 1 \\
1 & 1 & 0 & 0 & 1 & 1 & 0 & 0 & 0 & 0 & 0 & 1 & 1 & 1 & 1 & 1 \\
0 & 1 & 1 & 1 & 1 & 1 & 0 & 1 & 1 & 0 & 0 & 1 & 0 & 1 & 1 & 0 \\
0 & 1 & 1 & 1 & 1 & 0 & 0 & 1 & 0 & 1 & 0 & 1 & 0 & 1 & 1 & 1 \\
1 & 0 & 1 & 1 & 1 & 0 & 0 & 0 & 1 & 0 & 1 & 0 & 0 & 1 & 0 & 1 \\
0 & 1 & 0 & 1 & 1 & 1 & 0 & 1 & 1 & 1 & 0 & 1 & 0 & 0 & 1 & 1 \\
1 & 0 & 0 & 1 & 1 & 0 & 1 & 0 & 0 & 0 & 1 & 0 & 1 & 1 & 0 & 0 \\
1 & 1 & 1 & 1 & 0 & 1 & 1 & 0 & 1 & 0 & 0 & 1 & 0 & 1 & 0 & 1 \\
0 & 1 & 1 & 1 & 1 & 1 & 0 & 1 & 1 & 1 & 0 & 1 & 1 & 1 & 0 & 0 \\
1 & 0 & 0 & 0 & 1 & 1 & 1 & 1 & 1 & 1 & 0 & 0 & 1 & 1 & 1 & 0 \\
0 & 0 & 0 & 1 & 1 & 0 & 0 & 1 & 0 & 0 & 0 & 0 & 1 & 1 & 0 & 1 \\
\end{array} \right| 
\end{equation}

\begin{equation}
\label{eq:mesh_2}
\tiny
\textit{mesh\_2}=
\left| \begin{array}{@{}*{19}{c}@{}}
{{h1}} & {h2} & {h3} & {h4} & {h5} & {h6} & {h7} & {h8} & {h9} & {h10} & {h11} & {h12} & {h13} & {h14} & {h15} & {h16} \\
1 & 1 & 1 & 1 & 1 & 1 & 1 & 1 & 1 & 1 & 1 & 1 & 1 & 1 & 1 & 1 \\
\end{array} \right|
\end{equation}

\begin{table}
\centering
\caption{ACO Top Thirty Evolved Networks}
\label{table:top_three_evolve}
\resizebox{0.9\textwidth}{!}
{%
\begin{tabular}{ccccc|ccccc}
    \toprule
    \multicolumn{1}{c}{\bfseries No.} &
    \multicolumn{1}{c}{\bfseries Fitness} &
    \multicolumn{1}{c}{\bfseries Number} &
    \multicolumn{1}{c}{\bfseries Number} &
	\multicolumn{1}{c|}{\bfseries Total Number } &
    \multicolumn{1}{c}{\bfseries No.} &
    \multicolumn{1}{c}{\bfseries Fitness} &
    \multicolumn{1}{c}{\bfseries Number} &
    \multicolumn{1}{c}{\bfseries Number} &
	\multicolumn{1}{c}{\bfseries Total Number }\\
	&
	&
	\multicolumn{1}{c}{\bfseries of M1} &
	\multicolumn{1}{c}{\bfseries of M2} &
	\multicolumn{1}{c|}{\bfseries of} 	&
	&
	&
	\multicolumn{1}{c}{\bfseries of M1} &
	\multicolumn{1}{c}{\bfseries of M2} &
	\multicolumn{1}{c}{\bfseries of}\\
	&
	&
	\multicolumn{1}{c}{\bfseries Connections} &
	\multicolumn{1}{c}{\bfseries Connections} &
	\multicolumn{1}{c|}{\bfseries Connections} &
	&
	&
	\multicolumn{1}{c}{\bfseries Connections} &
	\multicolumn{1}{c}{\bfseries Connections} &
	\multicolumn{1}{c}{\bfseries Connections}\\
    \midrule
    1	& 0.041723	& 139	& 16 & 11,810 & 16 & 0.044603 	& 134	& 16 & 11,410 \\
    2	& 0.041927	& 136	& 16 & 11,570 & 17 	& 0.044680 	& 132	& 16 & 12,050 \\
    3	& 0.042549	& 144	& 16 & 12,210 & 18 	& 0.044719 	& 140	& 16 & 11,890 \\
    4	& 0.043000	& 137	& 16 & 11,650 & 19 	& 0.044838 	& 141	& 16 & 11,970 \\
    5	& 0.043134	& 137	& 16 & 11,650 & 20 	& 0.044922 	& 141	& 16 & 11,970 \\
    6 	& 0.043399 	& 138	& 16 & 11,730 & 21 	& 0.044949 	& 143	& 16 & 12,130 \\
    7 	& 0.043434 	& 137	& 16 & 11,650 & 22 	& 0.044982 	& 139	& 16 & 11,810 \\
    8 	& 0.043456 	& 143	& 16 & 12,130 & 23 	& 0.045013 	& 133	& 16 & 11,330 \\
    9 	& 0.043806 	& 139	& 16 & 11,810 & 24 	& 0.045117 	& 137	& 16 & 11,650 \\
    10 	& 0.043986 	& 144	& 16 & 12,210 & 25 	& 0.045240 	& 136	& 16 & 11,570 \\
    11 	& 0.044109 	& 141	& 16 & 11,970 & 26 	& 0.045302 	& 147	& 16 & 12,450 \\
    12 	& 0.044261 	& 137	& 16 & 11,650 & 27 	& 0.045306 	& 144	& 16 & 12,210 \\	
    13 	& 0.044362 	& 143	& 16 & 12,130 & 28 	& 0.045363 	& 147	& 16 & 12,450 \\
    14 	& 0.044483 	& 142	& 16 & 12,050 & 29 	& 0.045367 	& 133	& 16 & 11,330 \\
    15 	& 0.044493 	& 136	& 16 & 11,570 & 30 	& 0.045397 	& 146	& 16 & 12,370 \\
    
    \bottomrule
	
\end{tabular}
}
\end{table}

%
%

{
\begin{figure*}
\centering
\subfloat[ACO Architecture I First Best Fitness Mesh: 155 connections.]{\includegraphics[width=.3\textwidth, height=.8\textheight]{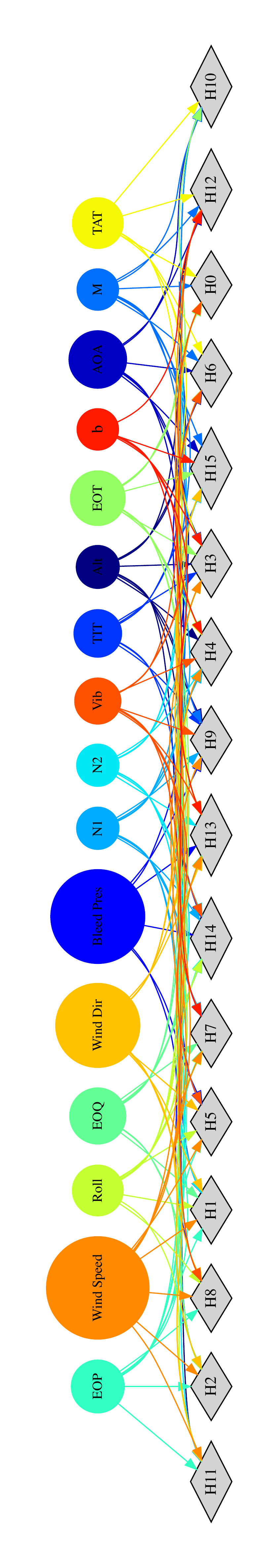}
\label{fig:ant_art_first_best}}
\quad
\subfloat[ACO Architecture I Second Best Fitness Mesh: 152 connections.]{\includegraphics[width=.3\textwidth, height=.8\textheight]{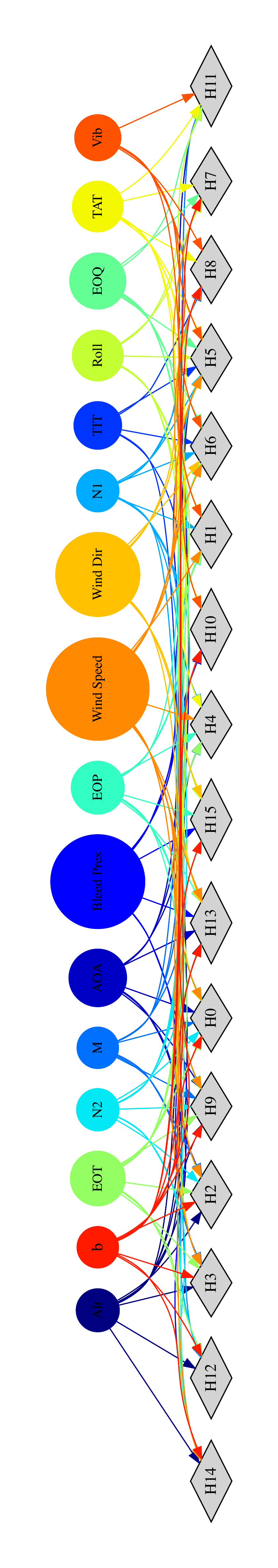}
\label{fig:ant_art_second_best}}
\quad
\subfloat[ACO Architecture I Third Best Fitness Mesh: 160 connections.]{\includegraphics[width=.3\textwidth, height=.8\textheight]{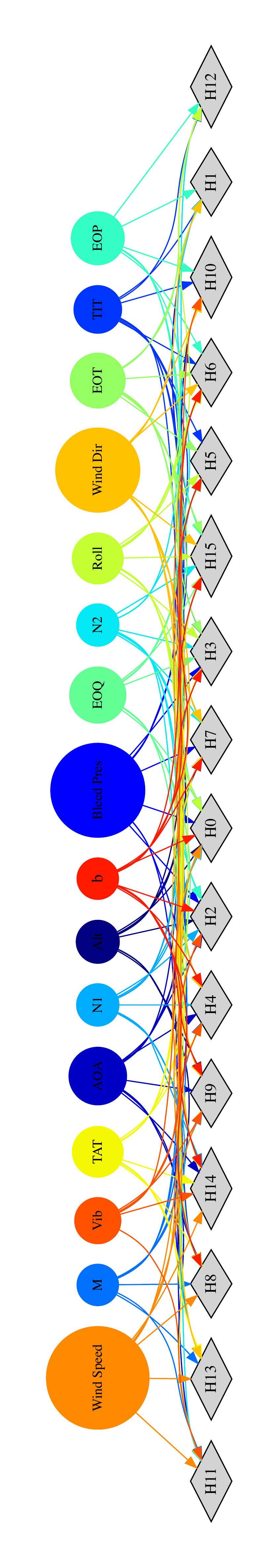}
\label{fig:ant_art_third_best}}
\caption{ACO Architecture I Best Fitness Topologies' Meshes (Equation~\ref{eq:mesh_1}) for at ``M1'' (Figure~\ref{fig:M1}) LSTM cells: 1000 Iterations, and 200 Ants.}
\label{fig:ant_best_fit}
\end{figure*}
}

\section{Discussion and Future Work}
\label{sec:conclusion}

The results have shown that the ACO approach for optimizing the gates within LSTM cells can dramatically reduce the number of connections required, while at the same time improve the predictive ability of the recurrent neural network.  However, as much of the matrix in Equation~\ref{eq:mesh_1} is sparse, none of the rows of this matrix had all their elements equal to zero, meaning that none of the inputs ended up being dropped out (which was a potential means for improving predictions, as discussed in Subsection~\ref{subsec:aero_select}). This is a indication that all the chosen parameters actually had a positive contributing influence on the vibration. On the other hand, it suggests that having additional connections can increase the difficulty of appropriately training the LSTM RNN, resulting in less predictive ability (as in the case of the original unoptimized LSTM RNN architectures).

This approach can open the door to further identify the highest contributors to the vibration problem by examining the number of the connections between the input neurons and the hidden layers, along with the magnitude of those weights. Furthermore, the combined effect of multiple parameters can also be investigated by looking at the input neurons connected to a certain hidden layer's neuron. This would give an idea about the effect those input neurons, which represents the input parameters, have on the vibration as a final output, which could aid in discovering actual cause of the vibration events.

This also opens up the potential for significant future work. While the optimized LSTM RNNs did not drop out any input connections, by increasing the number of flight parameters used as input (even using all available parameters), the algorithm has the potential to determine which parameters contribute most to the predictive ability, instead of relying on {\it a priori} expert knowledge to select parameters. Further, in this work one mesh of connections was generated and then used in all the LSTM cell gates at all time-steps. The future work will consider the meshes in the four gates of the LSTM cells at the various LSTM time-steps as variable and will apply the ACO on each of them simultaneously in every ACO iteration.

Future work will also consider optimizing the LSTM RNN structure. The connections of the structure shown in Figures~\ref{fig:art1_full} and ~\ref{fig_arts} will be subject to ACO process along with the optimization of the connections within the LSTM cells. This has the potential to make a large step forward in the evolution of the LSTM RNNs as it will allow for connections between non-adjacent cells, and potentailly even dropping out certain unused cells and potentailly even full layers from the LSTM RNN. 

Lastly, work investigating the tuning of the ACO hyperparameters can be done to improve how quickly the algorithm converges to optimal LSTM RNN structures. For example, modifying the number of ants, reducing pheromones on paths from LSTM RNNs with lower fitness, and periodically refreshing the pheromones levels by decreasing all of its levels by certain amount. 

\section*{Acknowledgments}
\footnotesize
We very much appreciate the help, patience and support of \textbf{\textit{Mr. Aaron Bergstrom}} of the University of North Dakota's Computational Research Center (CRC). This work used the high performance computing clusters at the CRC, where Mr. Bergstrom (the North Dakota University System HPC Specialist, UND Big Data Project Coordinator, and UND Campus Champion) offered his time and effort to much facilitate it. It was a pleasure working with him.

\section*{References}

\bibliography{bibliography_merged}

\end{document}